\documentclass[12pt,twoside]{report}
\usepackage[utf8]{inputenc}
\usepackage{graphicx}
\usepackage[a4paper,top=25mm,bottom=25mm,left=5mm, right= 5mm,bindingoffset=6mm]{geometry}
\graphicspath{ {images/} }
\usepackage{amsmath,amssymb,bm,amsthm,amsfonts}
\usepackage{comment}
\usepackage{anysize}
\usepackage{graphicx}
\usepackage{authblk}
\usepackage{float}
\usepackage{subfig}
\usepackage{exscale}
\usepackage{mathtools}
\usepackage{siunitx}
\usepackage{mathrsfs}
\usepackage{authblk}
\usepackage{mdframed}
\usepackage{hyperref}
\hypersetup{colorlinks=true,linkcolor=black,citecolor=black}   % marcadores no documento .pdf
\usepackage[ruled,vlined]{algorithm2e}
\usepackage{arydshln} %linha pontilhada
\usepackage{subfig} %side by side figures
\usepackage[font=small,labelfont=bf]{caption}
\usepackage{microtype}
\usepackage{breakcites}
\newcommand{\bfbeta}{\bm{\beta}}
\newcommand{\bfx}{\mathbf{X}}
\newcommand{\bbfx}{\mathbf{x}}
\newcommand{\bfe}{\mathbf{E}}
\newcommand{\bfy}{\mathbf{Y}}
\newcommand{\bbfy}{\mathbf{y}}
\newcommand{\bfb}{\mathbf{B}}
\newcommand{\bfhb}{\hat{\mathbf{B}}}
\newcommand{\bfj}{\mathbf{J}}
\newcommand{\bfs}{\mathbf{S}}
\newcommand{\bfi}{\mathbf{I}}
\newcommand{\bfv}{\mathbf{V}}
\newcommand{\bfq}{\mathbf{Q}}
\newcommand{\bfg}{\mathbf{G}}
\newcommand{\bfm}{\mathbf{M}}
\newcommand{\bfeps}{\bm{\varepsilon}}
\newcommand{\bfk}{\mathbf{K}}
\newcommand{\bflambda}{\bm{\Lambda}}
\newcommand{\bfalpha}{\bm{\alpha}}
\newcommand{\bfhbeta}{\hat{\bm{\beta}}}

\newcommand{\bfgamma}{\bm{\gamma}}
\newcommand{\bftheta}{\bm{\theta}}
\newcommand{\bfdelta}{\bm{\Delta}}
\newcommand{\bfxi}{\mathbf{\Xi}}
\newcommand{\bbe}{\mathbb{E}}

\newcommand{\mbbx}{\mathbf{x}}
\newcommand{\kl}{\text{KL}}
\newcommand{\elbo}{\text{ELBO}}
\newcommand{\diag}{\text{diag}}
\newcommand{\tr}{\text{tr}}
\newcommand{\vect}{\text{vec}}
\newcommand{\p}{\text{p}}
\newcommand{\q}{\text{q}}
\newcommand{\var}{\text{Var}}
\newcommand{\cov}{\text{Cov}}
\newcommand{\der}{\text{d}}
\newtheorem{theo}{Definition}
\newtheorem*{remark}{Remark}
\newtheorem{theorem}{Theorem}

\begin{document}

%title page -------------------------------------------------

\begin{titlepage}

    \begin{center}
    
    \includegraphics[width=1.\textwidth]{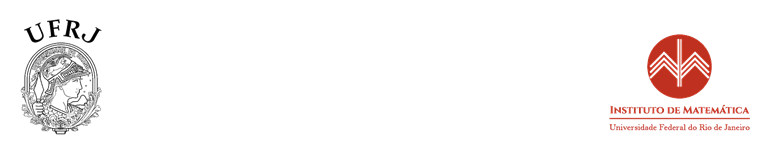}
    
    \Large
        Universidade Federal do Rio de Janeiro\\
        Instituto de Matemática \\
        Departamento de Métodos Estatísticos\\
 
         \vspace{5.cm}   
 
        \Huge
        \textbf{O Uso de Inferência Variacional para Reconhecimento de Emoções em Música}

        \vspace{1.cm}
        
        \large
        \textbf{Nathalie Fernanda Toffani Magliano di Domenico Fraçatho Deziderio}
 
        \vfill
        
        \normalsize
        Projeto final submetido ao Departamento de Métodos Estatísticos do Instituto de Matemática da Universidade Federal do Rio de Janeiro como parte dos requerimentos para obtenção do Bacharelado em Estatística.

        \vspace{0.8cm}
        
        Orientador: Hugo Tremonte de Carvalho

        Brasil, Rio de Janeiro, 5 de março de 2021
 
    \end{center}
\end{titlepage} 

%title page -------------------------------------------------

%fancy bit --------------------------------------------------
    \Huge{
    \textbf{Agradecimentos}} \\

\normalsize
Refletindo sobre todo o trabalho realizado e todas as pessoas que acompanharam minha trajetória até aqui, me vejo na privilegiada posição de ter tantas pessoas a agradecer pelo apoio que me deram ao longo do caminho. Pessoas essas que viram o bom, o mau e o feio na minha vida, sem nunca deixar que nossos laços se desfizessem. Todos os dias escolhemos fazer parte das histórias uns dos outros, nos acolhemos, nos valorizamos e aprendemos a abraçar completamente as peças que nos constituem. Serei eternamente grata pelos elos que me conectam a essas pessoas e pelo nosso esforço em preservá-los.

\vspace{5mm}

Não poderia começar sem agradecer à minha mãe, Marise Toffani, que sempre esteve ao meu lado e fez de tudo para que eu tivesse a melhor educação possível. Muito além disso, me deu apoio emocional em todos os momentos difíceis e me ouviu falar incessantemente sobre essa pesquisa mesmo detestando matemática com todas as forças.

\vspace{5mm}

Agradeço muito ao Aluisio, Cristina e Bárbara Mariante por serem pessoas tão amorosas e especiais na minha vida. Vocês são minha segunda família e têm meu eterno carinho.

\vspace{5mm}

Desde o início da minha vida acadêmica também tive a sorte de poder contar com tantos professores que ajudaram a me tornar quem sou. Meus queridos professores J. J., Cataldo e Felipe Ferreira, cujo carinho e didática impecável cultivaram meu amor pela Matemática durante o Ensino Médio. Ao meu Professor de Língua Portuguesa, Bernardo Miller, gostaria de agradecer pelos nossos diálogos sobre poesia e literatura que ainda enriquecem minha vida. Com certeza a qualidade desse texto está refletida nas suas rigorosíssimas aulas de produção escrita. Finalmente, gostaria de agradecer ao meu professor de teatro Rodrigo Gaia, que ajudou a desenvolver minha voz e habilidade de me fazer ser ouvida.

\vspace{5mm}

Ao longo da minha trajetória, também pude contar com amigos que espero levar para o resto da minha vida. Pessoas que me acompanharam desde os meus anos formativos, com quem posso dividir qualquer experiência e torná-la especial. Agradeço à Anna Carolina Bittencourt, Bruna Pessanha, Gabriela Carvalho, Julia Abib, Maria Emília, Patricia Menezes, Renata Soares, Teca Guedes e Viviane Romero por essa amizade. Seja para tomar vinho, ver um filme, brigar pelos nossos direitos, dividir uma refeição ou passar a noite toda conversando, sei que posso contar com a sua companhia. Foi um privilégio passar pelas agruras de se tornar mulher nesse mundo ao lado de vocês e hoje continuam sendo tão importantes quanto eram quando nos víamos todos os dias. Bernardo Werneck, Miguel Arraes, Thiago Duarte e Thiago Oliveira, todos os instantes passados com vocês viram as mais carinhosas lembranças. Não existe assunto sob o sol que não conseguimos transformar em discussões filosóficas (ou brigas homéricas) inesquecíveis. 

\vspace{5mm}

Já na UFRJ, pude conhecer novas pessoas que também passaram a ocupar um espaço importante na minha vida. Eduardo Leandro, Felipe Sadok, Gustavo Martins, Julia Cabral, Mariana Grijó, Mariana Ruma, Matheus Raposo, Rebeca Barros, Sérgio Vinícius e Stefanni Attina, poder partilhar o dia a dia da graduação com vocês tornou esse caminho muito mais feliz. Agradeço em especial ao Marcelo Carneiro, um grande amigo de impecável confiança. Agradeço também ao Vinícius Grijó pela sua ajuda tão cuidadosa na revisão deste trabalho. Meus queridos amigos do Matmúsica, que tanto apreciam ambos os assuntos discutidos aqui, me proporcionaram os mais deliciosos momentos. Começando por um agradecimento especial, destaco Iago Leal pelo seu incrível apoio para realizar esse trabalho. Sua placidez sem igual consegue acalmar até a mim quando tenho vontade de largar tudo e ir fazer apicultura. Agradeço também ao Bruno Lima Netto, Cynthia Herkenhoff, Felipe Paggineli, Gabriela Lewenfus, Pedro Xavier, Thiago Holleben por esses momentos. O fundador deste grupo, é alguém que merece muito mais que agradecimentos. Bernardo Freitas Paulo da Costa, obrigada por todo seu apoio e carinho, por ser tão bom para mim, por tantos momentos maravilhosos. Sua ajuda tornou este trabalho infinitamente melhor e sua presença alegra cada um dos meus dias. 

\vspace{5mm}

Agradeço também à minha amiga Maria Irene Tremonte, com quem consigo compartilhar bons momentos mesmo em meio a um turbilhão de dificuldades. 

\vspace{5mm}

Finalmente, agradeço aos professores Carlos Zanini e Carlos Pedreira por aceitarem fazer parte dessa banca, dando tanta atenção a um trabalho que tive infinito carinho para confeccionar, e ao meu orientador Hugo Carvalho pela sua dedicação e apoio. As valiosas lições que aprendi com você ao longo desses anos jamais serão esquecidas.

\newpage

    \Huge{
    \textbf{Resumo}} \\

\normalsize
Esse trabalho foi criado com o objetivo de aplicar técnicas de Estatística ao campo de Reconhecimento de Emoções em Música, uma área bem conhecida dentro do mundo de Processamento de sinais mas ainda pouco explorada do ponto de vista estatístico. Aqui, abrimos diversas possibilidades dentro do meio, aplicando técnicas modernas de Estatística Bayesiana de maneira a criar algoritmos eficientes, focando na aplicabilidade dos resultados. Apesar da motivação deste projeto ser a criação de um sistema de recomendação de músicas baseado em emoção, a principal contribuição aqui desenvolvida é um modelo multivariado altamente adaptável e pode ser útil para interpretar qualquer conjunto de dados onde se deseje aplicar regularização de maneira eficiente. De maneira geral, vamos explorar o papel que uma profunda análise teórica estatística pode ter na modelagem de um algoritmo que explore bem uma base de dados já conhecida e o que pode ser ganho com esse tipo de abordagem.
\newpage

    \Huge{
    \textbf{Abstract}} \\

\normalsize
This work was developed aiming to employ Statistical techniques to the field of Music Emotion Recognition, a well-recognized area within the Signal Processing world, but hardly explored from the statistical point of view. Here, we opened several possibilities within the field, applying modern Bayesian Statistics techniques and developing efficient algorithms, focusing on the applicability of the results obtained. Although the motivation for this project was the development of a emotion-based music recommendation system, its main contribution is a highly adaptable multivariate model that can be useful interpreting any database where there is an interest in applying regularization in an efficient manner. Broadly speaking, we will explore what role a sound theoretical statistical analysis can play in the modeling of an algorithm that is able to understand a well-known database and what can be gained with this kind of approach.

%fancy bit --------------------------------------------------

%indexes ----------------------------------------------------

\normalsize %keep the lists' items with normal size :)

\tableofcontents

\addcontentsline{toc}{chapter}{List of Tables}
\listoftables

\addcontentsline{toc}{chapter}{List of Figures}
\listoffigures

\addcontentsline{toc}{chapter}{List of Alghorithms}
\listofalgorithms

%indexes ----------------------------------------------------

%text ------------------------------------------------------- 

\chapter{Introduction}\label{introduction}

\begin{comment}

-AV model
-why it is relevant psychologically and for the problem

\end{comment}

The motivation for this work emerged from a deep love for music and the desire to better understand its relations to human psychology through Statistics. This led us into the world of Music Emotion Recognition (MER) \cite{Panda-dsc}, a research area within Music Information Retrieval, that has been steadily growing due to its challenging and thought-provoking nature, and which we briefly introduce below.% that arose from Music Information Retrieval in the Signal Processing field and has been steadily growing due to its challenging and thought provoking nature.

Music is an indispensable part of human culture and it is deeply intertwined with the emotional response it causes: ancient Greeks used it to tell epic stories about great wars and journeys that brought awe and heartbreak to their listeners;
the Apache people sang a war chant before striding into battle to give courage and unite their warriors; and even though Wolfgang Amadeus Mozart's motivation while composing \textit{``Le Nozze di Figaro"} was to criticize the royals in Versailles, we laugh and cry as the melody unfolds.

% Nowadays music is just as important in our lives as it was then, and
There are several studies that explore the relation of music to the human brain \cite{min82, jou08}, and one can also argue that nowadays it plays an even more important role in our lives than ever, since CD's, portable MP3 players and streaming services have made music broadly and easily accessible.
%is more accessible than ever before.
That overload of options, however, comes at a price: we are constantly drowning in new pieces from all over the world, with new artists and genres appearing every day, so it is not an easy task to keep track of what is relevant to each person. To tackle that issue, many streaming services have employed personalized recommendation systems \cite{son12}, that study what each user listens to and offer suggestions that are similar in style, time period, artist or genre, to name a few possibilities. Considering the aforementioned relationship between music and sentiment, it could be very interesting to create a system that takes this connection in consideration. In order to do that, we will explore several statistical methods to estimate emotions in music but before diving into it, let us set some boundaries, defining how this idea unfolds.

\begin{figure}[H]
    \centering
    \includegraphics[scale=0.5]{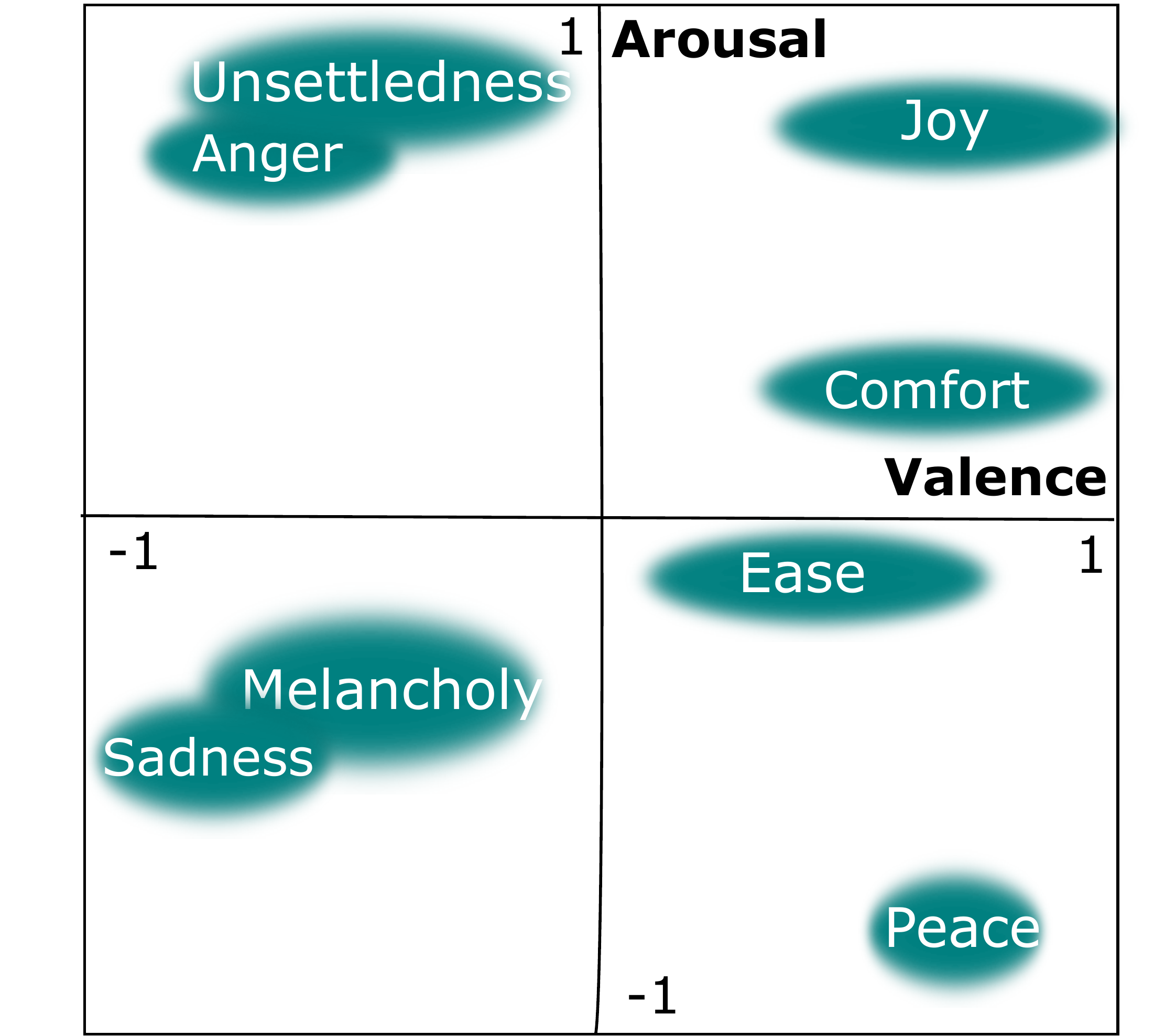}
    \caption[The Arousal-Valence Scale]{The Arousal-Valence Scale with examples of emotion placement.}
    \label{fig:AV-scale.pdf}
    \end{figure}

Since ``emotion" is a rather vague and imprecise target, the first step we need to take is coming up with more precise way to approach the subject. Exploring previous works in MER we came upon the Arousal-Valence (AV) scale, a measuring system created by psychologists in the 80's \cite{russell-av} thar has been widely used by the MER community. It measures respectively how agitated a song is (Arousal) and how positive or negative are the feelings it evokes (Valence), from $-1$ to $1$, as illustrated in Figure \ref{fig:AV-scale.pdf}. Since it is continuous rather than discrete, it more successfully addresses the fluid nature of emotions. Note that we seek to measure the sentiment a piece \textit{evokes} as opposed to the one a particular listener \textit{perceives}. This important distinction makes it possible to have consistent, straightforward data, independent from personal judgements that would be very difficult to predict without a great level of uncertainty.

It is easy to infer that Arousal and Valence - respectively A and V from now on - are positively correlated, since agitated songs tend to evoke more positive feelings and slower songs tend to evoke more negative feelings. Even though the exact correlation between A and V may vary with each dataset being considered, it is usually not negligible and, in the dataset here employed \cite{Ann17}, it reaches $0.59$.

Ignoring this fact can simplify the modeling process quite a lot, but taking it into consideration could be valuable. In particular, since Valence is significantly harder to estimate than Arousal \cite{yan08}, this correlation could help the model provide better estimates. Although that information was obtained empirically, it is not difficult to understand why that difference exists: intuitively, agitation is reasonably objective to perceive, as it is a movement. Positivity and negativity on the other hand are much more subtle, thus harder to put a pin on.

To predict the desired responses, we will build a model that explores the relation between a song's AV-values and several physical properties of the waveform. From a myriad of avaliable pieces of data, we choose acoustic features to work with. As we will detail in Chapter \ref{lit_review}, they are a reasonable choice in this scenario, since they reduce dimensionality by extracting important information contained within the waveform, being also easy to compute and, therefore, potentially numerous. However, in spite of their advantages, these low-level features are very lightly correlated to the response variables \cite{Panda-dsc}, and usual techniques will struggle from having a number of parameters potentially approaching the number of observations. To solve this particular issue, one can look for ways to select more important features in an efficient manner or calculate high-level features more relevant to the problem \cite{Panda-dsc}.

here, we seek to introduce more refined statistical techniques to perform feature selection automatically while estimating Arousal and Valence, and explore Bayesian inference to develop models that extract more interpretable results from the data. Apart from the theoretical improvement, we seek to produce a fast algorithm that can be easily adapted and then employed in a business setting, handling larger databases with multiple responses.

Moreover, we seek to introduce the MER problem to the statistical community, not only by use of the techniques employed here but also to encourage further investigation in the field. Still, we keep in mind applicability of the results and propose a preliminary prototype for an emotion-based recommendation system with statistical foundations, that helped motivate this endeavor.

After this brief introduction to the general basis of this work, Chapter \ref{lit_review} will carry a more detailed discussion on the subject of MER and literature review. Once we have a proper grasp of the field and some of its characteristics, we move on to Chapter \ref{database} where we offer some more detail into the database being employed. %so we are able to compare results between all approaches, discussing merits and downfalls of each one.
Statistical techniques will begin to be introduced in Chapter \ref{bayes}, more specifically by the discussion and application of Bayesian linear regression to the Arousal and Valence estimation problem, and Variational Inference within Automatic Relevance Determination (ARD) will be discussed in Chapter \ref{vi_ard}. Note that even though none of this approaches have been applied to the area of MER before, they are widely known in the Statistics community \cite{vi-rev-stat, cas01}. In Chapter \ref{MARD} we propose the Multivariate Automatic Relevance  Determination (MARD), a generalization of the ARD that is able to handle multiple correlated response variables at once. Finally, in Chapter \ref{Conclusions} we make the appropriate remarks on the entirety of the work done, as well as proposing further steps.

\chapter{A Brief Overview of Music Emotion Recognition}\label{lit_review}

\begin{comment}

-revise bibliography
-problems in the area
-weaknesses in other works
-our strengths

\end{comment}

Now that we have some notion of the Music Emotion Recognition (MER) problem itself and briefly presented some of our goals, let us walk through what has already been developed in the area in the past few decades. The first two sections of this chapter will focus on a literature review of the MER field: Section \ref{sec:disc} will describe discrete approaches %with some more detail,
while Section \ref{sec:cont} will account for continuous ones, more related to this work. In Section \ref{sec:our} we will briefly outline %what we will pursue here and
a new possibility to tackle this problem.

\section{Discrete approaches}\label{sec:disc}

The first attempt at estimating emotion from music was made in 1988 with a multi-label classification model \cite{Kat88}, with targets such as \textit{gloomy}, \textit{urbane}, \textit{pathetic}, \textit{serious}, \textit{hopeful} and others, but the results showed low classification accuracy. Even though different databases and label sets were contemplated by researchers throughout time, mostly categorical approaches were made until 2009. The challenge of using discrete values to address the fluid nature of the emotions %they were seeking to understand
led to scenarios where multiple overlapping categories (18 categories containing 135 tags) were used \cite{hu09}, or a few generic ones (\textit{happy}, \textit{sad}, \textit{angry} and \textit{fearful}) \cite{fen03} that rather simplified the goal of identifying emotion in music. However, the idea of classification is more adequate to other MIR tasks, such as genre, artist, or time period identification, since discrete response values properly describe the desired information.
%more recognizable ways of obtaining information from music, such as genre, artist or time period that are more easily identified with a discrete response value.
Not only emotions are more fluid but they are also hard to describe or objectively box, what could cause confusion in the human annotators needed for this task and, therefore, inconsistent data or very broad categories. %that fall short of what we seek to develop.

\section{Continuous approaches} \label{sec:cont}
Ultimately, discrete approaches were mostly dropped in favor of continuous ones that resonated more with the problem, and the aforementioned AV scale \cite{russell-av} became more widely used by the MER community, with its first use recorded in \cite{yan04}. In some cases \cite{lem05}, a 3D version of the scale is used, contemplating values for Valence, Arousal (in \cite{lem05} named Activity) and Interest or Dominance (exciting-boring).

%as pictured in Figure \ref{fig:AVDscale.PNG}. Each dot in the figure corresponds to an emotion in the adjacent table in a more or less arbitrary manner, as proposed in \cite{ver17}.

Although the addition of the Dominance axis expands the emotional spectre in one dimension, it also makes it a lot more complicated to understand and that can be troublesome, considering we need several humans to be capable of annotating this emotions with some level of precision.

Even though this version has show some good results in \cite{ver17}, the new axis has a subjectiveness akin to Valence, thus increasing the difficulty in the annotation process. Again, it is important to find balance between necessary complexity to make the results realistically useful and enough simplicity to keep the annotation less confusing as possible. 

    % \begin{figure}[H]
    % \centering
    % \includegraphics[scale=0.75]{Images/AVD.png}
    % \caption[The Arousal-Valence-Dominance Scale]{The Arousal-Valence-Dominance Scale with associating emotions (Figure 4 in \cite{ver17}).}
    % \label{fig:AVDscale.PNG}
    % \end{figure}
    % %https://www.springer.com/gp/rights-permissions/obtaining-permissions/882

Although many different datasets have been used in literature, the data used to relate the song to the response is frequently composed by the songs'  acoustic features \cite{Panda-dsc}, e.g. spectral centroid ( where the center of mass of the spectrum is located), range and Zero-crossing rate (instantaneous point at which there is no voltage present) of time signal among others \cite{Li11}. It is worth noting that although some of them, like loudness, fundamental frequency (the most prominent frequency present in a signal in a give time), skewness (3rd order moment), kurtosis (4th order moment) and others are known to be relevant in modeling Arousal, the correlation between acoustic features and Valence is less significant \cite{yan08}. This means that predicting both responses with this data have different difficulty levels.

These features can be extracted using softwares such as PsySound \cite{cabrebra-psysound} and Marsyas \cite{tzanetakis-cook-marsyas} which are open-source and capable of computing several of them rather quickly using Fast Fourier Transform \cite{din10}. Although easy and cheap to calculate, this data represent purely acoustical characteristics of the waveform and even the most relevant ones have relatively small correlation to the responses we expect. Testing the dataset employed here, the correlation reached around $0.3$ for Valence and $0.5$ for Arousal. This lead  us to employ a large number of features in order to interpret the desired variables, what increases the risk of overfitting and, in our database, many of those features had high correlation between others. To deal with those problems, we must build a model capable of selecting the most significant variables efficiently. 

Not only the scarcity of relevant features for Valence makes it generally more difficult to model than  Arousal. It is also worth noticing that the it has a much less precise interpretation, so an important challenge to have in mind is finding ways to improve Valence estimation. As mentioned in Chapter \ref{introduction} here we will try to tackle this problem by exploring the positive correlation between Arousal and Valence in a bivariate model, developed in Chapter \ref{MARD}.

Model-wise, several options were previously explored such as support vector machines \cite{li03}, $K$-nearest neighbors \cite{wie05} and even deep belief neural networks \cite{sch11} more recently, but one was particularly attention grabbing: a linear regression model built in \cite{yan08}.

\section{Our approach}\label{sec:our}
Our main goal is to find a model that is flexible, with continuous responses and easy to interpret, and then explore it from a statistical viewpoint. A linear regression seems a good choice, since it brings all this characteristics while also adding the benefit of allowing automatic feature selection. Although these attributes are not unique to a linear regression, it represents a good balance between simplicity and flexibility, as we open prospects of further investigation with more sophisticated models in the future. Regarding the pieces of information previously discussed we can set the roles of the AV values and acoustic features as response and explanatory variables, respectively.

\begin{comment}
Parting from the idea on \cite{yan08}, we replaced the manual feature selection process by implementing a LASSO penalization in the least-squares solution as in Equation \ref{eq:ls-lasso}, what also gave us some instant insight to what features were more relevant to each response.

\begin{equation} \label{eq:ls-lasso}
\begin{split}
 \underset{\beta}{\mathrm{argmin}} \sum_{i=1}^n \left(A_i - \beta_0 - \sum_{j=1}^p \beta_jx_{ij}\right)^2 + \lambda \sum_{j=1}^p |\beta_j|
\end{split}
\end{equation}
\end{comment}

Parting from the idea on \cite{yan08}, we implemented the LASSO to perform feature selection. It is easily merged to the least squares solution in a classic linear regression and it can also be implemented in a Bayesian linear regression as we will develop on Chapter \ref{bayes}. %, which means we will be able to compare results between these techniques. 
Expanding the possibilities of Bayesian models, we employ Automatic Relevance Determination (ARD) on Chapter \ref{vi_ard} (roughly explaining, the ARD is a penalization that excludes features based on how high their variance are and does not shrink all of them by the same amount as the LASSO). Up to this point, the estimation of Arousal and Valence was done separately, but finally, we developed a generalization of the ARD for multivariate responses in Chapter \ref{MARD}, the \textit{Multivariate Automatic Relevance Determination} (MARD).

We also used a different, broader dataset \cite{Ann17}, further explored in Chapter \ref{database}: while \cite{yan08} chose one containing only oriental pop pieces, we preferred other with many different genres and styles to increase the estimates robustness regarding the subject.

The $R^2$ metric is commonly used not only by \cite{yan08} but by many others in literature, but here we disclaim some concerns about it as well as propose a new one on Chapter \ref{bayes}: as the $R^2$ measures how much of the data variation is explained by the model it shows little insight whether we are correctly predicting the AV values of a song or not, so we need to find something more related to the subtlety of our problem.

\chapter{The Database}\label{database}

In order to test the proposed models, % and interpret results with some degree of confidence,
it was necessary to find a broad and reliable database, pertinent to the problem of Music Emotion Recognition. We will go through this database in detail in Section \ref{sec:deam}. It is also necessary to explore the chosen database and tailor it to our needs, so in Section \ref{sec:prep} we will look into the preprocessing we have done to the data. 

\section{The Database for Emotional Analysis of Music database} \label{sec:deam}

The Database for Emotional Analysis of Music (DEAM) database \cite{Ann17} was the one chosen by us, as it presents enough data to test our approaches and significant care in its construction, increasing its reliability. It it an aggregation of the datasets created for the ”Emotion in Music” task at MediaEval benchmarking campaign 2013-2015 \cite{alj15}. 
It carries $260$ acoustic, low-level features already calculated and $1802$ excerpts from royalty-free songs, annotated for Arousal and Valence by proper personnel, that is, several people were trained, tested and remunerated for this task. Although the features could have been easily calculated by us, the AV annotations would be much harder, time consuming and more expensive.

There was also a concern about stability of the data, meaning every piece was annotated by a minimum $10$ of people through $2013$ and $2014$, then annotated again in $2015$ by $5$ people, three of which were the most successful workers from the previous experiment. The results were verified to be very similar, even between annotators, thus giving even more credibility to this database. 
Even though the music genres were somewhat broad, it only adopted western music and without preliminary tests we cannot assure that a model trained in this database would interpret well genres not contemplated by it, but it is an aspect worth investigating in future research.  works in This direction was taken in \cite{cou14}, by transfer learning emotion across music and speech in the AV domain. 

Each piece in the $2013$ and $2014$ subsets was trimmed to $45$ seconds, while the $2015$ subset was annotated in full. They were all processed to have the same sampling frequency, i.e., 44,100 Hz and annotated throughout its length at a $2$ Hz rate. This means that every feature and AV value was measured two times per second throughout each song. Even though this greatly summarizes the information given (that would be 44100 per second otherwise) but it would still be a lot to process and would require a time sensitive model, which is not of our interest.

\section{Preprocessing}\label{sec:prep}

Some preprocessing was necessary in order to work with DEAM, as we are aiming for a global estimate for A and V, rather than one for every few seconds. A model that incorporated the variable time can be considered, but it will not be explored in this work.
We began by excluding the initial $10\%$ samples of every piece, since slow intros or fading effects could give erroneous information about the bulk of the song. Since some of the songs were snippets and others were full length, there were long strings of $0$'s recorded at the end of shorter pieces, so we also trimmed the endings to exclude them.

\begin{figure}[H]
    \centering
    \includegraphics[width=0.5\textwidth]{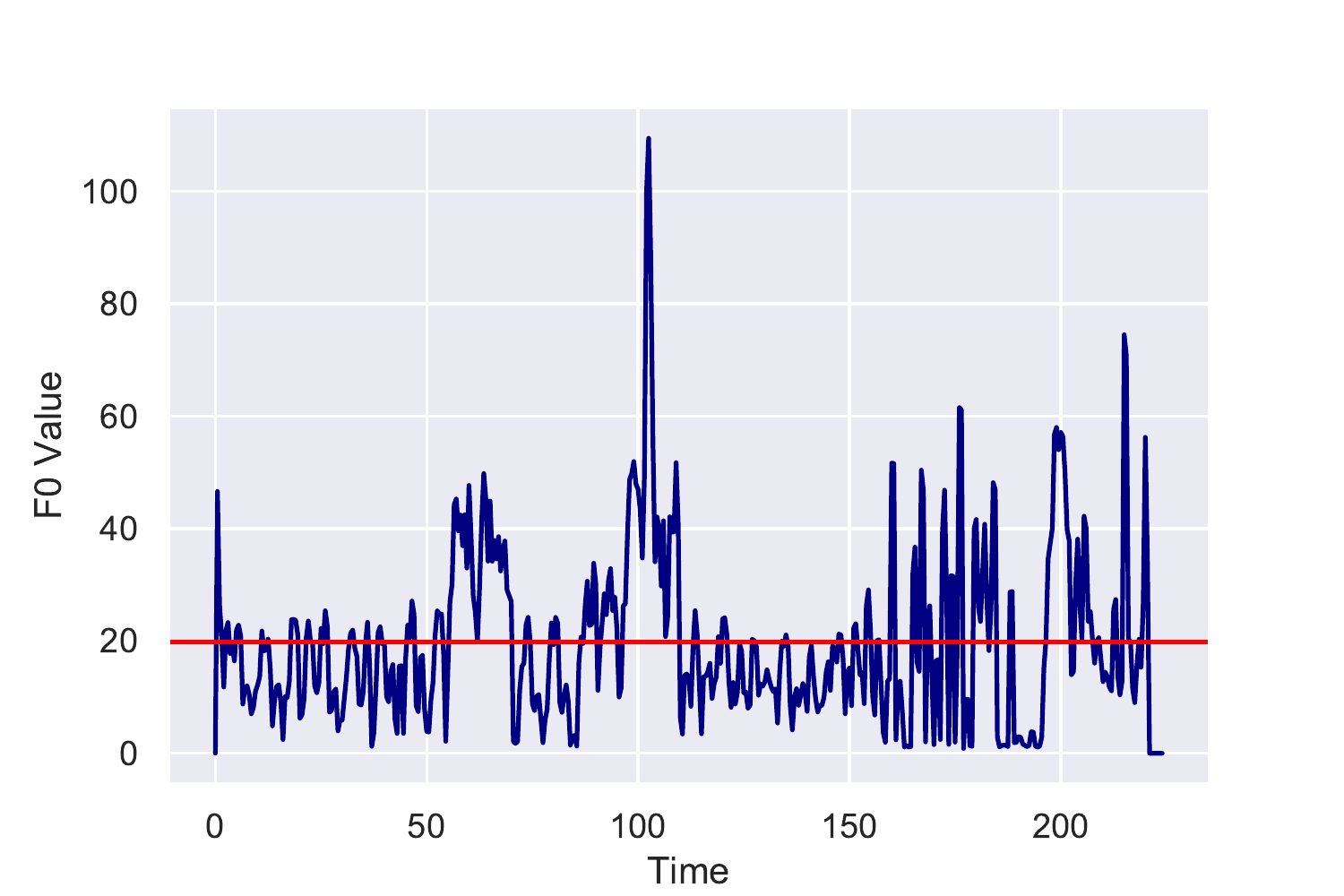}
    \caption[Example of fundamental frequency evolution throughout time for a single musical piece.]{Example of fundamental frequency evolution throughout time for a single musical piece.}
    \label{fig:f0_prep}
\end{figure}

Figure \ref{fig:f0_prep} shows an example of the evolution of one particular feature, the fundamental frequency for a single song in blue. In order to reduce this information to a single feature value, we calculate every feature's average for each song.

% we can see how the features and AV values were measured and recorded every few second throughout the song. This was solved by taking an average of every feature and AV-values for each song and stored so for every song in the dataset there was a single values for A, for V and for each of the $260$ features. We can also see in the Figure how the information was summarized by the red x for a single feature (the fundamental frequency) and for a few songs in the dataset. 

\begin{figure}[H]
    \centering
    \includegraphics[width=0.5\textwidth]{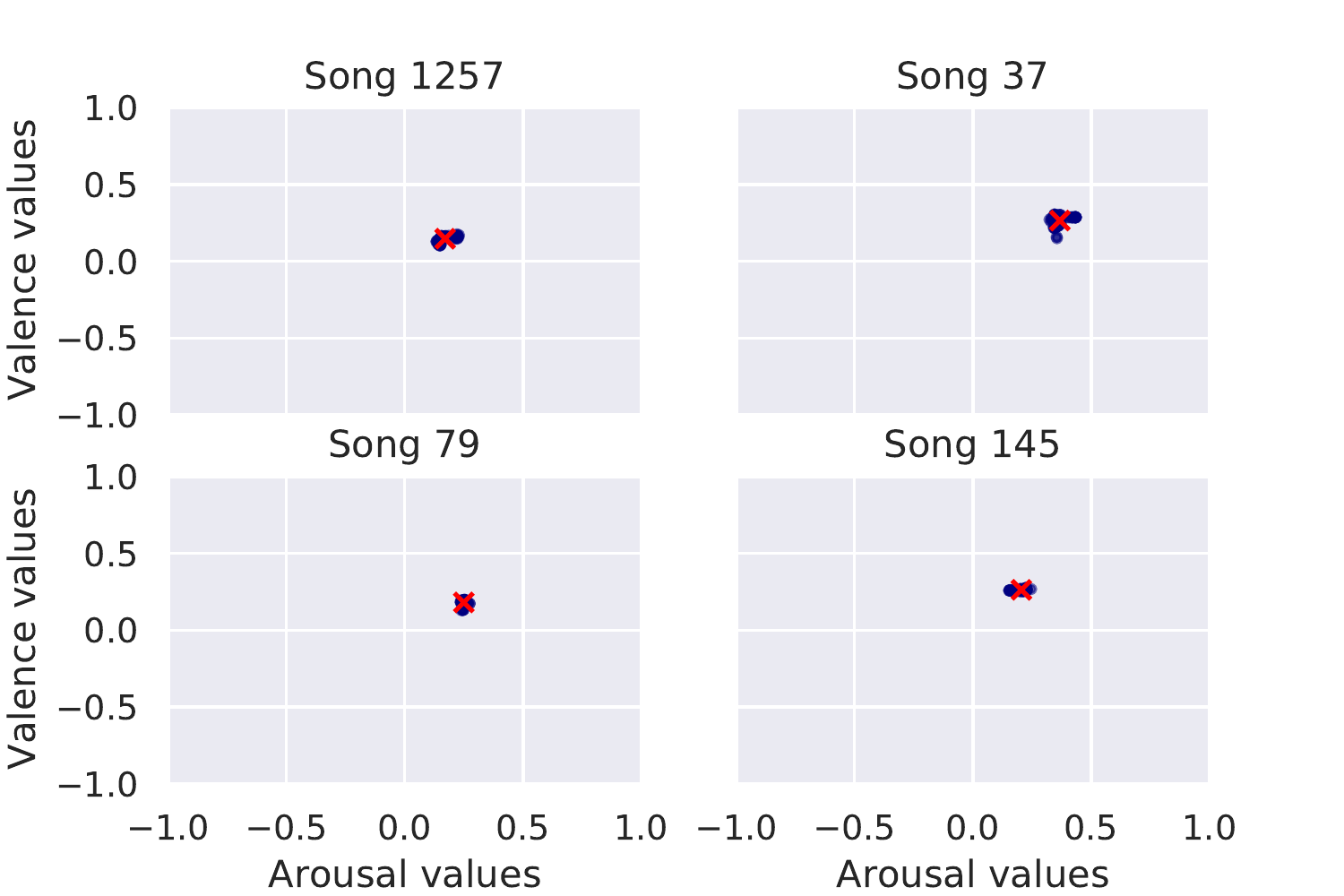}
    \caption[Example of annotated AV values for four songs and their reduction.]{Example of annotated AV values for four songs and their reduction. Blue dots represent annotations and red crosses the summarized value}
 \label{fig:AV_prep}
\end{figure}

Figure \ref{fig:AV_prep} on the other hand, shows the annotated AV values throughout time for four different songs by a single annotator. We can note that the values are well concentrated, something verified to be true throughout the database. This information was also reduced with the computation of the sample mean, here represented by the red crosses.

\begin{figure}[H]
    \centering
        \includegraphics[width=0.3\textwidth]{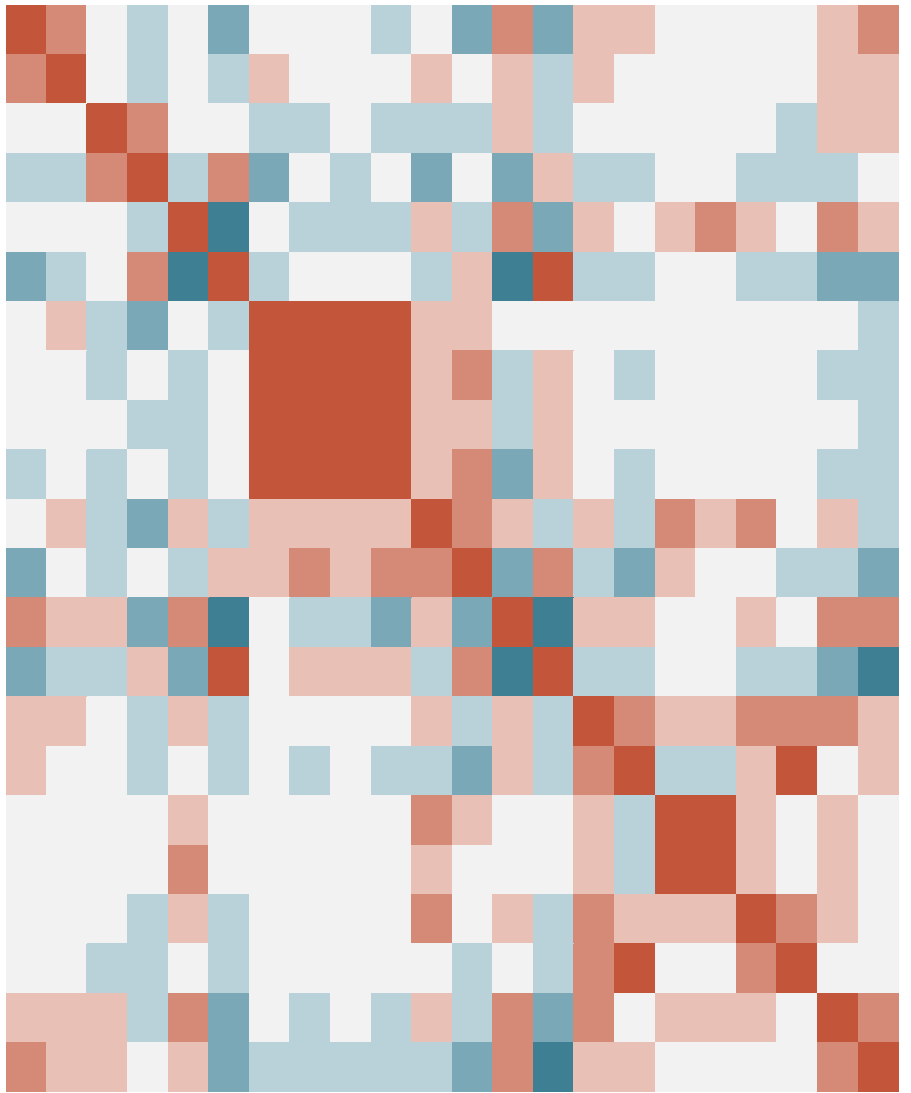}
        \caption{Excerpt of the correlation matrix between the 260 features, after preprocessing. Red values are close to $1$ and blue values are close to $-1$.}
        \label{fig:corr}
\end{figure}

We also noticed very high correlations between many of the features, as we can see in an excerpt of the $260 \times 260$ feature correlation matrix (after the aforementioned preprocessing) on Figure \ref{fig:corr}. Ideally, all but the main diagonal should be white, meaning zero correlation, but there are a significant amount of red (highly positively correlated) and blue (highly negatively correlated) pair of variables. Although shown in only an excerpt, relevant correlation values can be found in the whole correlation matrix between most of the features.

% the scales of red and blue represent respectively high and low correlation values. The most extreme values reach $0.8$ and $-0.8$, what can be damaging to he regression model, since this pattern is presented on the whole matrix, not just the excerpt. We chose to apply Principal Component Analysis (PCA) was applied to the features in order to mitigate that problem.

To mitigate the problem of high correlation, Principal Component Analysis (PCA) was applied to the features, implying that the linear regression models implemented here will have linear combinations of features as explanatory variables, rather than the features themselves.

The preprocessed data was then allocated in matrices used to train and test models. The response variables were stored in vectors $\mathbf{A}, \mathbf{V} \in \mathbb{R}^n, n=1802$. The explanatory variables (features after PCA) were placed in a matrix $\bfx \in \mathbb{R}^{n \times p}, n=1802, p=260$. 

 The database was randomly split in a training set containing $1262$ songs and a test set with $540$ songs, that is, $\frac{2}{3}$ and $\frac{1}{3}$ of the database, respectively.

\chapter{Music Emotion Recognition Via the Bayesian LASSO}\label{bayes}
    
\begin{comment}

-variational inference
-motivation for using VI 
-how CAVI works
-show the math

\end{comment}

In this Chapter we will explore how we can incorporate Bayesian statistics into our problem, what possibilities it adds to the modeling process and what can be done to push previous limitations. In Section \ref{sec:intro} we will lay out some basic insights on Bayesian modeling when applied to MER and in Section \ref{sec:lasso} we will work on a way to perform automatic feature selection, looking in how it fits in our scenario. The implementation of the model will be discussed in Section \ref{sec:bayes_lasso}, finally laying out the results we obtained in comparison with the ones achieved in the classical approach in Section \ref{sec:results} and wrapping up final thoughts in Section \ref{sec:final_bayes}.

\section{Why Bayesian Inference?}\label{sec:intro}
Keeping in mind the problem at hand, it is reasonable to consider changing from a classic approach to a Bayesian one. Emotions are naturally fluid and uncertain, so it makes sense to consider our response variables as random variables and instead of point estimates for the AV values we would have posterior probability distributions from which we can calculate credibility intervals. These intervals can be used not only for model evaluation but also for building a recommendation system, as we will see later on. A Bayesian approach also presents an advantage when smaller databases are considered and, particularly, the penalized model implemented here also takes that into consideration. In the scenario of MER, the annotation process can be slow and expensive so being able to work with smaller databases is not a disposable quality.

\section{Linear regression with the LASSO}  	\label{sec:lasso}
For each song $i$ in the dataset, let $A_i$ and $V_i$, denote the respective Arousal and Valence values, along with a vector $\mbbx_i \in \mathbb{R}^p$ containing the respective song's features \footnote{Let us remember that we are no longer using the actual features, but a linear combination of them due to the PCA. For simplicity we will continue to call this linear combination of features, ``features''}. Even though only numerical features are being used, this method can be adapted to include categorical ones as well \cite{mont13}.

We will model Arousal and Valence by two independent linear regressions, as stated in Equation \ref{eq:model}:
\begin{equation} \label{eq:model}
\begin{split}
A_i = \beta_0^a + \mathbf{x}_i^T\bfbeta^a + \varepsilon_i^a, ~~ i = 1, \dots, n; \\
V_i = \beta_0^v + \mathbf{x}_i^T\bfbeta^v + \varepsilon_i^v, ~~ i = 1, \dots, n,
\end{split}
\end{equation}
where $\varepsilon_i^a \sim \text{N}(0, \sigma_a^2)$ and $\varepsilon_i^v \sim \text{N}(0, \sigma_v^2)$ are the measurement errors, assumed independent for $i = 1, \dots, n$. Vectors $\bfbeta^a = [\beta_1^a, \dots , \beta_j^a, \dots , \beta_p^a], \bfbeta^v  = [\beta_1^v, \dots , \beta_j^v, \dots , \beta_p^v] \in \mathbb{R}^p$ and scalars $\beta_0^a, \beta_0^v$ are the regression coefficients, to be estimated from the observed data.

The LASSO (Least Absolute Shrinkage and Selection Operator), also known as the L$1$ penalty, was originally proposed in the geophysics literature in the 1980's \cite{san86}, and further popularized in Statistics in \cite{Tib96}. Roughly explaining, it performs feature selection by shrinking the regression coefficients and discarding the sufficiently small ones, excluding therefore the most irrelevant features. In the classical scenario, it can be incorporated in the least-squares estimator as:
\begin{equation} \label{eq:lasso-classic}
\begin{split}
\widehat{\bfbeta^a} &=  \underset{\bfbeta^a \in \mathbb{R}^p}{\mathrm{argmin}} \sum_{i=1}^n \left(A_i - \beta^a_0 - \sum_{j=1}^p \bfbeta^a x_{ij}\right)^2     + \lambda_a \sum_{j=1}^p |\beta^a_j|; \\
\widehat{\bfbeta^v} &= \underset{\bfbeta^v \in \mathbb{R}^p}{\mathrm{argmin}} \sum_{i=1}^n \left(V_i - \beta^v_0 - \sum_{j=1}^p \beta_j^v x_{ij}\right)^2     + \lambda_v \sum_{j=1}^p |\beta^v_j|,
\end{split}
\end{equation}
where the parameters $\lambda_a$ and $\lambda_v$ can be tuned to increase or decrease the penalization, and can be chosen, for example, via cross-validation \cite{tib13}. Note that, since the intercept is not related to any feature, it is not included in the penalization term.

\section{Bayesian LASSO} \label{sec:bayes_lasso}

The least-squares solution in Equation \ref{eq:lasso-classic} can easily be interpreted as the maximum a posteriori of a Bayesian model on which the prior over each component of $\bfbeta^a$ and $\bfbeta^v$ are independent Laplace distributions with fixed parameter $\lambda_a$ and $\lambda_v$, respectively \cite{par08}. This derivation is presented in Appendix \ref{appendix:A}.

Since Arousal and Valence are modeled analogously there is no need to present the model twice, and in order to overcome ambiguities, denote by $y_i$ the observed value of Arousal or Valence for song $i$, and the vector containing all of these observations by $\mathbf{y}$. We will also omit the superscript ``a" and ``v" on parameters and hyperparameters.

Equation \ref{eq:bayes-1} presents an extension of this model, where a prior distribution is also assigned for $\lambda$, here a Gamma distribution parametrized by shape and rate, for conjugation purposes. The conditional distribution of $\mathbf{y}$ reflects the linear regression model in Equation \ref{eq:model}. The Inverse-Gamma prior attributed to $\sigma^2$, parametrized by shape and scale, is a common choice for the variance of a Normal distribution as it conjugates with the likelihood. The intercept $\beta_0$ does not carry relevant prior information to the model, so we assign a flat improper prior for it. 

% a Bayesian regression in Equation \ref{eq:bayes-1}, with a Laplace prion distribution on the coefficients acting as penalization. Elaborations on this are presented on Appendix \ref{appendix:A}. 

% The hyperparameter $\lambda$ can be attributed

\begin{equation}\label{eq:bayes-1}
\begin{split}
\mathbf{y}|\bfbeta, \sigma^2 &\sim \text{N}(\beta_0 + \mathbf{x}_i^T\bfbeta, \sigma^2),~i = 1, \dots n \\
\beta_0 &\sim \text{Flat} \\
\sigma^2 &\sim \text{IG}(a, b) \\
\beta_j|\lambda &\sim \prod_{j = 1}^p\text{Lap}(\lambda) \\
\lambda &\sim \Gamma(c, d).
\end{split}
\end{equation}

With this we can induce sparsity on the regression coefficients and we are able to choose parameters for the prior distributions of $\sigma^2$ and $\lambda$ such that they are non informative. 
The model in Equation \ref{eq:bayes-1} is the Bayesian LASSO as proposed by \cite{par08}, but it will not result in closed form conditional posteriors, complicating the implementation of a Gibbs sampler. To overcome that, we will write the Laplace distribution as a mixture of Normal and Exponential \cite{bab10}, at the expense of increasing the number of parameters, being the full model rewritten as:

\begin{equation} \label{eq:bayes-2}
\begin{split}
\mathbf{y}|\bfbeta, \sigma^2 &\sim \text{N}(\beta_0 + \mathbf{x}_i^T\bfbeta, \sigma^2),~i = 1, \dots n \\
\beta_0 &\sim \text{Flat} \\
\sigma^2 &\sim \text{IG}(a, b) \\
\beta_j | \gamma_j &\sim \text{N}(0, \gamma_j),~j = 1, \dots, p \\
\gamma_j | \lambda &\sim \mathrm{Exp}(\lambda/2),~j = 1, \dots, p \\
\lambda &\sim \Gamma(c, d).
\end{split}
\end{equation}

Since the calculations for this Gibbs sampler are more commonplace, they can be found in Appendix \ref{Appendix:B}. The result is presented below, where $\mathbf{z}_{-k}$ denotes the vector $\mathbf{z}$ without the $k$-th entry:

\begin{equation} \label{eq:bayes-2-posteriors}
\begin{split}
\beta_0|\cdot  &\sim    \mathrm{N}\left(\beta_0   \left| \frac{1}{n}\sum_{i=1}^n \right. (y_i - \mathbf{x}_i^T \bfbeta), \frac{\sigma^2}{n}\right) \\
\sigma^2| \cdot   &\sim    \mathrm{IG}\Bigg(\sigma^2   \left| a + \frac{n}{2}, b + \frac{1}{2} \sum_{i=1}^n \right.  (y_i   -   \beta_0   -   \mathbf{x}_i^T \bfbeta)^2 \Bigg) \\
\lambda|\cdot   &\sim   \mathrm{G} \left(\lambda   \left| c + p, d + \frac{1}{2} \sum_{j=1}^p \right.
 \gamma_j \right) \\
\beta_j|\cdot   &\sim    \mathrm{N}\Bigg(\beta_j   \left| \mu_{\beta_j}, \sigma_{\beta_j}^2 \Bigg)\right., \\
& ~~~~ \sigma_{\beta_j}^2 = \left[\frac{1}{\gamma_j} + \frac{1}{\sigma^2}\sum_{i=1}^n (x_{i,j})^2  \right]^{-1} \\
& ~~~~ \mu_{\beta_j} = \frac{\sigma_{\beta_j}^2}{\sigma^2} \sum_{i=1}^n x_{i,j}(y_i - \beta_0  - \mathbf{x}_{i,-j}^T \bfbeta_{(-j)}) \\
\gamma_j|\cdot   &\sim    \mathrm{GIG}(\gamma_j | \lambda, \beta_j^2, 1/2),  \\
\end{split}
\end{equation}
where GIG denotes the Generalized Inverse Gaussian distribution \cite{jor81} and $x_{ij}$ is the $j^{th}$ component of $x_i$. Even though is not as common as the other distributions, we can sample from it by employing the algorithms proposed in \cite{dev14, sta17}, translated to Python by us.
 
\section{Implementation and results} \label{sec:results}

All algorithms were implemented in Python language running on Google Colab, a free environment that runs Jupyter notebooks in the cloud.

For comparison reasons, firstly we implemented a classical linear regression model with LASSO penalization as in Equation \ref{eq:lasso-classic} to evaluate the Bayesian model's contributions. We were able to obtain similar results to \cite{yan08} in the training scores, although the test ones were remarkably lesser.

For the Bayesian approach we implemented a basic Gibbs sampler from scratch. We ran $10,\!000$ iterations of it, with the whole process lasting $45$ minutes for each model, and considered the first $1,\!000$ as the burn-in time of the chain, that is, the number of iterations it took to achieve convergence. In order to compare the results with the closest approach in the literature and our prior inspiration \cite{yan08}, we first calculated the train and test $R^2$ obtaining $0.60$ and $0.58$ for Arousal and $0.45$ and $0.25$ for Valence, not surpassing results in other approaches \cite{Panda-dsc}. As discussed before, the $R^2$ metric can lead to misleading conclusions and is not quite adequate to measure how well we are capturing a piece's emotion. The $R^2$ estimates the percentage of data variation explained by the model, which is a general aim for several problems but since we have the very specific task of estimating Arousal and Valence, the results shown by the $R^2$ have little consequence as commented in the end of Chapter \ref{lit_review}. Since a Bayesian approach is being used, we are able to compute the predictive distributions for Arousal and Valence for each song in the test set and compute the credible intervals at $95\%$ for the training set carrying $66\%$ of our samples. In the test set, we found that $201$ of the intervals contained the measured value for Arousal and $185$ for Valence, as laid out in Tables \ref{table:results1A} and \ref{table:results1V}. Since this metric was not employed in other MER works there is no frame of reference for comparison with other works. Still, considering we have been able to encapsulate less than half of true AV values within the intervals, we can certainly look into reasons why the results have not been better and ways to improve them.

\begin{table}[H]
\begin{center}
\begin{tabular}{||c c c||} 
\hline
\textbf{Arousal} & Classic & Bayesian \\ [0.5ex] 
 \hline\hline
Training $R^2$ &  $0.22$  & $0.60$  \\
\hline
Test $R^2$ & $0.15$ & $0.58$  \\
\hline
Credible intervals containing measured value & -- & $201$ \\[1ex]
\hline
 \hline
\end{tabular}
\end{center}
\caption{Comparison between classic and Bayesian LASSO for Arousal.}
\label{table:results1A}
\end{table}

\begin{table}[H]
\begin{center}
\begin{tabular}{||c c c||} 
\hline
\textbf{Valence}  & Classic & Bayesian \\ [0.5ex] 
 \hline\hline
Training $R^2$ &  $0.12$  & $0.45$ \\
\hline
Test $R^2$ & $0.06$ & $0.25$ \\
\hline
Credible intervals containing measured value & -- & $185$ \\[1ex]
\hline
 \hline
\end{tabular}
\end{center}
\caption{Comparison between classic and Bayesian LASSO for Valence.}
\label{table:results1V}
\end{table}

\begin{figure}[ht]
    \centering
        \subfloat{\includegraphics[width=0.45\textwidth]{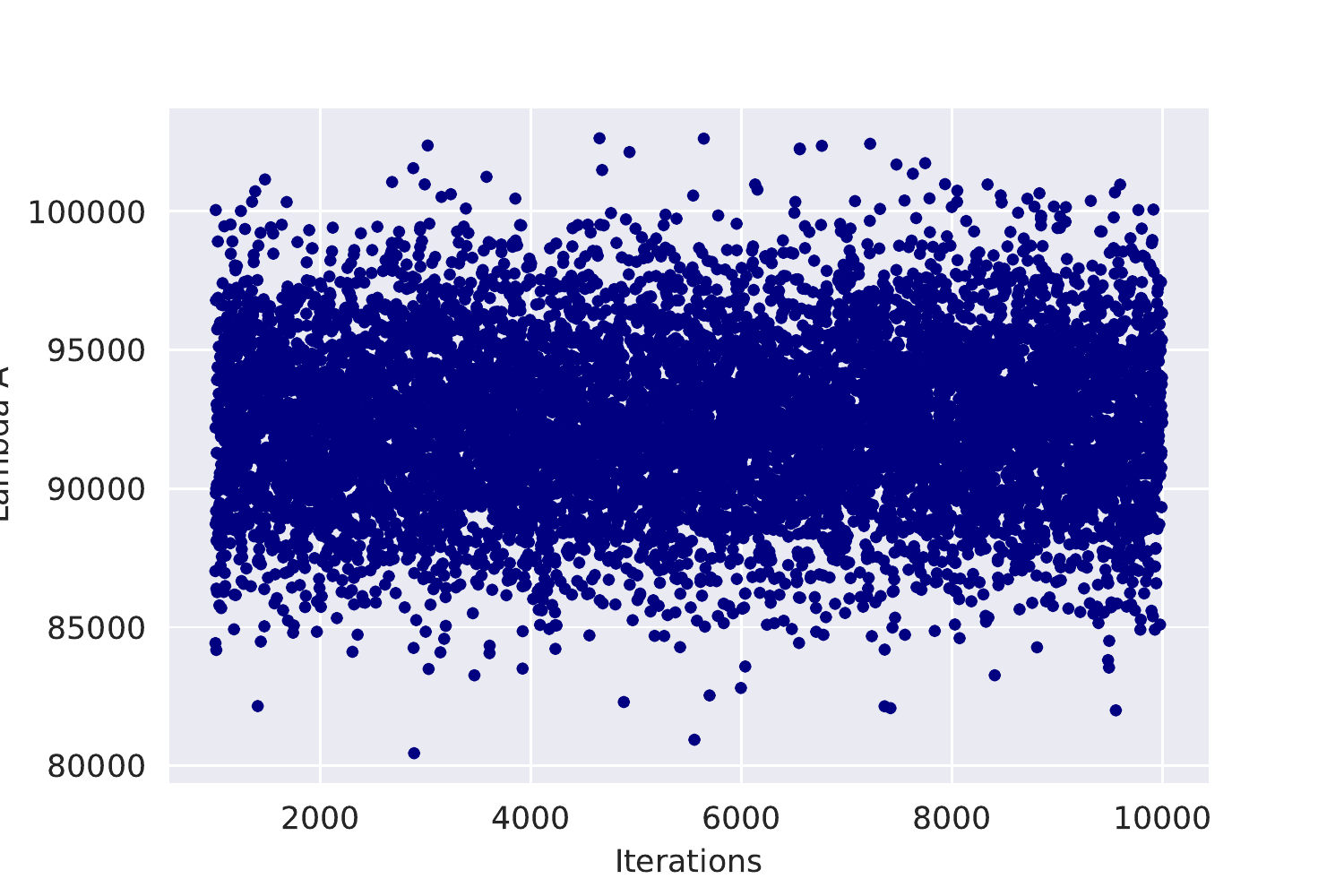}}
        \subfloat{\includegraphics[width=0.45\textwidth]{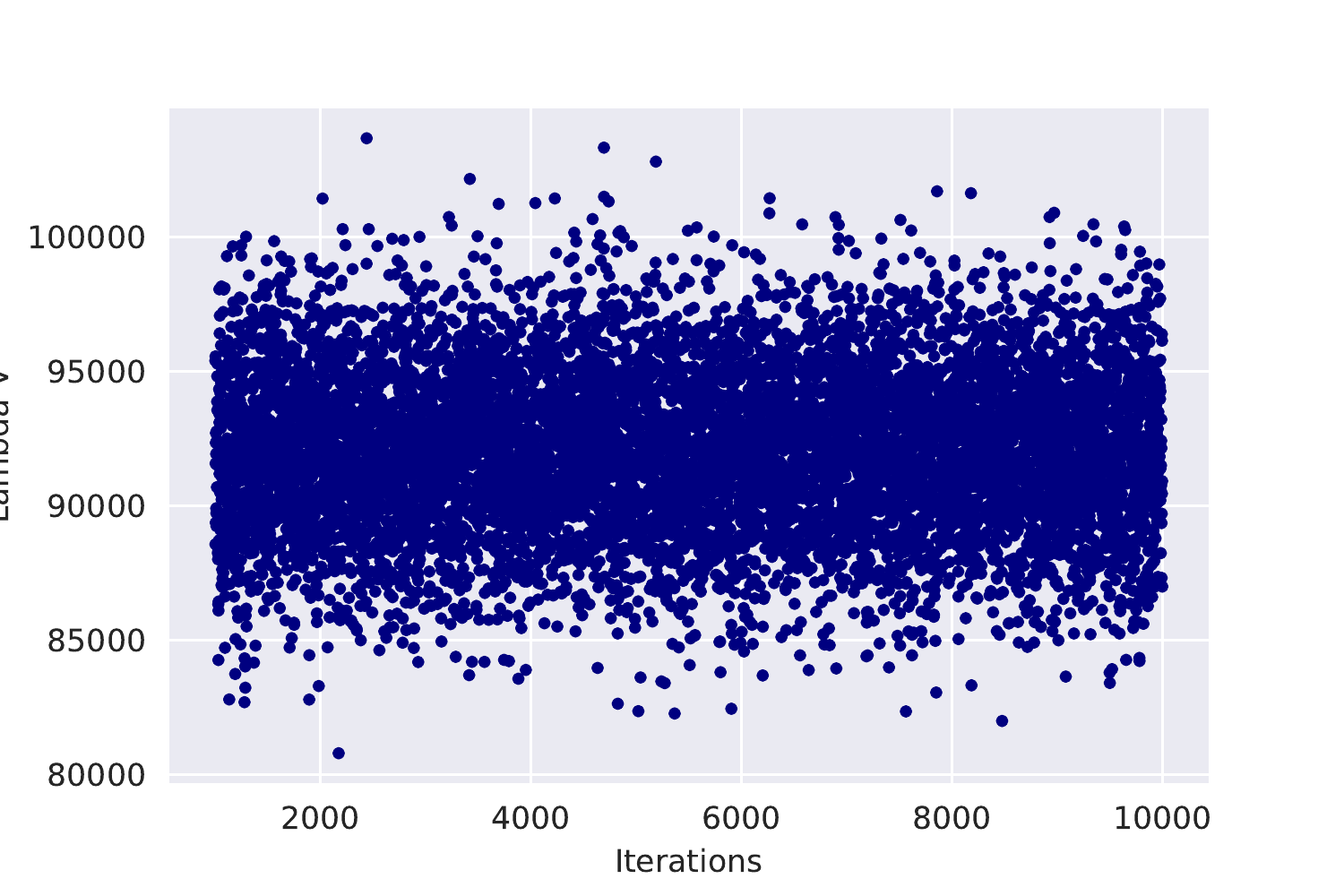}}
        \caption{Sampled values for $\lambda_a$ and $\lambda_v$ in the Bayesian LASSO.}
        \label{fig:lambda_conv}
\end{figure}

Now, let us study the role the penalization parameters $\lambda_a$ and $\lambda_v$ took in the modeling. In Figure \ref{fig:lambda_conv} we display the values encountered for $\lambda_a$ and $\lambda_v$ after discarding the burn-in. There we can see how $\lambda_a$ and $\lambda_v$ converge to high values, indicating the important role regularization takes place in this problem. 

Also note that the mean of Arousal's $\lambda$ seems higher, probably due to the fact that less information about Valence is gained from the features and the model tries to preserve as much information as possible. This hypothesis cannot be confirmed by visual inspection of Figure \ref{fig:lambda_conv} alone, so in order to investigate the issue further we would need to calculate each posterior mean, or provide a more accurate visual aid, such as their estimated densities. 

\begin{figure}[ht]
    \centering
        \subfloat{\includegraphics[width=0.5\textwidth]{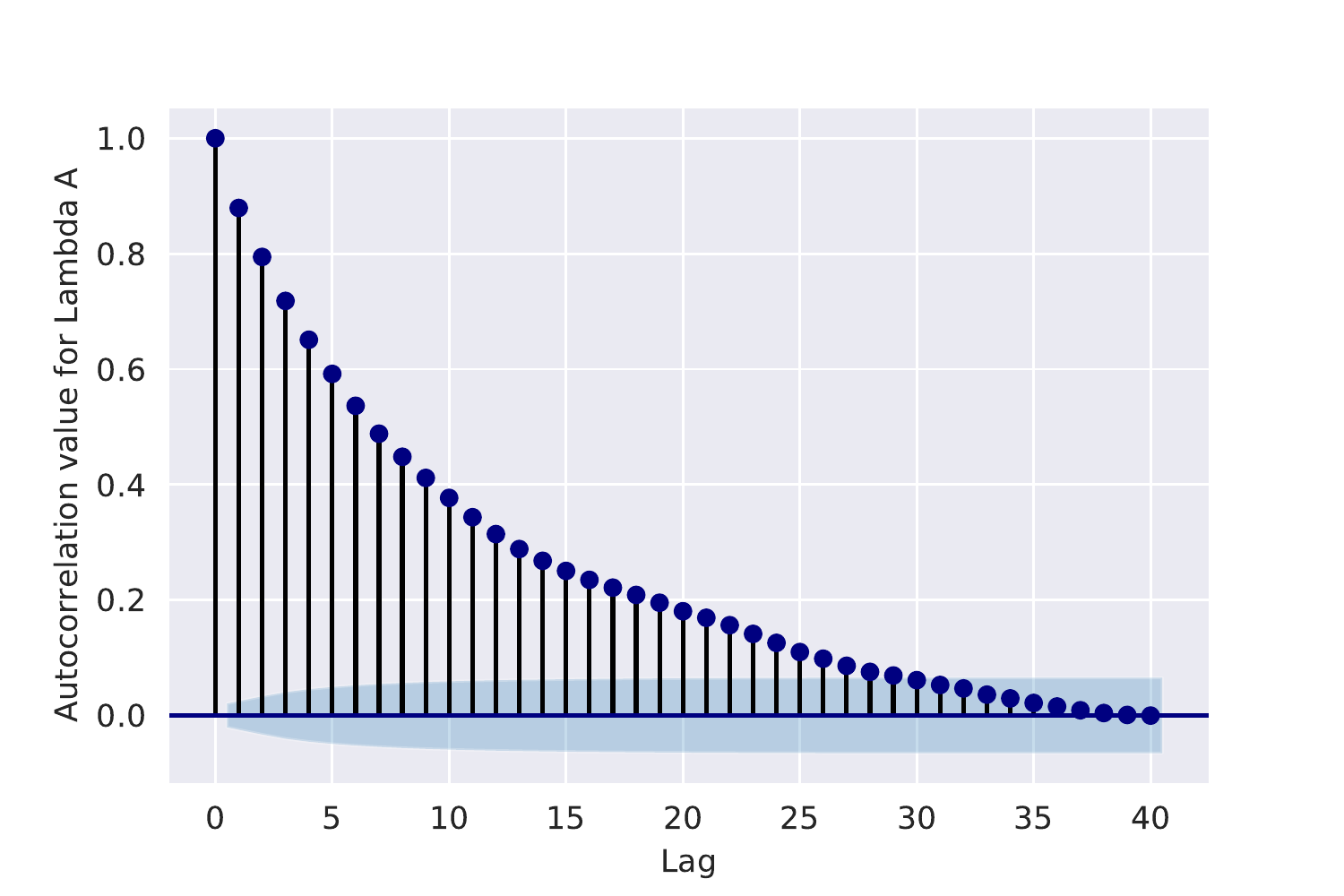}}
        \subfloat{\includegraphics[width=0.5\textwidth]{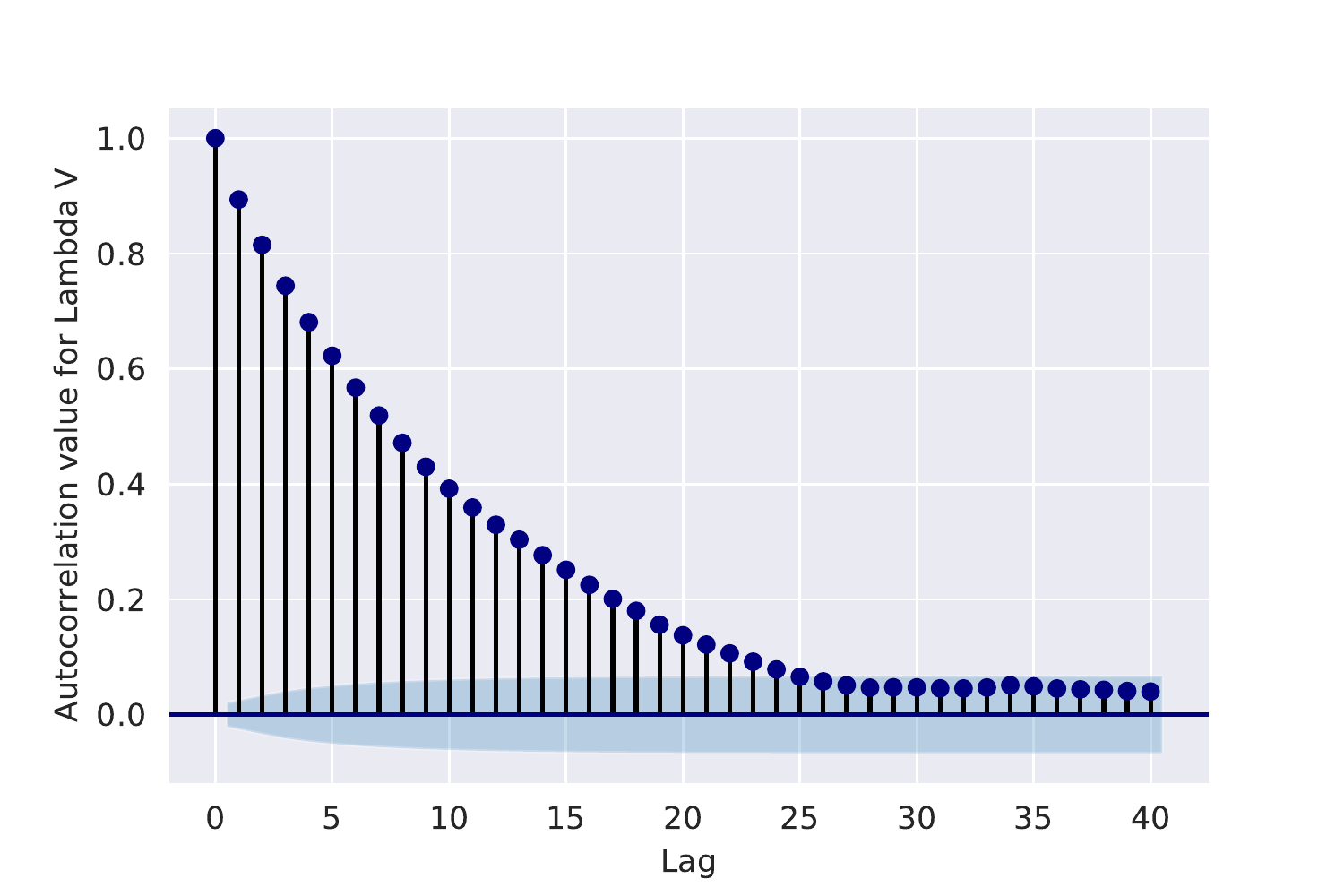}}
        \caption{Autocorrelation function of the sampled values for $\lambda_a$ and $\lambda_v$ in the Bayesian LASSO.}
        \label{fig:acf_lambda}
\end{figure}

Firstly note that Figure \ref{fig:lambda_conv} still spread out quite a bit, indicating the sampled values of $\lambda_a$ and $\lambda_v$ are substantially correlated, as confirmed by the autocorrelation functions in Figure \ref{fig:acf_lambda}. Before we make any assessments, only one in every $30$ samples should be considered for $\lambda_a$ and $25$ for $\lambda_v$, implying we would significantly reduce our amount of samples.% to achieve independence among.

\begin{figure}[ht]
    \centering
        \subfloat{\includegraphics[width=0.5\textwidth]{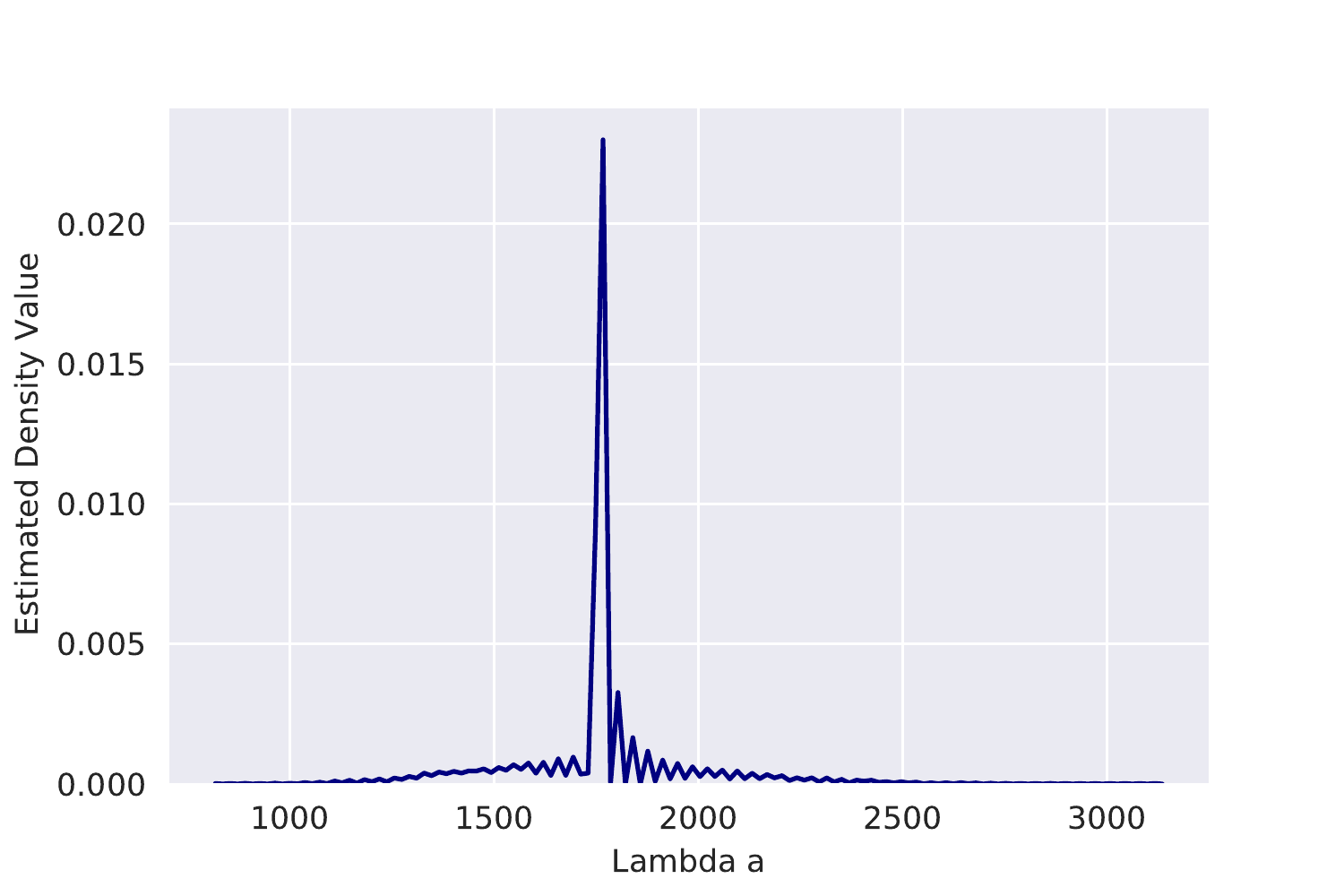}}
        \subfloat{\includegraphics[width=0.5\textwidth]{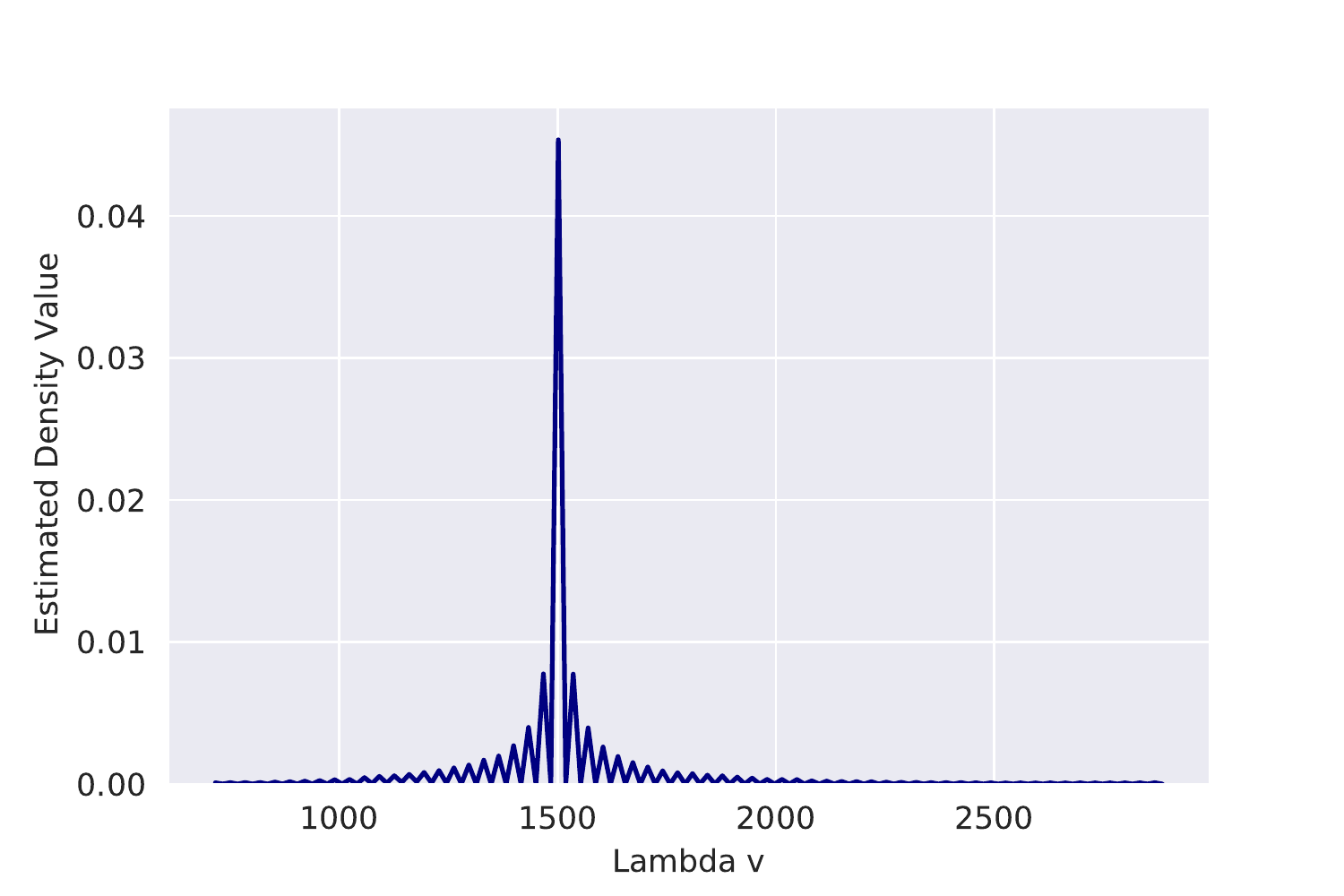}}
        \caption{Density estimation of $\lambda_a$ and $\lambda_v$ from samples for Arousal and Valence for the Bayesian LASSO.}
        \label{fig:lambda_dist}
\end{figure}

After discarding the appropriate amount of samples we are able to better estimate the posterior mean from both chains, returning $1753.3$ for $\lambda_a$ and $1501.6$ for $\lambda_v$, confirming our previous hypothesis. In  Figure \ref{fig:lambda_dist} we can see an estimation of the density of their respective distributions and how concentrated they are around their respective posterior means.

\begin{figure}[ht]
    \centering
        \includegraphics[width=0.7\textwidth]{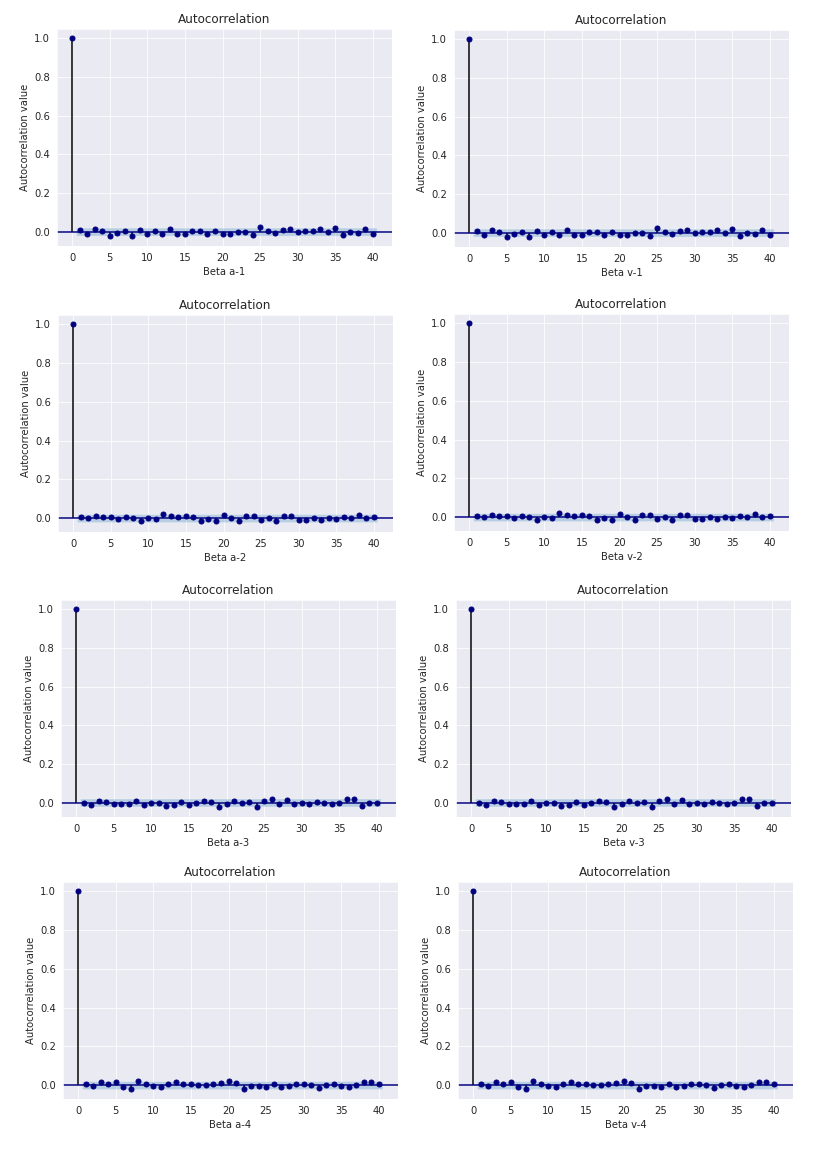}
        \caption{Autocorrelation function of the sampled values for first few $\beta_a$'s and $\beta_v$'s in the Bayesian LASSO.}
        \label{fig:beta_cov}
\end{figure}

Since we have auto-correlated samples for both regularization parameters, it is worth analysing the autocorrelation for a few $\beta$'s, let us say the first five for Arousal and Valence, since the $\beta$'s are directed linked with the respective regularization parameters. As we can see on Figure \ref{fig:beta_cov}, the correlation between $\lambda$'s did not contaminate the $\beta$'s.

It is also worth analysing the convergence of $\sigma^2_a$ and $\sigma^2_v$ in Figure \ref{fig:sigma_conv} as evidence of the convergence for the other variables' chains:

\begin{figure}[ht]
    \centering
        \subfloat{\includegraphics[width=0.5\textwidth]{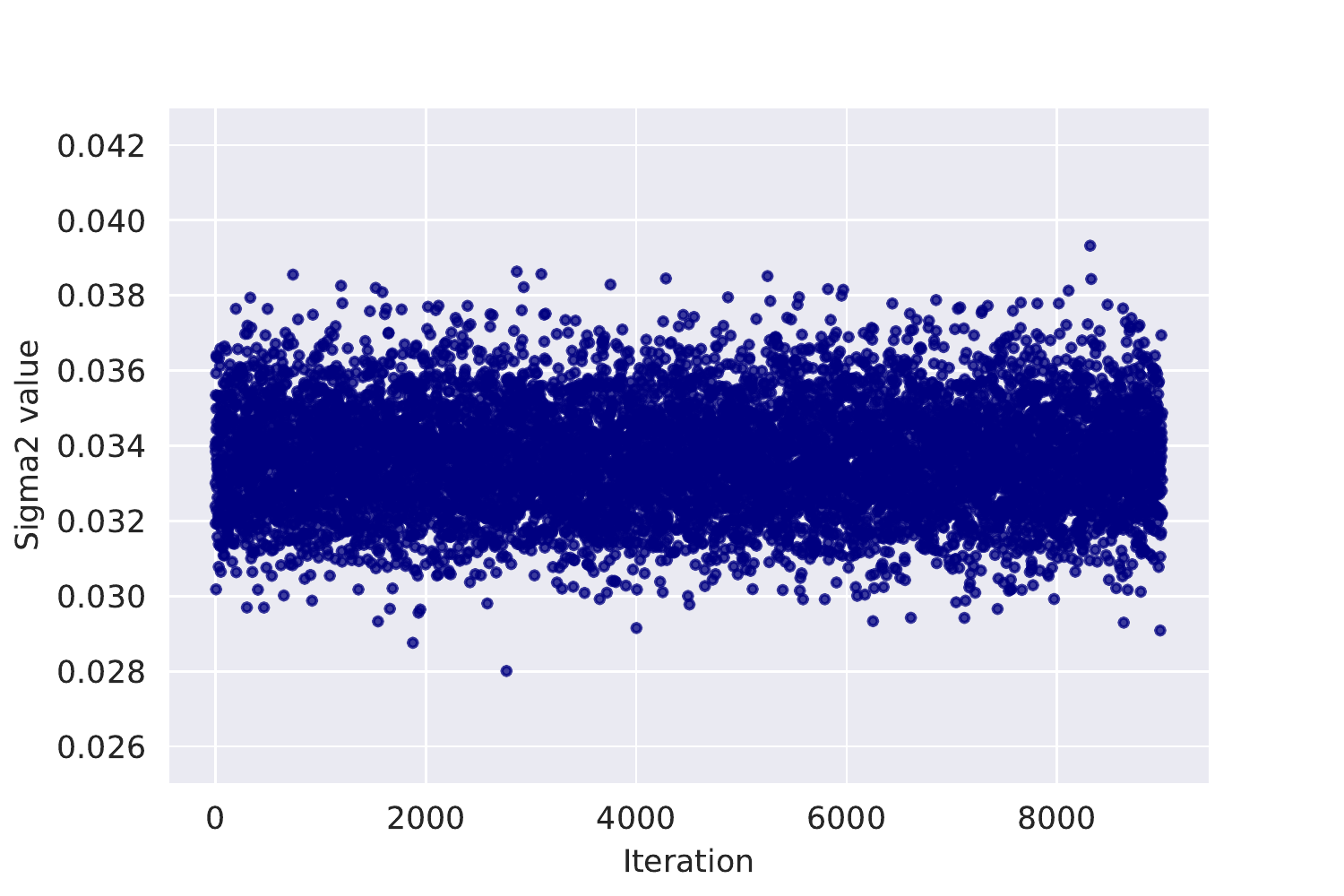}}
        \subfloat{\includegraphics[width=0.5\textwidth]{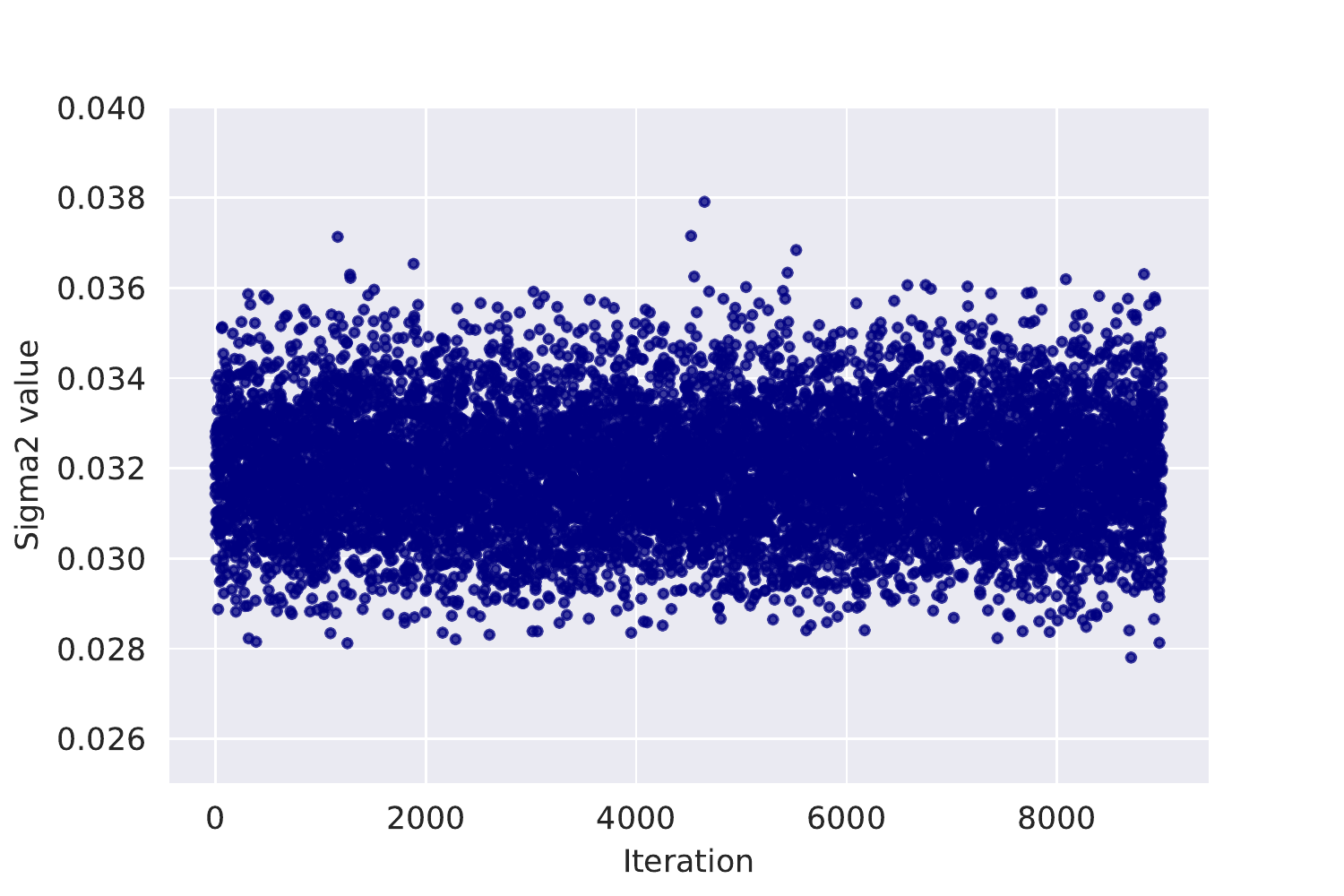}}
        \caption{Convergence of $\sigma_a^2$ and $\sigma_v^2$ samples in the Bayesian LASSO.}
        \label{fig:sigma_conv}
\end{figure}

We can also look at the autocorrelation plot in Figure \ref{fig:sigma_auto}, as an indicator that our samples to be uncorrelated:

\begin{figure}[ht]
    \centering
        \subfloat{\includegraphics[width=0.5\textwidth]{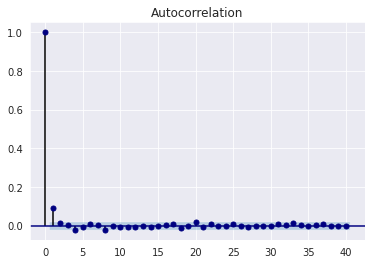}}
        \subfloat{\includegraphics[width=0.5\textwidth]{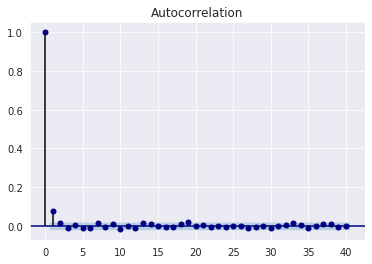}}
        \caption{Autocorrelation function of the sampled values for $\sigma_a^2$ and $\sigma_v^2$ in the Bayesian LASSO.}
        \label{fig:sigma_auto}
\end{figure}

\begin{figure}[H]
    \centering
        \subfloat{\includegraphics[width=0.5\textwidth]{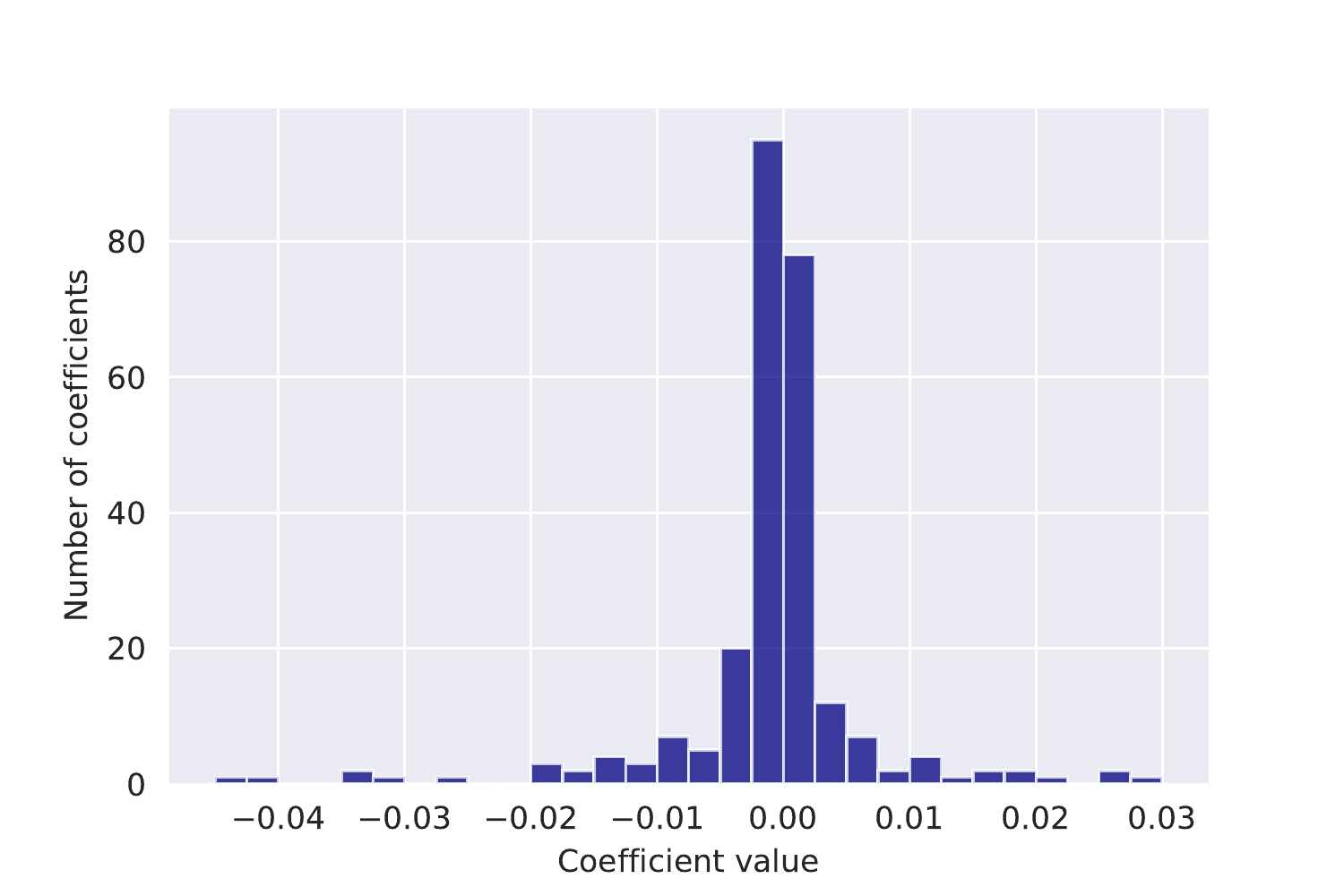}}
        \subfloat{\includegraphics[width=0.5\textwidth]{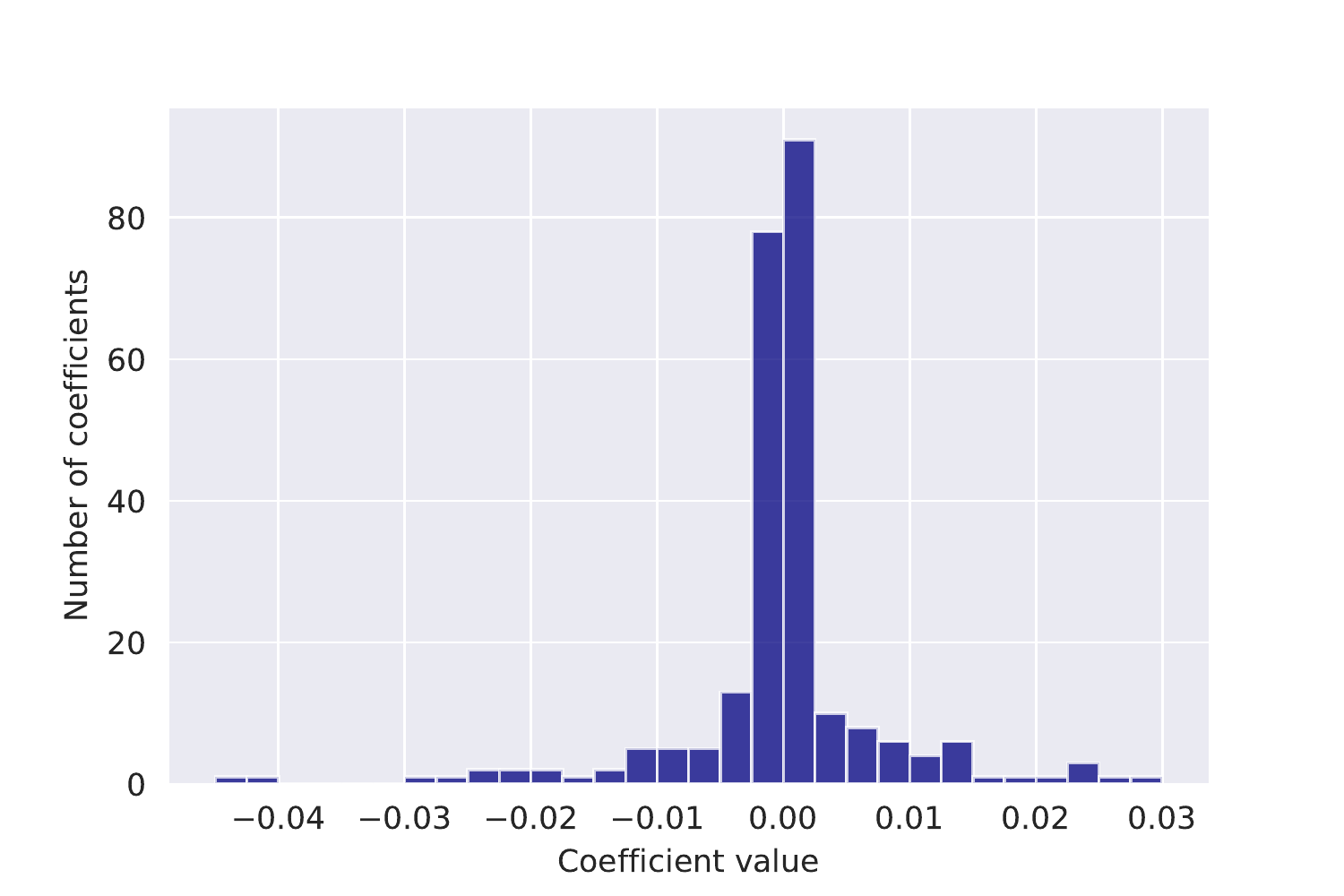}}
        \caption{Coefficient values for $\beta_a$ and $\beta_v$ in the Gibbs model.}
        \label{fig:beta}
        
\end{figure}

Now, let us analyse how LASSO regularization affects the parameters, that is, how the components of $\bfbeta_a$ and $\bfbeta_v$ are shrank towards zero in the regularized model. In Figure \ref{fig:beta} we can see how feature selection affects the model's coefficients and pulls the majority of them towards zero. 

\section{Final thoughts on the basic Bayesian model} \label{sec:final_bayes}

Although we have clearly obtained some improvement regarding the classical approach, there is still more we can work with in the Bayesian scenario. Since the LASSO shrinks every coefficient by the same amount, we can lose some important information by lowering coefficients that matter to our problem, so we need another approach to select the most relevant features. Potentially, other means to introduce sparsity can lead to models where closed form conditional posteriors simple to sample from are unobtainable. Alternatives that could introduce sparsity in the model would render a rather complicated posterior, that would take much too long to explore via Markov-Chain Monte Carlo. With that in mind, it seems reasonable to find an approximated posterior, rather than an exact one, and employing a model that performs a feature selection more tailored to the problem of MER.

\chapter{Employing Variational Inference and Automatic Relevance Determination in Music Emotion Recognition}\label{vi_ard}

In this chapter we will introduce the basics of Variational Inference, as means to perform approximated rather than exact inference. This allows us have more freedom when it comes to the modeling process and also potentially reduces by great lengths the time spent training the model in comparison to MCMC methods, making this approach more attractive when dealing with large databases. The downside of this method is having to work with an approximation of the posterior distribution, rather than the exact one. We will also go into Automatic Relevance Determination as means of performing feature selection in a more efficient way than the LASSO to overcome the difficulties we have established in Chapter \ref{bayes}. Although this model does not return closed form conditional posterior distributions, we can perform the aforementioned method and work with an approximation, rather than implementing something like a Metropolis step within a Gibbs sampler that would be required to perform exact inference, increasing by some measure the time needed to run the model.

In Section \ref{sec:VI} we will go into the basics of Variational Inference and how it works, to then introduce Automatic Relevance Determination as means to perform feature selection in Section \ref{sec:ARD}. This two ideas will be joined and applied to our problem in Section \ref{sec:vi-ard}, with its results discussed in Section \ref{sec:vi-results}, where we will be able to compare and discuss the models previously tested alongside this one.

\section{Modeling with Variational Inference} \label{sec:VI}
Variational Inference (VI) \cite{Tam14, Bis06} uses optimization to approximate a posterior distribution without a closed form by a distribution easier to calculate, and this process is usually much faster than employing MCMC. %, that would perform an exact inference.
But how can we obtain that approximating distribution? Let us assume that our real posterior distribution for a parameter $\bftheta$ given some observed data $\bfx$ is $\p  (\bftheta|\bfx)$ and we would like to pick the most similar distribution $\q^*(\bftheta)$ from a family $\mathcal{D}$ of known and simpler distributions. There are several similarity measures that can be used \cite{Inm07}, but VI employs the Kullback-Leibler divergence, given by
\begin{equation} \label{eq:kl}
    \kl(\q(\bftheta) || \p  (\bftheta|\mathbf{X})) := \int_{-\infty}^{+ \infty} \q(\bftheta) \log \Bigg(\frac{\q(\bftheta)}{\p  (\bftheta | \bfx)} \Bigg) d\bftheta,
\end{equation}
a quantity %called relative entropy 
inspired by Information Theory \cite{thomas-cover}. It is an asymmetric and positive measure of the dissimilarity between two distributions, with $\kl = 0$ meaning they are the same. In Statistics it can be loosely interpreted as how much information will be lost when making the proposed approximation \cite{kul51}. 

\vspace{5mm}

So we must find $q^*(\bftheta)$ so that the KL-divergence is minimum:
$$\q^*(\bftheta) = \underset{q \in \mathcal{D}}{\mathrm{argmin}} ~ \kl (\q(\bftheta) || \p  (\bftheta|\mathbf{x})) $$

In order to calculate the exact value of the KL divergence we would need to know the exact distribution $\p  (\bftheta)$ which we do not. In Equation \ref{eq:ard} we can see how we can bypass this issue, by interpreting the KL definition as a relation of expected values and rearranging them: 

\begin{equation} \label{eq:ard}
\begin{split}
\kl (\q(\bftheta) || \p  (\bftheta|\bfx)) & = \int_{-\infty}^{+ \infty} \q(\bftheta) \log \Bigg(\frac{\q(\bftheta)}{\p  (\bftheta|\bfx)} \Bigg) d\bftheta \\
& = \bbe_\q[\log \q(\bftheta)] - \bbe_\q[\log \p  (\bftheta|\bfx)] \\
& =\bbe_\q[\log \q(\bftheta)] - \bbe_\q[\log \p  (\bftheta,\bfx)] + \bbe_\q[\log \p  (\bfx)]\\
& =\bbe_\q[\log \q(\bftheta)] - \bbe_\q[\log \p  (\bftheta,\bfx)] + \log \p  (\bfx)\\
\end{split}
\end{equation}

$$\Rightarrow \log ~ \p  (\bfx) = \kl (\q(\bftheta) || \p  (\bftheta|\bfx)) + \underbrace{\bbe_\q[\log \p  (\bftheta,\bfx)] -  \bbe_\q[\log \q(\bftheta)]}_{\elbo(\q)} $$

Note that the right side of the equation is constant over $\bftheta$ and the left is independent from $\theta$, so minimizing KL increases a quantity called \textit{variational lower bound} or \textit{evidence lower bound} (ELBO) \cite{min05}. That is, since the lowest possible value for KL is zero, the ELBO is the lowest value that can be achieved by $\log\p (\bfx)$, also known as the log marginal likelihood or log evidence of the model, the latter being more used in Bayesian inference, meaning the logarithm of the probability of the data given the model type \cite{Bos02}. 

One drawback of using the ELBO is that since we do not have access to the real posterior, we have no way of knowing whether we have reached the global minimum or a local one. One way to assess this is choosing random starting values and analysing results, but even if they remain identical there is no way of being certain, meaning the distribution $\q$ we are using to approximate $\p$ might not be the optimal one.

Here we will implement mean field VI \cite{vi-rev-stat}, meaning that the unknown parameters will be partitioned as $\bftheta = \{\bftheta_1, \dots, \bftheta_m\}$, which means we are assuming each partition is independent from the others, that is, members of the variational family $\mathcal{D}$ are given by $\q(\bftheta) = \q_1(\bftheta_1) \dots \q_m(\bftheta_m)$. In this scenario, we are able to employ the Coordinate Ascending Variational Inference (CAVI) algorithm \cite{Bis06, vi-rev-stat}. It can be proven that the CAVI algorithm converges to a local maximum of the ELBO \cite{Bis06}, and it can be then fully described as \cite{vi-rev-stat}:

\vspace{5mm}

\begin{algorithm}[H]
\caption{Coordinate Ascending Variational Inference.}
\SetAlgoLined

Input: A model $\p  (\bftheta, \bfx)$ and a dataset $\bfx$
    
 Output: A variational density $\q^*(\bftheta) = \q_1^*(\bftheta_1) \dots \q_m^*(\bftheta_m)$
    
Initialize the variational factors $\q_j(\bftheta_j)$
    
\While{ELBO has not converged}{
    \begin{itemize}
        \item [i) ] For $j = 1, \dots, m$, update $\q_j(\bftheta_j) \propto \exp\{\bbe_{-\q_j}[\log \p  (\bftheta_j, \bftheta_{-j}, \bfx)]\}$
        
        \item [ii) ] Compute the ELBO
    \end{itemize}}
    
Return: $\q^*(\bftheta)$
\end{algorithm}

\vspace{5mm}

%Where

%$$\q(\bftheta) = \prod_{j=1}^m \q_j (\bftheta_j, \eta_j) \Rightarrow \elbo(q) = ELBO(\eta_1, \dots, \eta_m) : \mathbb{R}^m \mapsto \mathbb{R}$$

CAVI's update rule can be derived from the ELBO by reorganizing its equation, as stated in Equation \ref{eq:elbo}. The first step is to rewrite the subtraction in function of $\q_j(\theta_j)$, letting the terms independent from it be absorbed into the constant $\mathcal{C} = \bbe_{-\q_j(\theta_{-j})}[\log(\p(\theta_{-j},\bfx))]$ and exploring the assumed independence between the variables in the second term:

\begin{equation} \label{eq:elbo}
\begin{split}
\elbo(\q)  & =  \bbe_{\q}[\log \p  (\bftheta,\bfx)] -  \bbe_{\q}[\log \q(\bftheta)] \\
\elbo(\q_j) & = \bbe_{\q_j} [\bbe_{-\q_j} [\log \p  (\bftheta_j, \bftheta_{-j}, \bfx)] ] -  \bbe_{\q_j}[\log \q_j(\bftheta_j)] + \bbe_{-\q_j}[\log(\p(\theta_{-j},\bfx))]\\
& = \bbe_{\q_j} [\bbe_{-\q_j} [\log \p  (\bftheta_j, \bftheta_{-j}, \bfx)] ] -  \bbe_{\q_j}[\log \q_j(\bftheta_j)] + \mathcal{C}\\
 \end{split}
\end{equation}
 
 Where $\bbe_{-\q_j}$ is the expected value with respect to $\q$, except on the variable $j$. If we make an educated guess, it is reasonable to assume $ \q_j(\bftheta_j) = \exp\{\bbe_{-\q_j}[\log \p  (\bftheta_j, \bftheta_{-j}, \bfx)]\}$, which would render $\bbe_{\q_j} = \bbe_{-\q_j(\theta_{-j})}[\log(\p(\theta_{-j},\bfx))]$, meaning $ \q_j(\bftheta_j) = \exp\{\bbe_{-\q_j}[\log \p  (\bftheta_j, \bftheta_{-j}, \bfx)]\}$ up to a normalization factor. Thus, by making $ \q^*_j(\bftheta_j) \propto \exp\{\bbe_{-\q_j}[\log \p  (\bftheta_j, \bftheta_{-j}, \bfx)]\}$ we maximize the $\elbo$.

Now there is an alternative to MCMC and we are able to explore other ways to introduce sparsity in our model.

\section{Feature selection with Automatic Relevance Determination} \label{sec:ARD}

Automatic Relevance Determination (ARD) \cite{WN07, Jan13, vi-rev-stat} is another way of introducing sparsity in a model by using a parameterized prior distribution that effectively prunes away redundant or superfluous coefficients instead of shrinking all of them by the same amount as we were doing before. It performs features selection in a similar manner of the LASSO, but it takes into consideration the observed error within each parameter to evaluate the importance of each feature separately, attributing individual weights to each one. The hyper parameters  $a_0, b_0, c_0$ and $d_0$ can be chosen to achieve purposely uninformative distributions to allow the model to learn more from the data. From here on, normal distributions will be parametrized by precision: $[\text{varicance}]^{-1}$. To implement ARD we are going to keep the linear regression model $\bfy = \bfx\bfbeta + \varepsilon $ but with the following modifications:

\begin{equation} \label{eq:ard_model}
\begin{split}
y_i | \bfbeta, \tau & \sim \text{N}(\mathbf{x}_i^T\bfbeta, \tau),~i=1, \dots, n \\
\bfbeta, \tau | \bfalpha & \sim \text{N}(\bfbeta | \mathbf{0}, \tau ~ \diag(\bfalpha)) \Gamma(\tau|a_0, b_0) \\
\alpha_j & \sim \Gamma(\alpha_j|c_0, d_0),~j = 1, \dots, p \\
\end{split}
\end{equation}

%Where

%\vspace{5mm}

%$\bfy = \left [ \begin{array}{c}

%a_1 \\
%\vdots \\
%a_n\ \end{array} \right] _ {n \times 1}$, where each $y_i$ correspond to the A value of the observation $y_i$.

%\vspace{5mm}

%$\bfbeta = \left [ \begin{array}{c}

%\bfbeta_{a1} \\
% \vdots \\
%\bfbeta_{an} \ \end{array} \right] _ {p \times1}$, where the array contain the coefficient values associated to each of the p parameters, the 1 being the intercept

%\vspace{5mm}

%$\bfx = \left [ \begin{array}{ccc}

 %- & \mathbf{x}_1^T & - \\
 % & \vdots &  \\
% - & \mathbf{x}_n^T & -\ \end{array} \right] _ {n\times p}$, where the arrays are explanatory variables corresponding to each observation $i$.
 
 \vspace{5mm}

 Notice that $\tau ~ \diag(\alpha)$ is a precision matrix rather than covariance as it is ordinary in Automatic Relevance Determination and we will continue to employ that for the remainder of this work.

\vspace{5mm}

In our scenario, inspired by the prior structure, we will assume that the distributions in the variational family factorize as

\begin{equation} \label{eq:mean-field}
\q(\bfbeta, \tau, \bfalpha ) = \q(\bfbeta, \tau) \prod_{j = 1}^{p} \q( \alpha_j ).
\end{equation}

\section{Employing VI and ARD} \label{sec:vi-ard}
And now we need to apply VI to our model in Equation \ref{eq:ard_model} so our conditional posterior will be:

\begin{equation} \label{eq:posterior_aprox}
\begin{split}
\p  (\bfbeta, \tau, \bfalpha|\bfy,\bfx) & \propto \p  (\bfy|\bfbeta, \tau, \bfalpha, \bfx)\p  (\bfbeta, \tau|\bfalpha)\p  (\bfalpha) \\
& = \p  (\bfy, \bfbeta, \tau, \bfalpha| \bfx) \\
& \approx \q(\bfbeta, \tau, \bfalpha) \\
& = \q(\bfalpha)\q(\bfbeta, \tau) \\
& = \Bigg[\prod_{j=1}^p \Gamma(\alpha_j|c_{j_*}, j_{j_*}) \Bigg] \times \Bigg[\text{N}(\bfbeta|\bfbeta_*, \tau V_*^{-1}) \Bigg] \times \Bigg[\Gamma(\tau| a_*, b_*) \Bigg],\\
\end{split}
\end{equation}
where the quantities $c_{j_*}, d_{j_*}, \beta_*, V_*^{-1}, a_*$ and $b_*$ are the parameters of the variational distribution defined previously.

% \begin{equation} \label{eq:posterior_model_ard_1}
% \begin{split}
% \p  (\bfy, \bfbeta, \tau, \bfalpha| \bfx)&  = \Bigg[ \prod_{i=1}^n \text{N}(y_i|\mathbf{x}_i^T\bfbeta, \tau) \Bigg] \times [\text{N}(\bfbeta|\mathbf{0}, \tau \diag(\bfalpha)) \Gamma(\tau|\theta_0,b_0)] \times \\
% & ~~~~ \Bigg[\prod_{d=1}^p \Gamma(\alpha_j|c_0, d_0) \Bigg] 
% \end{split}
% \end{equation}

Since these calculations are not authorial \cite{vi-rev-stat}, they will be omitted for the time being and detailed in Appendix \ref{Appendix:C} with only the results presented below:

 $$\q^*(\bfbeta, \tau, \bfalpha) = \q^*(\bfbeta, \tau)\q^*(\alpha_j) = \text{N}\left(\bfbeta|\bfbeta_*, \tau V_*^{-1} \right)\Gamma(\tau | a_*, b_*)\Gamma (c_{j*},d_{j*}),~j = 1, \dots, p$$ 
 
The updated parameters are given by:

\begin{equation} \label{eq:q-beta-tau}
\begin{split}
& \bfbeta_* = V_*\Bigg[\sum_{i=1}^n \mathbf{x}_i y_i \Bigg] \\
& V_*^{-1} = \sum_{i=1}^n \mathbf{x}_i\mathbf{x}_i^T - \bbe_{\bfalpha}[\diag(\bfalpha)] \\
& a_* = a_0 + \frac{n}{2} \\
& b_* = b_0 + \frac{1}{2}\sum_{i=1}^n y_i^2 + \bfbeta_*^T V_*^{-1}\bfbeta_*\\
& c_{j*} = c_0 + \frac{p}{2} \\
& d_{j*} =  d_0 + \frac{1}{2} \Bigg[[V_*]_{jj} + \beta_{*j}^2 \frac{a_*}{b_*}\Bigg],~j = 1, \dots, p.\\
\end{split}
\end{equation}

Now we are able to implement CAVI's algorithm as we did before, inspired by the one in \cite{vi-rev-stat}. The implementation od the algorithm was written for MATLAB in \cite{Jan13} and transcribed by us for Pyhton:

\vspace{5mm}

\begin{algorithm}[H] \label{alg:ACAVI}
\SetAlgoLined

Input: A model $\p  (\bfbeta, \tau, \bfalpha, \bfx)$ and a dataset $\bfx$
    
 Output: A variational density $\q^*(\bfbeta, \tau, \bfalpha) = \q^*(\bfbeta, \tau) \prod_{j=1}^p \q^*(\alpha_j)$
    
Initialize the parameters of the variational distributions $\q(\bfbeta, \tau) \prod_{j=1}^p \q_j(\alpha_j)$
\begin{itemize}
     \item [i) ] Update $a_*  = a_0 + \frac{n}{2} $
     \item [ii) ] Update $ c_{j*}  = c_0 + \frac{1}{2},~j = 1, \dots, p$
\end{itemize}
    
\While{ELBO has not converged}{
    \begin{itemize}
        \item [i) ] Update $\bfbeta_* = V_* \left[\sum_{i=1}^n \mathbf{x}_iy_i \right]$
        \item [ii) ] Update $V_*^{-1} = \sum_{i=1}^n \mathbf{x}_i\mathbf{x}_i^T + \diag(c_{1*}/d_{1*}, \dots, c_{p*}/d_{p*})$
        \item [iii) ] Update $b_*  = b_0 + \frac{1}{2}\Bigg(\sum_{i=1}^n y_i^2 - \bfbeta_*^T V_*^{-1}\bfbeta_*\Bigg) $
        \item [iv) ] Update $ d_{j*}  =  d_0 + \frac{1}{2} \Bigg[[V_*]_{jj} + \beta_{*j}^2 \frac{a_*}{b_*}\Bigg],~j = 1, \dots, p$
        \item [v) ] Compute the ELBO $= \bbe_{\q}[\log(\text{N}(\bfbeta_*, \tau V_*^{-1})\Gamma(a_*,b_*))] + \bbe[\log(\Gamma(c_*,d_*))]$
    \end{itemize}}
    
Return: $\q^*(\bfbeta, \tau, \bfalpha)$
 \caption{Coordinate Ascending Variational Inference for Music Emotion Recognition.}
 \end{algorithm}
 
 With these parameters we can use the approximated posterior distribution to implement the approximated predictive distribution in order to make predictions from a set of new data. 
 
 In this scenario, this approximated predictive density follows a Student's t distribution with location $\bfbeta_*^T\bbfx$, scale $(1 + \bbfx^T \bfv_*\bbfx)^{-1} \frac{a_*}{b_*}$ and $2a_*$ degrees of freedom, as demonstrated in \cite{Jan13} and adapted for the parameters and data we have in this work.
 
\section{Results} \label{sec:vi-results}

The ARD model was implemented with an adapted code from \cite{Jan13} and translated from MATLAB/Octave language.

The credibility intervals calculated for the ARD model were obtained from the approximated predictive density mentioned in Section \ref{sec:vi-ard}.

We can see in Tables \ref{table:results2A} and \ref{table:results2V} that ARD implemented with the CAVI algorithm showed significant improvement upon the previous results, with $270$ intervals containing the right values for Arousal and $231$ for Valence, and only taking a few seconds to run and draw samples from the variational distribution. The subset of songs that have both A-V values within the intervals built for them is more modest, reaching $131$. The training and test $R^2$ metric were also improved, reaching $0.75$ and $0.66$ for Arousal and $0.59$ and $0.29$ for Valence. 

\begin{table}[H]
\begin{center}
\begin{tabular}{||c c c c||} 
\hline
\textbf{Arousal} & Classic & Bayesian & ARD  \\ [0.5ex] 
 \hline\hline
Training $R^2$ &  $0.22$  & $0.60$ & $0.75$ \\
\hline
Test $R^2$ & $0.15$ & $0.58$  & $0.66$\\
\hline
Credibility intervals & - & $201$ & $270$\\[1ex]
\hline
 \hline
\end{tabular}
\end{center}
\caption{Comparison between classical, Bayesian and ARD models for Arousal.}
\label{table:results2A}
\end{table}

\begin{table}[H]
\begin{center}
\begin{tabular}{||c c c c||} 
\hline
\textbf{Valence}  & Classic & Bayesian & ARD \\ [0.5ex] 
 \hline\hline
Training $R^2$ &  $0.12$  & $0.45$ & $0.59$ \\
\hline
Test $R^2$ & $0.06$ & $0.25$  & $0.29$\\
\hline
Credibility intervals & - & $185$ & $231$\\[1ex]
\hline
 \hline
\end{tabular}
\end{center}
\caption{Comparison between classical, Bayesian and ARD models for Valence.}
\label{table:results2V}
\end{table}

\begin{figure}[H]
    \centering
        \subfloat{\includegraphics[width=0.5\textwidth]{Images/Beta_A.pdf}}
        \subfloat{\includegraphics[width=0.5\textwidth]{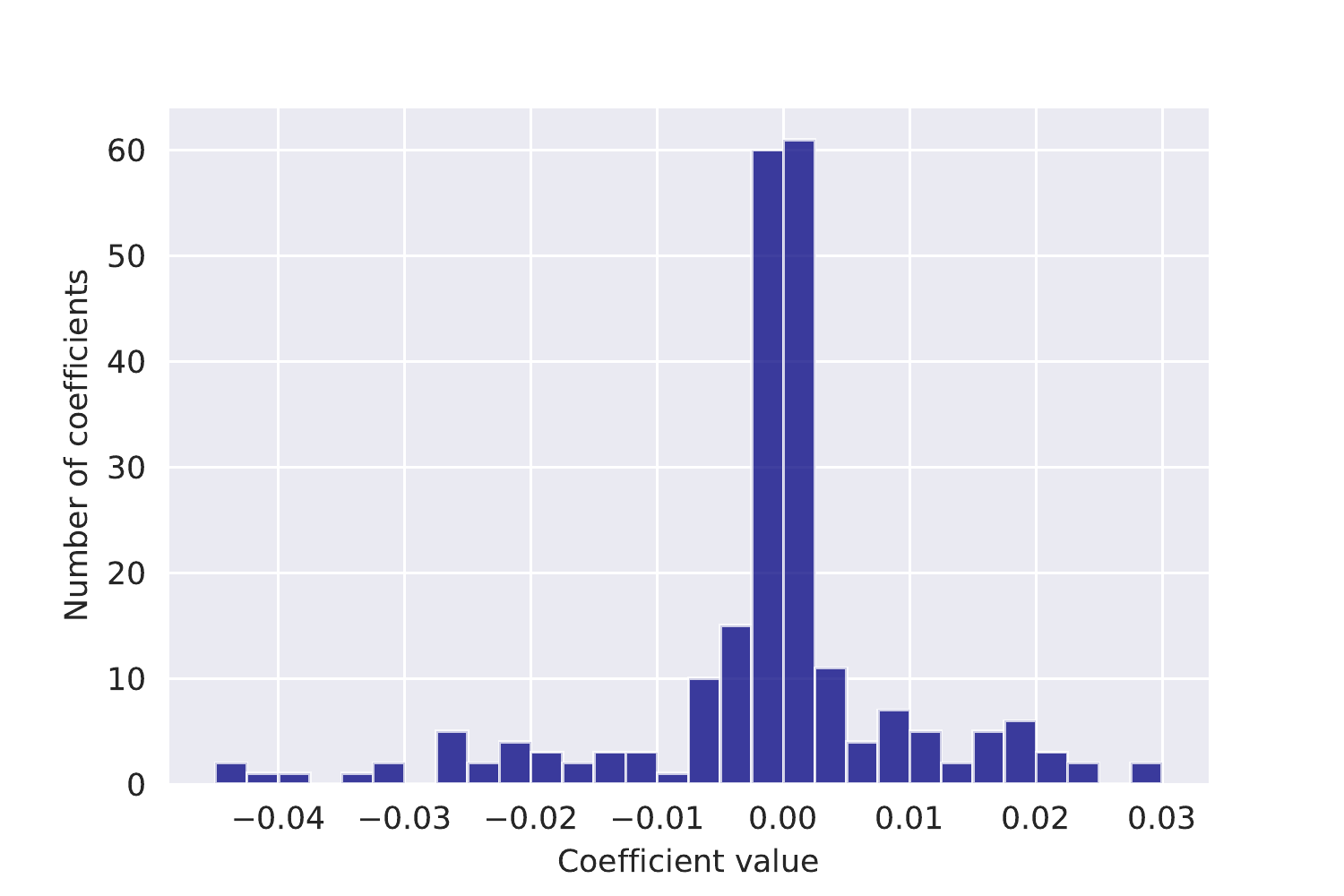}}
        \caption{Histograms of linear regression coefficients for Arousal with feature selection via LASSO (left) and  ARD (right).}
        \label{fig:beta_A}
        
\end{figure}

\begin{figure}[H]
    \centering
        \subfloat{\includegraphics[width=0.5\textwidth]{Images/Beta_V.pdf}}
        \subfloat{\includegraphics[width=0.5\textwidth]{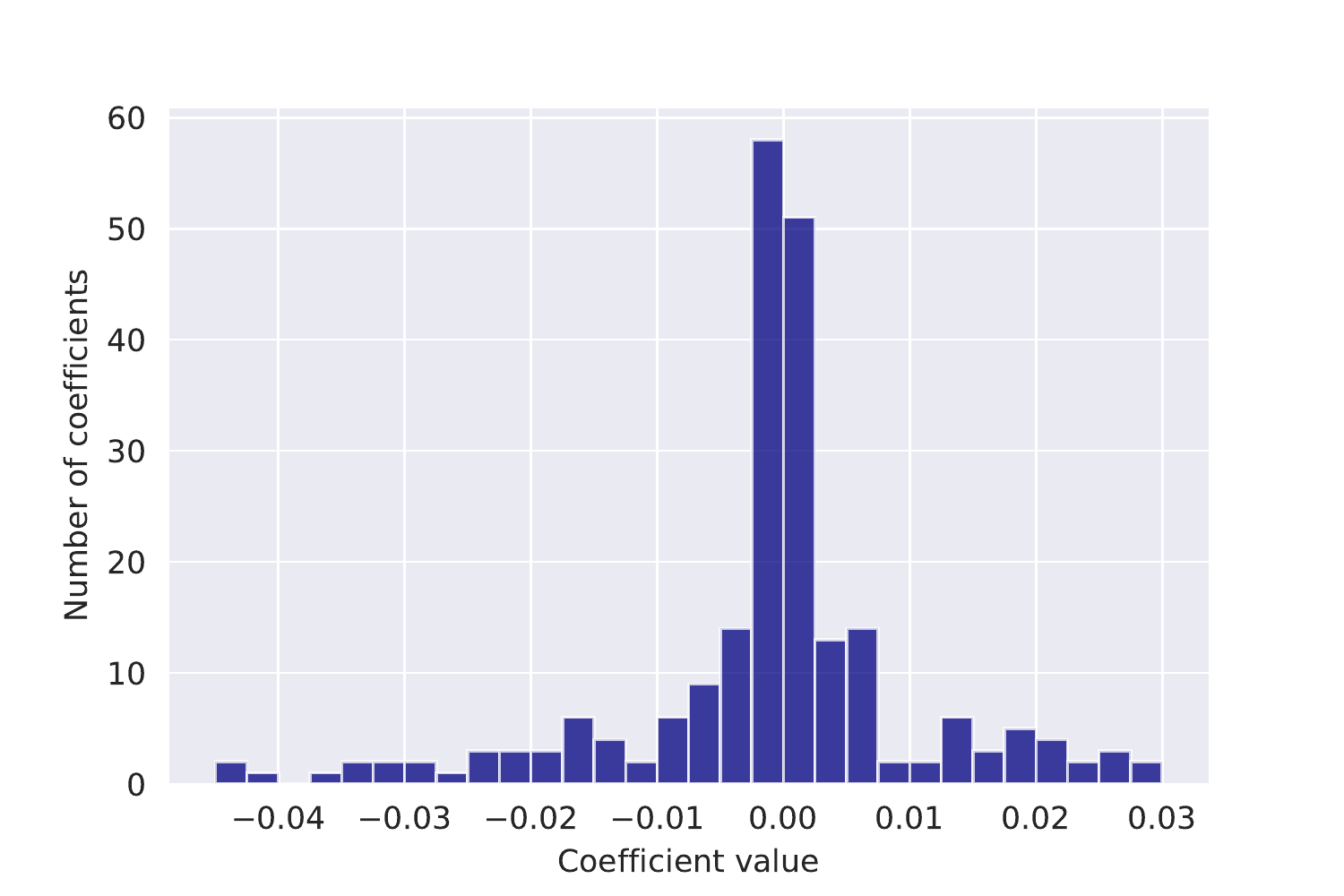}}
        \caption{Histograms of linear regression coefficients for Valence with feature selection via LASSO (left) and  ARD (right).}
        \label{fig:beta_V}
        
\end{figure}

Now, let us analyse how regularization affects the parameters, that is, the amount of regression coefficients that became smaller than $10^{-6}$ in absolute value after being regularized. Comparing  the Histograms in Figures \ref{fig:beta_A} and \ref{fig:beta_V} we can see how these feature selection approaches differ when applied to the same dataset. 

From the volume distribution around the columns centered in zero, we can see that the ARD model turns fewer coefficients to zero and allows others to be slightly higher, thus preserving information that was otherwise discarded in the LASSO model.

\chapter{Developing Multivariate Automatic Relevance Determination in Music Emotion Recognition}\label{MARD}

\begin{comment}

-my model
-how it is and how suitable it is
-how Cavi works i this scenario
-pseudo-code
\end{comment}

The model presented in Chapter \ref{vi_ard} ignored the substantial correlation between Arousal and Valence by treating these two response variables as independent, but the Multivariate Automatic Relevance Determination (MARD) model we seek to develop here accounts for that relationship. More generally, this model is capable of being adapted to deal with a dataset carrying $m$ dependent response variables, considering valuable information that the ARD model does not account for. In this work, however, we will only consider the scenario where $m=2$. We extended the ARD by generalizing the Normal-Gamma prior distribution jointly attributed to $\bfbeta$ and $\tau$ with a Normal-Wishart distribution attributed to $\bfbeta$ and the observations' precision matrix, setting requirements to adjust the generalized ARD to any problem.

We will begin by defining the bivariate model we are going to explore in Section \ref{sec:biv-model}, then develop it in Section \ref{sec:dev-model} where several calculations will be made to find its updating rules. The predictive distribution will be calculated and discussed in Section \ref{sec:pred-dist}, before testing the model in Section \ref{sec:mard-test} and analyzing results in \ref{sec:mard-results}. 

\section{Introducing a bivariate model}\label{sec:biv-model}
The model will follow the same structure as described in Equation \ref{eq:model}, but here it will be defined as a multivariate linear regression \cite{Ros05} $\bfy = \bfx\bfb + \bfe $, where each pair ($a_i, v_i$) is written as $\bbfy_i$ and the remaining variables will be described below. Underneath each matrix is stated its size, as to provide more clarity in the calculations that follow.

\vspace{5mm}

$\bfy = \left [ \begin{array}{cc}

a_1 & v_1\\
\vdots & \vdots \\
a_n & v_n\ \end{array} \right] _ {n\times2}$, where each pair $(a_i, v_i)$ correspond to the AV values of the observation $i$

\vspace{5mm}

$\bfb = \left [ \begin{array}{cc}

\beta^a_{1} & \beta^v_{1}\\
 \vdots & \vdots\\
\beta^a_{n} & \beta^v_{n} \ \end{array} \right] _ {p \times 2}$, where each pair $(\beta^a_{j}, \beta^v_{j})$ contains the regression parameters for Arousal and Valence
\vspace{5mm}

$\bfx = \left [ \begin{array}{ccc}

 - & \mathbf{x}_1^T & - \\
  & \vdots &  \\
 - & \mathbf{x}_n^T & -\ \end{array} \right] _ {n \times p}$, where the arrays $\mathbf{x}^T_i$ are explanatory variables corresponding to each observation $i$, with the first entry of each $\mathbf{x}_i^T$ is 1 to account for the intercept.

\vspace{5mm}

$\bfe = \left [ \begin{array}{cc}

\varepsilon_1^a & \varepsilon_1^v \\
 \vdots & \vdots   \\
\varepsilon_n^a & \varepsilon_n^a \ \end{array} \right] _ {n\times2}$, where each pair $\varepsilon_i^a, \varepsilon_i^v$ correspond to the observation errors  $i$ associated to the regression model employed and will be represented by $\varepsilon_i = \left[ \begin{array}{cc}
    \varepsilon_1^a  \\
     \varepsilon_1^v 
\end{array} \right]  $.

\subsection{Creating the foundation for MARD} \label{sec:found_mard}

 In order to implement ARD's intuition in a multivariate scenario several adaptations to that model had to be made, most remarkably by creating a precision matrix for the parameters that is related to $\tau\diag(\alpha)$ used by ARD in a way that allowed MARD to support any size of correlated responses. Note that a precision matrix will be attributed to the Normal distributions rather than a covariance matrix, just as defined in Chapter \ref{vi_ard}. 
 
 Before we can state this generalization, the Kronecker Product must be defined, as it is essential for its construction:
 
 \begin{theo}{\textbf{Kronecker Product:}}
Consider the matrices $\mathbf{A}_{m\times n}$ and $\mathbf{B}_{p\times q}$. The Kronecker Product between them is defined as:

\vspace{5mm}

$~~~~~~~~~~~~~~~~~~~~~~~~~\mathbf{A} \otimes \mathbf{B} = \left [ \begin{array}{ccc}

a_{11}\mathbf{B} & \dots & a_{1n}\mathbf{B} \\
 \vdots & \ddots & \vdots  \\
a_{m1}\mathbf{B} & \dots & a_{mn}\mathbf{B}\ \end{array} \right] _ {pm\times qn}$
\end{theo}

Thus, we can elaborate on ARD in order to create the generalized model MARD: 

\begin{equation}
\begin{split}
\bbfy_i & \sim \text{N}(x_i^T \bfb, \bfk)\\
\bfbeta|\bfk & \sim \text{N} (\mathbf{0}, \bfk \otimes \bfdelta) \\
\bfk & \sim \text{W}(\bfv_0,\nu_0) \\
\alpha_j & \sim \Gamma(c_{}, d_{}), j= 1, \dots p ~\text{i.i.d.}
\end{split}
\end{equation}
\vspace{5mm}
where $\bfk =
 \left [ \begin{array}{cc}

\tau_a & \kappa_{av} \\
\kappa_{av} & \tau_v   \ \end{array} \right] _ {2\times2}$ and
$\bfdelta = \left [ \begin{array}{ccc}

\alpha_{1} \\
 & \ddots \\
 & & \alpha_{p} \ \end{array} \right] _ {p\times p}$
 
\vspace{5mm}

Where $\text{W}$ denotes the Wishart distribution. Also, note that $\bfbeta$ is not $\bfb$, but its vectorization:

\begin{theo}{\textbf{Vectorization:}}
Consider the matrix $\mathbf{A}_{m \times n}$. The vectorization of $\mathbf{A}$, denoted by $\vect(\mathbf{A})$, is the $mn \times 1$ column vector obtained by stacking the columns of the matrix $\mathbf{A}$ on top of one another:

\vspace{5mm}

$~~~~~~~~~~~~~~~~~~~~~~~~~~~~~~\vect(\mathbf{A}) = \left [ \begin{array}{c}
a_{1,1}\\
\vdots \\
a_{m,1}\\
\vdots \\
a_{1,n} \\
\vdots \\
a_{m,n}\ \end{array} \right]$
\end{theo}

 \vspace{5mm} 
 
Therefore, the prior precision matrix of $\beta | \bfk$ will be given by

 \vspace{5mm}
 
 $\bfk \otimes \bfdelta = \left [ \begin{array}{c:c}

\tau_a \left [ \begin{array}{ccc} 
\alpha_{1} \\
& \ddots \\
& & \alpha_p
\end{array} \right] & \kappa \left [ \begin{array}{ccc}
\alpha_{1} \\
& \ddots \\
& & \alpha_p
\end{array} \right]\\  \hdashline
 \kappa \left [ \begin{array}{ccc}
\alpha_{1} \\
& \ddots \\
& & \alpha_p
\end{array} \right] & \tau_v \left [ \begin{array}{ccc}
\alpha_{1} \\
& \ddots \\
& & \alpha_p
\end{array} \right]\\
\end{array} \right] _ {2p\times2p}$
 
 \vspace{5mm}

Further along, we will denote $\bfk \otimes \bfdelta$ by $\bfq$ to facilitate understanding. In this notation, $\bfb | \bfk$ follows the so-called matrix normal distribution \cite{gup99}, therefore, $\bfk^{-1}$ represents covariance on its columns while $\bfdelta^{-1}$ accounts for the lines, so we are assuming uncorrelated features as priors but the columns can have some correlation influenced by the correlation between Arousal and Valence.

\begin{remark}
Notice that even though the matrix $\bfk \otimes \bfdelta$ is related to the precision matrix % $\Sigma^{-1} = 
$\tau \diag(\alpha)$ in  ARD, initially we hoped to use the precision matrix $\bflambda$ as defined below:

 \vspace{5mm}
 
$\bflambda = \left [ \begin{array}{cccccc}
\tau_a\alpha_{a1} \\
 & \ddots \\
 & & \tau_{a}\alpha_{ap} \\
 & & & \tau_{v}\alpha_{v1} \\
 & & & & \ddots  \\
 & & & & &  \tau_{v}\alpha_{vp}\ \end{array} \right] _ {2p\times2p}$

 \vspace{5mm}

As we can see, $\bflambda$ is more similar to $\tau\diag(\alpha)$ and it allowed each feature to have its own regularization parameter for Arousal and Valence. This option does not need to be excluded as it is possible to perform MCMC, utilize some form of Variational Inference that is not CAVI or choose a variational family where this precision matrix works, but those were not adequate for this work and another approach became preferable.
So even though we have lost the information different $\alpha$'s would have given to each parameter related to Arousal and Valence separately it is easier to render a model with joint alphas while keeping some of the information they carry.
\end{remark} 

\section{Developing the model}\label{sec:dev-model}

Here we will work through the necessary calculations to deduce CAVI's updating rules for MARD and the variational families within CAVI as done in Chapter \ref{vi_ard}.

\subsection{Likelihood}

With the definitions set in Section \ref{sec:found_mard} we can begin to calculate the model's likelihood in the following way, considering the errors to be independent and follow a Normal distribution \cite{Ros05}.

\begin{equation} \label{eq:likili_multi_model}
\begin{split}
\p (\bfe | \bfk) & \propto \prod_{i=1}^n |\bfk ^{-1}|^{-\frac{1}{2}}\exp \Bigg\{-\frac{1}{2} \bfeps_i^T\bfk \bfeps_i \Bigg\} \\
& \propto |\bfk|^{\frac{n}{2}} \exp \Bigg\{-\frac{1}{2} \sum_{i=1}^n\bfeps_i^T  \bfk \bfeps_i \Bigg\} \\
& \propto |\bfk|^{\frac{n}{2}} \exp \Bigg\{-\frac{1}{2} \tr(\bfe^T \bfe \bfk)\Bigg\} \\
\end{split}
\end{equation}

With $\tr$ being the trace of $\bfe^T \bfe \bfk = \left [ \begin{array}{cc}
\sum_{i=1}^n (\varepsilon_i^a)^2 & \sum_{i=1}^n \varepsilon_i^a\varepsilon_i^v \\
 \sum_{i=1}^n \varepsilon_i^a\varepsilon_i^v & \sum_{i=1}^n (\varepsilon_i^v)^2
 \ \end{array} \right]\bfk$, that is, the sum of its main diagonal: $\sum_{i=1}^n \tau_a(\varepsilon_i^a)^2 + 2\varepsilon_i^a\varepsilon_i^v \kappa_{av} + \tau_v(\varepsilon_i^v)^2 = \sum_{i=1}^n\bfeps_i^T  \bfk \bfeps_i$.

Now we explore the relation $ \bfy = \bfx\bfb - \bfe \Rightarrow \bfe  = \bfy - \bfx\bfb$ to further the calculations. It performs a change of variable from $\bfe$ to $\bfy$, and the jacobian is constant and unitary \cite{Ros05}. The variable $\mathbf{S} = \bfy - \bfx\hat{\bfb}$, $\hat{\bfb} = (\bfx^T\bfx)^{-1}\bfx^T\bfy$ being the least squares estimate of $\bfb$, will also be introduced to facilitate calculations. Below we have manipulated the terms inside the exponential in order to obtain something quadratic in $\bfb$ which is close to the desired normal distribution.

\begin{equation} \label{eq:like-new}
\begin{split}
\p (\bfy | &\bfx, \bfbeta, \bfk)  \propto |\bfk|^{\frac{n}{2}}\exp \Bigg\{-\frac{1}{2} \tr((\bfy - \bfx\bfb)^T(\bfy - \bfx\bfb)\bfk)\Bigg\} \\
&  \propto  |\bfk|^{\frac{n}{2}}  \exp \Bigg\{ -\tr\Bigg(\frac{1}{2}\mathbf{S}^T\mathbf{S} \bfk \Bigg)\Bigg\} \exp \Bigg\{-\frac{1}{2} \tr((\bfb - \hat{\bfb})^T \bfx^T\bfx (\bfb - \hat{\bfb}) \bfk)\Bigg\} \\
\end{split}
\end{equation}

\vspace{5mm}

The result we achieve on Equation \ref{eq:like-new} very closely resembles a normal distribution and we will use Vectorization and the Kronkcker product to derive some important relations to finish that off.

\vspace{10mm}

The following relations occur from properties of vectorizarion as defined above, further explored in \cite{mac13} and proved in Chapter $2$ of \cite{mag88}

\begin{theorem}{\textbf{Properties of Vectorization and the Kronecker Product:}} \label{theo:tr_vec_kro}
Consider the matrices $A,B ~~ \text{and} ~~ C$ as matrices such that the matrix product ABC is defined. We can relate vectorization and the Kronecker product as follows:

\vspace{5mm}

$$   \tr(AB) = \vect(A^T)^T \vect(B)$$
$$ \vect(ABC) = (C^T \otimes A) \vect (B)$$
\end{theorem}

Recalling that $\vect(\bfb) = \bfbeta$ we can develop a inner section of Equation \ref{eq:like-new} as:

\begin{equation} \label{vectorization}
\begin{split}
 \tr((\bfb - \hat{\bfb})^T \bfx^T\bfx (\bfb - \hat{\bfb}) \bfk) &  = \vect(\bfb - \hat{\bfb})^T \vect(\bfx^T\bfx(\bfb - \hat{\bfb})\bfk) \\
&   = \vect(\bfb - \hat{\bfb})^T\bfk \otimes \bfx^T\bfx (\bfbeta - \hat{\bfbeta}) \\
&  = (\bfbeta - \hat{\bfbeta})^T \bfk \otimes \bfx^T\bfx (\bfbeta - \hat{\bfbeta})\\
\end{split}
\end{equation}

\vspace{5mm}

So we can follow up Equation \ref{eq:like-new} with

\begin{equation}
\begin{split}
\p (\bfy | \bfx, \bfbeta, \bfk) & \!\propto\! |\bfk|^{\frac{n}{2}}\!\exp\!\Bigg\{\!-\!\tr\!\Bigg(\!\frac{1}{2}\mathbf{S}^T\mathbf{S} \bfk \!\Bigg)\!\Bigg\}\! \exp\! \Bigg\{\!-\!\frac{1}{2} (\bfbeta - \hat{\bfbeta})^T \bfk\! \otimes\! \bfx^T\bfx (\bfbeta - \hat{\bfbeta})\!\Bigg\}  
\end{split}
\end{equation}

\vspace{5mm}

To simplify the notation, $\bfk \otimes \bfx^T\bfx$ will be written as $\bfj$.

\subsection{Computation of variational distributions}

Joining our likelihood and prior distributions previously set, we obtain the following posterior:

\begin{equation}
\begin{split}
\p (\bfy, \bfbeta, \bfk, \bfalpha| \bfx) & \propto \p (\bfy | \bfbeta, \bfk, \bfalpha, \bfx) \times \p (\bfbeta, \bfk | \bfalpha) \times \p (\bfalpha)\\
& \propto \! |\bfk|^{\frac{n}{2}}\!\exp\! \Bigg\{\! -\! \frac{1}{2} \tr (\bfs^T\bfs\bfk)\! \Bigg\}\! \exp\! \Bigg\{\!-\!\frac{1}{2} (\bfbeta\! -\! \bfhbeta)^T\!\bfj(\bfbeta \!-\! \bfhbeta))\!\Bigg\}\! \\
& ~~~~ \times \!|\bfq|^{\frac{1}{2}}\! \exp\! \Bigg\{\!\! -\!\frac{1}{2} \bfbeta^T\bfq\bfbeta\! \Bigg\}\! \times\! |\bfk|^{\frac{\nu_0 - 3}{2}} \!\exp \!\Bigg\{\!\! - \!\frac{1}{2} \tr (\bfv_0^{-1} \bfk)\! \Bigg\}\\
& ~~~~  \times \prod_{j=1}^p \alpha_{j}^{c -1} \exp \Bigg\{-d \alpha_j\Bigg\} \\
\end{split}
\end{equation}

\vspace{5mm}

The mean-field variational model will be written as 

$$\q(\bfbeta, \bfk) \prod_{j=1}^p \q(\alpha_{j}),$$
Which is reasonable, taking into consideration MARD's model. The posterior above will be used to find CAVI's updating rule as shown in Chapter \ref{vi_ard}.

To carry on some calculations in this scenario, it will be necessary using
the following properties, proven in, \cite{hor91}:

\vspace{30mm}

\begin{theorem}{\textbf{Kronecker Product Properties:}} \label{theo:kro_prop}
Let $\mathbf{A}, \mathbf{B}, \mathbf{C}, \mathbf{D}, \mathbf{E}_{n \times n}$ and  $\mathbf{F}_{m \times m}$ be matrices of appropriate sizes 
\begin{equation} \nonumber
\begin{split}
\mathbf{A} \otimes(\mathbf{B} + \mathbf{C}) & = \mathbf{A} \otimes \mathbf{B} + \mathbf{A} \otimes \mathbf{C} \\
(\mathbf{A} \otimes \mathbf{B})^{-1} & =  \mathbf{A}^{-1} \otimes \mathbf{B}^{-1}\\
(\mathbf{A} \otimes \mathbf{B})^T & =  \mathbf{A}^T \otimes \mathbf{B}^T \\
(\mathbf{A} \otimes \mathbf{B})(\mathbf{C} \otimes \mathbf{D}) & = (\mathbf{AC}) \otimes (\mathbf{BD}) \\|\mathbf{E} \otimes \mathbf{F}| & = |\mathbf{E}|^m|\mathbf{F}|^n
\end{split}
\end{equation}
\end{theorem}

\subsection{Developing calculations for CAVI's updating rules}

Firstly, let us observe this posterior to find the CAVI's updating rules for $\bfbeta, \bfk$:

\begin{equation} \label{eq:q_beta_k}
\begin{split}
\q^*(\bfbeta, \bfk) & \propto \exp \Bigg\{\bbe_{- \bfbeta, \bfk}\Bigg[\log \p (\bfy | \bfbeta, \bfk, \bfalpha, \bfx) + \log \p (\bfbeta, \bfk | \bfalpha) + \log \p (\bfalpha)\Bigg]  \Bigg\} \\
& \propto\! \exp\! \Bigg\{\!\bbe_{- \bfbeta, \bfk}\!\Bigg[\!\log \!\!\Bigg (\!\! |\bfk|^{\frac{n}{2}}\!\exp\! \Bigg\{\!\! -\! \frac{1}{2} \tr (\bfs^T\bfs\bfk)\! \Bigg\}\!\! \exp\! \Bigg\{\!-\!\frac{1}{2} (\bfbeta\! - \!\bfhbeta)^T\!\bfj(\bfbeta \!-\! \bfhbeta))\!\!\Bigg\}\!\! \Bigg)\! +\\
& ~~~~ \log\! \Bigg(\! |\bfq|^{\frac{1}{2}}\! \exp \!\Bigg\{\! -\! \frac{1}{2}\! \bfbeta^T \bfq \bfbeta\! \Bigg\}\! \Bigg)\!  +\! \log\! \Bigg(\!   |\bfk|^{\frac{\nu_0 - 3}{2}} \exp\!  \Bigg\{\! -\! \frac{1}{2} \tr \bfv_0^{-1} \bfk\!\Bigg\} \!\Bigg)\!\Bigg]\! \Bigg\} \\
& \propto \exp \Bigg\{\bbe_{- \bfbeta, \bfk}\Bigg[\frac{n}{2} \log|\bfk|  - \frac{1}{2} \tr (\bfs^T\bfs\bfk)  -\frac{1}{2} (\bfbeta - \bfhbeta)^T\bfj(\bfbeta - \bfhbeta)) + \\
& ~~~~ \frac{1}{2} \log |\bfk|^p |\bfdelta|^2 - \frac{1}{2} \bfbeta^T \bfq \bfbeta  + \frac{\nu_0 - 3}{2} \log |\bfk| 
- \tr \frac{1}{2} \bfv_0^{-1} \bfk \Bigg] \Bigg\} \\
& \propto \exp \Bigg\{ -\frac{1}{2} (\bfbeta - \bfhbeta)^T\bfj(\bfbeta - \bfhbeta)) + \frac{\nu_0 - 3 + n + p}{2} \log |\bfk| \\
& ~~~~  - \frac{1}{2} \bfbeta^T \bbe_{- \bfbeta, \bfk}[\bfq] \bfbeta 
- \frac{1}{2} \tr (\bfv_0^{-1} \bfk + \bfs^T\bfs\bfk) \Bigg\}
\end{split}
\end{equation}

Firstly, let us make a brief pause here to explain some actions in the computations above. The value $\bbe_{- \bfbeta, \bfk} [\log(\alpha_{a1} \dots \alpha_{ap},\alpha_{v1} \dots \alpha_{vp})]$ is unrelated to our variables of interest $\bfbeta$ and $\bfk$ so it can be disregarded. The expected value of $\bfq$ requires more care:

$$\bbe_{- \bfbeta, \bfk} [\bfq] ~~~~~~ =  \left [ \begin{array}{c:c}

\tau_a \left [ \begin{array}{ccc} 
\bbe_{- \bfbeta, \bfk} [\alpha_{1}] \\
& \ddots \\
& & \bbe_{- \bfbeta, \bfk} [\alpha_p]
\end{array} \right] & \kappa \left [ \begin{array}{ccc}
\bbe_{- \bfbeta, \bfk} [\alpha_{1}] \\
& \ddots \\
& & \bbe_{- \bfbeta, \bfk} [\alpha_p]
\end{array} \right]\\  \hdashline
 \kappa \left [ \begin{array}{ccc}
\bbe_{- \bfbeta, \bfk} [\alpha_{1}] \\
& \ddots \\
& & \bbe_{- \bfbeta, \bfk} [\alpha_p]
\end{array} \right] & \tau_v \left [ \begin{array}{ccc}
\bbe_{- \bfbeta, \bfk} [\alpha_{1}] \\
& \ddots \\
& & \bbe_{- \bfbeta, \bfk} [\alpha_p]
\end{array} \right]\\
\end{array} \right] _ {2p\times2p} $$

$$ =  \left [ \begin{array}{c:c}

\tau_a \left [ \begin{array}{ccc} 
\frac{c_{*}}{d_{*}} \\
& \ddots \\
& & \frac{c_{*}}{d_{*}}
\end{array} \right] & \kappa \left [ \begin{array}{ccc}
\frac{c_{*}}{d_{*}} \\
& \ddots \\
& & \frac{c_{*}}{d_{*}}
\end{array} \right]\\  \hdashline
 \kappa \left [ \begin{array}{ccc}
\frac{c_{*}}{d_{*}} \\
& \ddots \\
& & \frac{c_{*}}{d_{*}}
\end{array} \right] & \tau_v \left [ \begin{array}{ccc}
\frac{c_{*}}{d_{*}} \\
& \ddots \\
& & \frac{c_{*}}{d_{*}}
\end{array} \right]\\
\end{array} \right] _ {2p\times2p} $$

To make the notation more efficient the last matrix will be noted as $$\bfq_* = \bfk \otimes \bfdelta_*, \text{ being } \bfdelta_* = \left [ \begin{array}{ccc}

\frac{c_{*}}{d_{*}} \\
& \ddots \\
& & \frac{c_{*}}{d_{*}} \ \end{array} \right] _ {p\times p}$$ 
as we further the calculations in Equation \ref{eq:q_beta_k}. The exponentials will also be manipulated to fulfill our goal of identifying a Normal-Wishart distribution in this posterior. This distribution would parallel well with the idea of ARD and the posterior we have indicates that this this can be achieved. Continuing Equation \ref{eq:q_beta_k}, we have:

\begin{equation} \label{eq:q_beta_k_2}
\begin{split}
\q^*(\bfbeta, \bfk) & \propto \exp \Bigg\{\Bigg[ -\frac{1}{2} ((\bfbeta - \bfhbeta)^T\bfj(\bfbeta - \bfhbeta) +  \bfbeta^T \bfq_* \bfbeta) \Bigg] \\
& ~~~~ + \frac{\nu_0 - 3 + n + p}{2} \log |\bfk| + 
- \frac{1}{2} \tr (\bfv_0^{-1} \bfk + \bfs^T\bfs\bfk)  \Bigg\} \\
\end{split}
\end{equation}

\vspace{5mm}

From Equation \ref{eq:q_beta_k_2} we will find a Normal-Wishart distribution and therefore, the updating rules for the respective variational parameters. In order to do that we need only to identify the quadratic part of the Normal distribution and discover its parameters, so let us consider 
$$f(\bfbeta) = (\bfbeta - \hat{\bfbeta})^T \bfj(\bfbeta - \hat{\bfbeta}) + \bfbeta^T \bfq_* \bfbeta $$
and
$$g(\bfbeta) = (\bfbeta - \bfbeta_*)^T \bfk_* (\bfbeta - \bfbeta_*) $$
where
$$f(\bfbeta) = g(\bfbeta)+{\gamma}$$

with $\gamma$ being a constant value independent from $\bfbeta$.

\vspace{5mm}

First we will aim to identify a Normal distribution exploring similarities in $f(\bfbeta)$ and $g(\bfbeta)$. Ideally, the differences between them ($\gamma$) will be reallocated in a Wishart distribution in $\q^*(\bfbeta, \bfk)$

\vspace{5mm}

To find the mean and precision matrix of this distribution we know that 
$$g'(\bfbeta) = 2(\bfbeta - \bfbeta_*)^T \bfk_* = 0 \Rightarrow \bfbeta = \bfbeta_*$$
and
$$g''(\bfbeta) = 2\bfk_* $$
So we will obtain these parameters for $f(\bfbeta)$ in the same way:
\begin{equation} \label{eq:f_mean}
\begin{split}
f'(\bfbeta) = 2(\bfbeta - \hat{\bfbeta})^T \bfj + 2\bfbeta^T \bfq_* &  = 0\\
2[(\bfbeta^T - \hat{\bfbeta}^T) \bfj + \bfbeta^T \bfq_*] & = 0 \\
\bfbeta^T\bfj - \hat{\bfbeta}^T \bfj + \bfbeta^T \bfq_* & = 0 \\
\bfbeta^T(\bfj + \bfq_*)  & = \hat{\bfbeta}^T \bfj  \\
\bfbeta^T & = \hat{\bfbeta}^T \bfj (\bfj + \bfq_*)^{-1} \\
\bfbeta & = (\bfj + \bfq_*)^{-1} \bfj \bfhbeta \\
\end{split}
\end{equation}
\begin{equation} \label{eq:f_std}
\begin{split}
f''(\bfbeta) =2( \bfj + \bfq_*)
\end{split}
\end{equation}

So the first part of our variational distribution follows a normal distribution with parameters $\bfbeta_* = (\bfj + \bfq_*)^{-1} \bfj \bfhbeta$ and $\bfk_* = \bfj + \bfq_*$. Now we must separate what differs $f(\bfbeta)$ and $g(\bfbeta)$. 

Opening up the terms if $f(\bfbeta)$ we have:
\begin{equation} \label{eq:fopen}
\begin{split}
f(\bfbeta) = \bfbeta^T\bfj\bfbeta - 2\bfhbeta^T\bfj\bfbeta  + \bfbeta^T\bfq_*\bfbeta + \bfhbeta^T\bfj\bfhbeta
\end{split}
\end{equation}

Opening up the terms in $g(\bfbeta)$ we have the following:
\begin{equation} \label{eq:gopen}
\begin{split}
g(\bfbeta) & =(\bfbeta - \bfbeta_*)^T\bfk_*(\bfbeta - \bfbeta_*) \\
 & = [\bfbeta - (\bfj + \bfq_*)^{-1} \bfj \bfhbeta]^T(\bfj + \bfq_*)[\bfbeta - (\bfj + \bfq_*)^{-1} \bfj \bfhbeta] \\
 & =  [\bfbeta^T - \bfhbeta^T\bfj(\bfj + \bfq_*)^{-1}](\bfj + \bfq_*)[\bfbeta - (\bfj + \bfq_*)^{-1} \bfj \bfhbeta]\\
 & = [\bfbeta^T(\bfj + \bfq_*) - \bfhbeta^T\bfj(\bfj + \bfq_*)^{-1}(\bfj + \bfq_*)][\bfbeta - (\bfj + \bfq_*)^{-1} \bfj \bfhbeta] \\
 & = [\bfbeta^T(\bfj + \bfq_*) - \bfhbeta^T\bfj][\bfbeta - (\bfj + \bfq_*)^{-1} \bfj \bfhbeta] \\
 & = \bfbeta^T(\bfj + \bfq_*)\bfbeta - \bfhbeta^T\bfj\bfbeta - \bfbeta^T\bfj\bfhbeta + \bfhbeta^T\bfj(\bfj + \bfq_*)^{-1}\bfj\bfhbeta \\
 & = \bfbeta^T\bfj\bfbeta + \bfbeta^T\bfq_*\bfbeta - \bfhbeta^T\bfj\bfbeta - \bfbeta^T\bfj\bfhbeta + \bfhbeta^T\bfj(\bfj + \bfq_*)^{-1}\bfj\bfhbeta \\
& = \bfbeta^T\bfj\bfbeta + \bfbeta^T\bfq_*\bfbeta - 2\bfhbeta^T\bfj\bfbeta + \bfhbeta^T\bfj(\bfj + \bfq_*)^{-1}\bfj\bfhbeta
\end{split}
\end{equation}

From the four terms in Equation \ref{eq:fopen} we can already see the first in Equation \ref{eq:gopen}, respectively. To explore the last term in $g(\bfbeta)$, the Woodbury matrix identity will be applied.

\begin{theorem}{\textbf{Woodbury Identity:}}
Let $\mathbf{A}, \mathbf{U}, \mathbf{C}$ and $\mathbf{V}$ be matrices of adequate sizes, then
$(\mathbf{A} + \mathbf{UCV})^{-1} = \mathbf{A}^{-1} - \mathbf{A}^{-1}\mathbf{U}(\mathbf{C}^{-1} + \mathbf{VA}^{-1}\mathbf{U}^{-1})^{-1}\mathbf{VA}^{-1}$
\end{theorem}

Considering $\mathbf{A} = \bfj, \mathbf{C} = \bfq_*, \mathbf{U} ~~ \text{and} ~~ V = \bfi$, the identity matrix, we can write $(\bfj + \bfq_*)^{-1} = \bfj^{-1} - \bfj^{-1}(\bfq_*^{-1} + \bfj^{-1})^{-1}\bfj^{-1} = \bfj^{-1} - \bfj^{-1}(\bfq_*^{-1} + \bfj^{-1})^{-1}\bfj^{-1}$. With this, the remaining term of $g(\bfbeta)$ will be opened as well:
\begin{equation} \label{eq:gopen-cont}
\begin{split}
g(\bfbeta) & = \bfbeta^T\bfj\bfbeta + \bfbeta^T\bfq_*\bfbeta - 2\bfhbeta^T\bfj\bfbeta - \bfhbeta^T\bfj(\bfj + \bfq_*)^{-1}\bfj\bfhbeta \\
 & = \bfbeta^T\bfj\bfbeta + \bfbeta^T\bfq_*\bfbeta - 2\bfhbeta^T\bfj\bfbeta - \bfhbeta^T\bfj[\bfj^{-1} - \bfj^{-1}(\bfq_*^{-1} + \bfj^{-1})^{-1}\bfj^{-1}]\bfj\bfhbeta \\
 & = \bfbeta^T\bfj\bfbeta + \bfbeta^T\bfq_*\bfbeta - 2\bfhbeta^T\bfj\bfbeta -\bfhbeta^T\bfj\bfhbeta + \bfhbeta^T(\bfq_*^{-1} + \bfj^{-1})^{-1}\bfhbeta
\end{split}
\end{equation}

Now we have found $f(\bfbeta)$ and the last term in $f(\bfbeta)$ should be allocated in the Wishart part of the variational distribution. So we can rewrite Equation \ref{eq:q_beta_k_2} as:
\begin{equation} \label{eq:ard_b_q}
\begin{split}
\q^*(\bfbeta, \bfk) & \propto \exp \Bigg\{\Bigg[ -\frac{1}{2} ((\bfbeta - \bfbeta_*)^T\bfk_*(\bfbeta - \bfbeta_*) - \bfhbeta^T(\bfq_*^{-1} + \bfj^{-1})^{-1}\bfhbeta) \Bigg] \\
& ~~~~ + \frac{\nu_0 - 3 + n + p}{2} \log |\bfk| + 
- \frac{1}{2} \tr ((\bfv_0^{-1} + \bfs^T\bfs)\bfk)  \Bigg\},
\end{split}
\end{equation}
where $\bfbeta_* = (\bfj + \bfq_*)^{-1} \bfj \bfhbeta$ and $\bfk_* = \bfj + \bfq_*$. The term $\bfhbeta^T(\bfq_*^{-1} + \bfj^{-1})^{-1}\bfhbeta$ will be transformed into something that can be incorporated in the Wishart part of Equation \ref{eq:ard_b_q}. 

We will also employ the properties listed in Theorem \ref{theo:tr_vec_kro} and the ones in Definition \ref{theo:kro_prop}:
\begin{equation}
\begin{split}
\bfhbeta^T(\bfq_*^{-1} + \bfj^{-1})^{-1}\bfhbeta) & = \bfhbeta^T[(\bfk \otimes \bfdelta_*)^{-1} + (\bfk \otimes \bfx^T\bfx)^{-1}]^{-1}\bfhbeta \\
& = \bfhbeta^T[\bfk^{-1} \otimes \bfdelta_*^{-1} + \bfk^{-1} \otimes \bfx^T\bfx]^{-1} \bfhbeta \\
& = \bfhbeta^T[\bfk^{-1} \otimes (\bfdelta_*^{-1} + \bfx^T\bfx^{-1})]^{-1} \bfhbeta\\
& = \bfhbeta^T [\bfk \otimes (\bfdelta_*^{-1} + \bfx^T\bfx^{-1})^{-1}]\bfhbeta \\
& = \tr(\bfhb^T(\bfdelta_*^{-1} + \bfx^T\bfx^{-1})^{-1}\bfhb \bfk),
\end{split}
\end{equation}
that can be inserted in Equation \ref{eq:ard_b_q} as
\begin{equation} \label{eq:ard_b_q_final}
\begin{split}
\q^*(\bfbeta, \bfk) & \propto \exp \Bigg\{ -\frac{1}{2} ((\bfbeta - \bfbeta_*)^T\bfk_*(\bfbeta - \bfbeta_*)) + \frac{\nu_* - 3}{2} \log |\bfk| - \frac{1}{2} \tr (\bfv_*^{-1} \bfk)  \Bigg\},
\end{split}
\end{equation}
where
\begin{equation}
\begin{split}
\nu_* & = \nu_0 + n + p \\
\bfv_*^{-1} & = \bfv_0^{-1} + \bfs^T\bfs + \bfhb^T(\bfdelta_*^{-1} + \bfx^T\bfx^{-1})^{-1}\bfhb.
\end{split}
\end{equation}

Now we need to find the distributions for each $\q^*(\alpha_{j})$. Since they will be identically calculated, suffices to calculate for one of theses $\alpha$'s

\begin{equation} \label{eq:q-alpha-model-ard-2}
\begin{split}
\q^*(\alpha_j) & \propto \exp \Bigg\{\bbe_{-\alpha_j} \Bigg[\log \p (\bfy, \bfbeta, \bfk, \bfalpha|\mathbf{X}) \Bigg]   \Bigg\} \\
& \propto \exp \Bigg\{ \bbe_{-\alpha_j} \Bigg[\log  \p (\bfy| \bfbeta, \bfk, \bfalpha, \mathbf{X}) + \log \p (\bfbeta, \bfk| \bfalpha) + \log \p (\bfalpha) \Bigg]   \Bigg\} \\
& \propto \exp \Bigg\{\bbe_{-\alpha_j} \Bigg[\log (\text{N}(\bfbeta|\mathbf{0}, \bfk \otimes \bfdelta)\text{W}(\bfv_0, \nu_0) + \log (\Gamma(\alpha_j|c,d)) \Bigg]   \Bigg\} \\
& \propto \exp \Bigg\{ \bbe_{-\alpha_j} \Bigg[ \log \Bigg( |\bfq|^{\frac{1}{2}} \exp \Bigg\{ - \frac{1}{2} \bfbeta^T \bfq \bfbeta \Bigg\} \Bigg) + \\
& ~~~~ \log \Bigg( |\bfk|^{\frac{\nu_0 - 3}{2}} \exp \Bigg\{ -\frac{1}{2 }\tr \bfv_0^{-1}\bfk \Bigg\}\Bigg) +  \log \Bigg( \alpha_j^{c_{a0-1}} \exp \Bigg\{ \alpha_j d  \Bigg\} \Bigg)  \Bigg] \Bigg\} \\
& \propto\! \exp \!\Bigg\{\! \bbe_{-\alpha_j}\! \Bigg[\! \frac{1}{2} \log |\bfk|^p|\bfdelta|^2\! -\! \frac{1}{2} \bfbeta^T \bfq \bfbeta \! +\! \frac{\nu_0\! -\! 3}{2} \log |\bfk|\!  -\!\frac{1}{2 }\tr \bfv_0^{-1}\bfk   + \\
& ~~~~(\!c\!-\!1\!) \!\log (\alpha_j) \! - \!\alpha_j d   \!\Bigg]\!  \Bigg\} \\
& \propto \exp \Bigg\{  \log |\bfdelta|  - \frac{1}{2} \bbe_{-\alpha_j} [\bfbeta^T \bfq \bfbeta]    + (c-1) \log (\alpha_j) -\alpha_j d \Bigg\} \\
& \propto \exp \Bigg\{  \log (\alpha_j)  - \frac{1}{2} \bbe_{-\alpha_j} [\bfbeta^T \bfq \bfbeta]    + (c-1) \log (\alpha_j) -\alpha_j d \Bigg\} \\
& \propto \exp \Bigg\{  (c+1 -1) \log (\alpha_j)  - \frac{1}{2} \bbe_{-\alpha_j} [\bfbeta^T \bfq \bfbeta] -\alpha_j d \Bigg\} \\
\end{split} 
\end{equation}

As in Equation \ref{eq:q_beta_k_2}, the terms that did not relate to $\alpha_j$ were disregarded, and to find $\bbe_{-\alpha_j} [\bfbeta^T \bfq \bfbeta ]$ the following calculations were made:

\begin{comment}

$$\bbe_{- \alpha_j}[\bfbeta^T \bfq \bfbeta] =\bbe_{- \alpha_j}\left[ \left[\beta_{a1} \dots \beta_{ap} \beta_{v1} \dots \beta_{vp} \right]_{1 \times 2p} \left [ \begin{array}{c:c}

\tau_a \bfdelta & \kappa \bfdelta\\  \hdashline
 \kappa \bfdelta & \tau_v \bfdelta \\
\end{array} \right] _ {2p\times2p} 
\left [ \begin{array}{c}
\beta_{a1} \\
\vdots\\
\beta_{ap}\\
\beta_{v1}\\
\vdots \\
\beta_{vp}\end{array}  \right]_{2p \times 1}\right]$$

$$\bfbeta^T \bfq \bfbeta = \left[\beta_{a1} \dots \beta_{ap} \beta_{v1} \dots \beta_{vp} \right]_{1 \times 2p} \left [ \begin{array}{cccccc}

\tau_a\alpha_{1} & & & \kappa\alpha_{1} & \\
& \ddots & & & \ddots\\
& & \tau_a\alpha_p & & &\kappa\alpha_p\\
\kappa\alpha_{1} & & & \tau_v\alpha_{1} & \\
& \ddots & & & \ddots\\
& & \kappa\alpha_p & & & \tau_v\alpha_p\\
\end{array}  \right]_{2p \times 2p}
\left [ \begin{array}{c}
\beta_{a1} \\
\vdots\\
\beta_{ap}\\
\beta_{v1}\\
\vdots \\
\beta_{vp}\end{array}  \right]_{2p \times 1}$$

$$\bfbeta^T\bfq\bfbeta = \left[ \begin{array}{c} 
\beta_{a1}\tau_a\alpha_1 + \beta_{v1}\kappa\alpha_1 \\
\vdots \\
\beta_{ap}\tau_a\alpha_p + \beta_{vp}\kappa\alpha_p \\
\beta_{a1}\kappa\alpha_1 + \beta_{v1} \tau_v\alpha_1 \\
\vdots \\
\beta_{ap}\kappa\alpha_p + \beta_{vp}\tau_v\alpha_p \end{array} \right]^T_{2p \times 1} \left [ \begin{array}{c}
\beta_{a1} \\
\vdots\\
\beta_{ap}\\
\beta_{v1}\\
\vdots \\
\beta_{vp}\end{array}  \right]_{2p \times 1}$$

\end{comment}

\begin{equation} 
\begin{split}
\bbe_{- \alpha_j}[\bfbeta^T \bfq \bfbeta]& =\bbe_{- \alpha_j}\left[ \left[\beta_{a1} \dots \beta_{ap} \beta_{v1} \dots \beta_{vp} \right]_{1 \times 2p} \begin{bmatrix} \tau_a \bfdelta & \kappa \bfdelta\\ 
 \kappa \bfdelta & \tau_v \bfdelta\\   \end{bmatrix}_ {2p\times2p} \begin{bmatrix}\beta_{a1} \\
\vdots\\
\beta_{ap}\\
\beta_{v1}\\
\vdots \\
\beta_{vp} \end{bmatrix}_{2p \times 1} \right]\\
& = \bbe_{- \alpha_j}\Bigg[
\beta_{a1}^2\tau_a\alpha_1 + \beta_{a1}\beta_{v1}\kappa\alpha_1 + \dots + \beta_{ap}^2\tau_a\alpha_p + \beta_{ap}\beta_{vp}\kappa\alpha_p \\
& ~~~~ + \beta_{v1}\beta_{a1}\kappa\alpha_1 + \beta_{v1}^2 \tau_v\alpha_1 + \dots 
+ \beta_{vp}\beta_{ap}\kappa\alpha_p + \beta_{vp}^2\tau_v\alpha_p \Bigg] \\
& = \bbe_{- \alpha_j}\Bigg[\tau_a\sum_{j=1}^p \beta_{aj}^2\alpha_j + 2\kappa \sum_{j=1}^p \alpha_j\beta_{aj}\beta_{vj} +\tau_v\sum_{j=1}^p \alpha_j\beta_{vj}^2 \Bigg]\\
& = \alpha_j(\bbe_{- \alpha_j}[\beta_{aj}^2\tau_a] + 2 \bbe_{- \alpha_j}[\beta_{aj}\beta_{vj}\kappa] + \bbe_{- \alpha_j}[\beta_{vj}^2\tau_v]) \\
\end{split}
\end{equation}

By discarding whatever does not relate to the variable of interest $\alpha_j$ we can explore the conditional expected value and linearity properties to further the calculations.
\begin{equation} \nonumber
\begin{split}
\bbe_{- \alpha_j}[\bfbeta^T\bfq\bfbeta]  & = \alpha_j(\bbe_{\bfk}[\bbe_{- \alpha_j}[\beta_{aj}^2\tau_a|\bfk]] + 2 \bbe_{\bfk}[\bbe_{- \alpha_j}[\beta_{aj}\beta_{vj}\kappa|\bfk]] + \bbe_{\bfk}[\bbe_{- \alpha_j}[\beta_{vj}^2\tau_v|\bfk]]) \\
 & = \alpha_j(\bbe_{\bfk}[\tau_a\bbe_{- \alpha_j}[\beta_{aj}^2|\bfk]] + 2 \bbe_{\bfk}[\bbe_{- \alpha_j}\kappa[\beta_{vj}\beta_{aj}|\bfk]] + \bbe_{\bfk}[\tau_v\bbe_{- \alpha_j}[\beta_{vj}^2|\bfk]]) \\
\end{split}
\end{equation}

Note that the expected value with respect to the variational distribution considers the variational distribution of $\bfbeta$. We know that $\bbe[\beta_d^2|\bfk] = \bbe^2[\beta_j|\bfk] + \var[\beta_j|\bfk]$ for both Arousal and Valence and  $\text{Cov}(\beta_{vj},\beta_{aj}|\bfk) = \bbe[\beta_{vj}\beta_{aj}|\bfk] - \bbe[\beta_{vj}|\bfk]\bbe[\beta_{aj}|\bfk]$. From now on, this equation will be split into three parts, in order to improve its understanding.
\begin{equation} \label{eq:expt_alpha}
\begin{split}
\bbe_{- \alpha_j}[\bfbeta^T\bfq\bfbeta] & = \alpha_j(\bbe_{\bfk}[\tau_a\bbe_{- \alpha_j}[\beta_{aj}|\bfk]^2 + \var_{- \alpha_j}(\beta_{aj}|\bfk)] \\
& ~~~~ + 2 \bbe_{\bfk}[\kappa \bbe_{- \alpha_j}[\beta_{vj}\beta_{aj}|\bfk]]\\
& ~~~~ + \bbe_{\bfk}[\tau_v\bbe_{- \alpha_j}[\beta_{vj}|\bfk]^2 + \var_{- \alpha_j}(\beta_{vj}|\bfk)]) \\
\end{split}
\end{equation}

Now let us develop some calculations to work the expected values on Equation \ref{eq:expt_alpha}. Note that $\bfk_* = \bfj + \bfq_* = \bfk \otimes \bfx^T\bfx + \bfk \otimes \bfdelta_* = \bfk \otimes (\bfx^T\bfx + \bfdelta_*)$. Since it will be a recurring value throughout the remaining of this work, let us attribute $\bfm_* = \bfx^T\bfx + \bfdelta_*$ 

\vspace{5mm}

Also note that calculating $\bfbeta_*$ does not involve $\bfk$:

\begin{equation} \label{eq:expt_alpha_ii}
\begin{split}
\bfbeta^* & = [\bfk \otimes \bfm_*]^{-1} [\bfk \otimes \bfx^T\bfx] \bfhbeta \\
& = [\bfk^{-1} \otimes \bfm_*^{-1}] [\bfk \otimes \bfx^T\bfx] \bfhbeta \\
& = [\bfi \otimes (\bfm_*^{-1} \bfx^T\bfx)] \bfhbeta \\
& = [\bfi \otimes ((\bfx^T\bfx + \bfdelta_*)^{-1} \bfx^T\bfx)] \vect((\bfx^T\bfx)^{-1}\bfx^T\bfy) \\
& = [\bfi \otimes ((\bfx^T\bfx + \bfdelta_*)^{-1} \bfx^T\bfx)] [\bfy^T \otimes (\bfx^T\bfx)^{-1} \vect (\bfx^T)] \\
& = [\bfy^T \otimes ((\bfm_*)^{-1} \bfx^T\bfx)] \vect (\bfx^T) \\
& = \vect((\bfm_*^{-1})\bfx^T\bfy)
\end{split}
\end{equation}

Therefore is constant with respect to $\bbe_\bfk$ and this allow us to continue this calculations with more simplicity.

\vspace{5mm}

So $\bfbeta$'s covariance matrix is $(\bfk_*)^{-1} = \left[ \begin{array}{c:c} 
(\bfk^{-1})_{11}(\bfm_*)^{-1} & (\bfk^{-1})_{12}(\bfm_*)^{-1}\\ \hdashline
(\bfk^{-1})_{21}(\bfm_*)^{-1} & (\bfk^{-1})_{22}(\bfm_*)^{-1}\end{array} \right]$,

\vspace{5mm}
\noindent what makes $\var_{-\alpha_j} (\beta_{aj}|\bfk)$ equals the $(j,j)$ entry of the first block in $(\bfk_*)^{-1}$ and $\var_{-\alpha_j} (\beta_{vj}|\bfk)$ equals the $(j,j)$ entry of the fourth block. So we can carry on Equation \ref{eq:expt_alpha} with
\begin{equation} \label{eq:expt_alpha_iii}
\begin{split}
\bbe_{- \alpha_j}[\bfbeta^T\bfq\bfbeta] & = \alpha_j(\bbe_{\bfk}[\tau_a( \beta_{aj}^{*2} + (\bfk^{-1})_{11}[(\bfm_*)^{-1}]_{jj})] \\
& ~~~~ + 2 \bbe_{\bfk}[\kappa( \cov_{- \alpha_j}(\beta_{aj},\beta_{vj}|\bfk) + \bbe_{- \alpha_j}[\beta_{aj}|\bfk]\bbe_{- \alpha_j}[\beta_{vj}|\bfk]]\\
& ~~~~ + \bbe_{\bfk}[\tau_v (\beta_{vj}^{*2} + (\bfk^{-1})_{22}[(\bfm_*)^{-1}]_{jj})]) \\
& = \alpha_j(\bbe_{\bfk}[\tau_a( \beta_{aj}^{*2} + (\bfk^{-1})_{11}[(\bfm_*)^{-1}]_{jj})] \\
& ~~~~ + 2 \bbe_{\bfk}[\kappa ((\bfk^{-1})_{12}[(\bfm_*)^{-1}]_{jj}] + \beta_{aj}^*\beta_{vj}^*)]\\
& ~~~~ +  \bbe_{\bfk}[\tau_v (\beta_{vj}^{*2} + (\bfk^{-1})_{22}[(\bfm_*)^{-1}]_{jj})] )\\
& = \alpha_j(\beta_{aj}^{*2} \bbe_{\bfk}[\tau_a]  + [(\bfm_*)^{-1}]_{jj}] \bbe_{\bfk}[\tau_a(\bfk^{-1})_{11}] \\
& ~~~~ + 2 [(\bfm_*)^{-1}]_{jj}] \bbe_{\bfk}[\kappa (\bfk^{-1})_{12}]  + \beta_{ad}^*\beta_{vj}^*\bbe_{\bfk}[\kappa] \\
& ~~~~ + \beta_{vj}^{*2} \bbe_{\bfk}[\tau_v]  + [(\bfm_*)^{-1}]_{jj}]\bbe_{\bfk}[\tau_v(\bfk^{-1})_{22}]] ).\\
\end{split}
\end{equation}

\vspace{5mm}

Now, to calculate $\bbe_{\bfk}[\tau_a], \bbe_{\bfk}[\tau_v], \bbe_{\bfk}[\kappa], \bbe_{\bfk}[\kappa(\bfk^{-1})_{12}], \bbe_{\bfk}[\tau_a(\bfk^{-1})_{11}]]$ and  $\bbe_{\bfk}[\tau_v(\bfk^{-1})_{22}]]$ we will need the following Definition and Theorem \cite{gup99}:

\begin{theo}{\textbf{Commutation matrix:}}
A commutation matrix $\bfg_{pq}$ is a $pq \times pq$ matrix that transforms $\vect(\mathbf{A})$ in $\vect(\mathbf{A})^T$, where $\mathbf{A}$ is a $p\times q$  matrix. $\bfg_{pq}$ is given by $\sum_{i=1}^p\sum_{j=1}^q \mathbf{H}_{ij} \otimes \mathbf{H}_{ij}^T$, where the $(i, j)$ entry of $\mathbf{H}_{ij}$ is $1$ and the others are $0$.
\end{theo}

\vspace{5mm}

\begin{theorem}{\textbf{Expectation value of Wishart distribution:}}
Let $\mathbf{S} \sim \text{W} (\Pi, u)$ and $\bfg_{qq}$ be a $q \times q$ commutation matrix. Then:
$$\bbe[\mathbf{S}_{ij}] = u \Pi_{ij}$$
$$\bbe[\mathbf{S}^{-1} \otimes \mathbf{S}] = \frac{u}{u-q-1} \Pi^{-1} \otimes \Pi - \frac{1}{u-q-1}[\vect(\bfi_q)\vect(\bfi_q)^T +\bfg_{qq}], u-q-1 >0 $$
\end{theorem}

Let us consider $\bfv_* = \begin{bmatrix} v_{11} & v_{22}\\ v_{21} & v_{22} \end{bmatrix}$. Applying this theorem to our case we have:

\begin{equation}
\begin{split}
\bbe[\bfk^{-1} \otimes \bfk] & = \frac{\nu_*}{\nu_*-2-1} ((\bfv_*)^{-1} \otimes \bfv_*) - \frac{1}{\nu_*-2-1}[\vect(\bfi_2)\vect(\bfi_2)^T +\mathbf{G}_{dd}].
\end{split}
\end{equation}

%\begin{equation}
%\begin{split}
%\bbe[\bfk^{-1} \otimes \bfk] & = \frac{\nu_*}{\nu_*-2-1} ((\bfv_*)^{-1} \otimes \bfv_*) - \frac{1}{\nu_*-2-1}[\vect(\bfi_2)\vect(\bfi_2)^T +\text{G}_{pp}]\\
%& = \frac{\nu_*}{\nu_* - 3} \frac{1}{v_{11}v_{22}-v_{12}v_{21}}  \begin{bmatrix} v_{22}v_{11} & v_{22}v_{12} & -v_{12}v_{11} & -v_{12}v_{12} \\ v_{22}v_{21} & v_{22}v_{22} & -v_{12}v_{21} &-v_{12}v_{22} \\ -v_{21}v_{11} & -v_{21}v_{12} & v_{11}v_{11} & v_{11}v_{12} \\ -v_{21}v_{21} & -v_{21}v_{22} & v_{11}v_{21}& v_{11}v_{22}\ \end{bmatrix} - \\ 
%& ~~~~ \frac{1}{\nu_*-3}\left[\begin{bmatrix} 1 \\0\\0\\1 \ \end{bmatrix} \begin{bmatrix} 1 & 0 & 0 & 1 \end{bmatrix} + \begin{bmatrix} 1 & 0 & 0 & 0 \\ 0 & 0 & 1 & 0 \\ 0 & 1 & 0 & 0 \\ 0 & 0 & 0 & 1 \ \end{bmatrix}  \right] \\
%& = \frac{\nu_*}{\nu_*-3} \frac{1}{v_{11}v_{22}-v_{12}v_{21}}  \begin{bmatrix} v_{22}v_{11} & v_{22}v_{12} & -v_{12}v_{11} & -v_{12}v_{12} \\ v_{22}v_{21} & v_{22}v_{22} & -v_{12}v_{21} &v_{12}v_{22} \\ -v_{21}v_{11} & -v_{21}v_{12} & v_{11}v_{11} & v_{11}v_{12} \\ -v_{21}v_{21} & -v_{21}v_{22} & v_{11}v_{21}& v_{11}v_{22}\ \end{bmatrix} - \\ 
%& ~~~~ \frac{1}{\nu_*-3} \begin{bmatrix} 2 & 0 & 0 & 1 \\ 0 & 0 & 1 & 0 \\ 0 & 1 & 0 & 0 \\ 1 & 0 & 0 & 2 \ \end{bmatrix}
%\end{split}
%\end{equation}

Collecting the values we need from the equations above we can fulfill the expected values below:
\begin{equation} \nonumber
\begin{split}
\bbe_{\bfk}[\tau_a] & = \nu_* (\bfv_*)_{11}\\
\bbe_{\bfk}[\tau_v] & = \nu_* (\bfv_*)_{22}\\
\bbe_{\bfk}[\kappa] & = \nu_* (\bfv_*)_{12}\\
\bbe_{\bfk}[\tau_a(\bfk^{-1})_{11}]] & = \bbe_{\bfk}[(\bfk)_{11}(\bfk^{-1})_{11}] = \frac{\nu_*}{\nu_*-3}\left(\bfv_*^{-1} \otimes \bfv_*\right)_{11} - \frac{2}{\nu_* - 3}\\
\bbe_{\bfk}[\tau_v(\bfk^{-1})_{22}]] & = \bbe_{\bfk}[(\bfk)_{22}(\bfk^{-1})_{22}] = \frac{\nu_*}{\nu_*-3}\left(\bfv_*^{-1} \otimes \bfv_*\right)_{22} - \frac{2}{\nu_* - 3}\\
\bbe_{\bfk}[\kappa(\bfk^{-1})_{12}] & = \bbe_{\bfk}[(\bfk)_{12}(\bfk^{-1})_{12}] = \frac{\nu_*}{\nu_*-3}\left(\bfv_*^{-1} \otimes \bfv_*\right)_{12} - \frac{1}{\nu_* - 3}\\
\end{split}
\end{equation}

With the information obtained in Equation \ref{eq:expt_alpha} we can continue Equation \ref{eq:q-alpha-model-ard-2} as
\begin{equation}
\begin{split}
\q^*(\alpha_j) & \propto \exp \Bigg\{ \log(\alpha_j)(c^*-1) - \alpha_j d^* \Bigg\},  \\
\end{split}
\end{equation}
where
\begin{equation} \nonumber
\begin{split}
c^* & = c + 1 \\
d_j^* & =  d + \frac{1}{2} \Bigg[ \beta_{aj}^{*2}\nu_* (\bfv_*)_{11} + [(\bfm_*)^{-1}]_{jj}]  \Bigg( \frac{\nu_*}{\nu_*-3}\left(\bfv_*^{-1} \otimes \bfv_*\right)_{11} - \frac{2}{\nu_* - 3}\Bigg)\\
& ~~~~ + 2 \Bigg[\beta_{aj}^*\beta_{vj}^*\nu_* (\bfv_*)_{12} + [(\bfm_*)^{-1}]_{jj}  \Bigg( \frac{\nu_*}{\nu_*-3}\left(\bfv_*^{-1} \otimes \bfv_*\right)_{12} - \frac{1}{\nu_* - 3} \Bigg)\Bigg]   \\
& ~~~~ + \beta_{vj}^{*2} \nu_* (\bfv_*)_{22}  + [(\bfm_*)^{-1}]_{jj}]  \Bigg(\frac{\nu_*}{\nu_*-3}\left(\bfv_*^{-1} \otimes \bfv_*\right)_{22} - \frac{2}{\nu_* - 3}\Bigg) \Bigg].\\
\end{split}
\end{equation}

With this method, we are able to maintain the variational family given and slightly modify the Algorithm \ref{alg:ACAVI} to implement CAVI in MARD:

\begin{algorithm}[H] \label{alg:MARD}
\SetAlgoLined

Input: A model $\p  (\bfbeta, \bfk, \bfalpha, \bfx)$ and a dataset $\bfx$, where the response is $\bfy \in \mathbb{R}^2 $
    
 Output: A variational density $\q^*(\bfbeta, \bfk, \bfalpha) = \q^*(\bfbeta, \bfk) \prod_{j=1}^p \q^*(\alpha_j)$
    
Initialize the variational factors $\q(\bfbeta, \bfk) \q_j(\alpha_j)$
    
\While{ELBO has not converged}{
    \begin{itemize}
        \item [i) ] Update $ c_{}^*  = c + 1$
        \item [ii) ] Update $\nu_* = \nu_0 + n +p$
        \item [iii) ] Update $\bfm_* = \bfx^T\bfx + \bfdelta_*$
        \item [iv) ] Update $\bfb* = (\bfm_*^{-1})\bfx^T\bfy$ 
        \item [v) ] Update $ \bfv_*^{-1}  = \bfv_0^{-1} + \bfs^T\bfs + \bfhb^T(\bfdelta_*^{-1} + \bfx^T\bfx^{-1})^{-1}\bfhb $
        \item [vi) ] Update $ d_{j}^* = d + \frac{1}{2} \Bigg[ \beta_{aj}^{*2}\nu_* (\bfv_*)_{11} + [(\bfm_*)^{-1}]_{jj}]  \Bigg( \frac{\nu_*}{\nu_*-3}\left(\bfv_*^{-1} \otimes \bfv_*\right)_{11} - \frac{2}{\nu_* - 3}\Bigg)+ 2 \Bigg[\beta_{aj}^*\beta_{vj}^*\nu_* (\bfv_*)_{12} + [(\bfm_*)^{-1}]_{jj}  \Bigg( \frac{\nu_*}{\nu_*-3}\left(\bfv_*^{-1} \otimes \bfv_*\right)_{14} - \frac{1}{\nu_* - 3} \Bigg)\Bigg] + \beta_{vj}^{*2} \nu_* (\bfv_*)_{44}  + [(\bfm_*)^{-1}]_{jj}]  \Bigg(\frac{\nu_*}{\nu_*-3}\left(\bfv_*^{-1} \otimes \bfv_*\right)_{22} - \frac{2}{\nu_* - 3}\Bigg) \Bigg]$, for all $j$
        \item [vii) ] Update $\bfdelta_* = \left [ \begin{smallmatrix} \frac{c_{*}}{d_{*}} \\ & \ddots \\
& & \frac{c_{*}}{d_{*}} \ \end{smallmatrix} \right] _ {p\times p}$
    \end{itemize}}
    
Return: $\q^*(\bfbeta, \bfk, \bfalpha)$
\caption{Adapted Coordinate Ascending Variational Inference.}
\end{algorithm}

\vspace{5mm}

Note that we will not calculate the ELBO as in Algorithm \ref{alg:ACAVI}, as it is much more difficult, but we will rely on the calculations provided and the theoretical evidence that guarantees convergence to a local maximum of the ELBO. The values obtained from this implementation will be considered converged when their precision reach $10^{-3}$.
\section{Predictive distribution}\label{sec:pred-dist}

%\bfb^T \bfx != \bfx\bfb
% \bfx \bfbeta -> nxp  2px1 -> incompatível
%Como \bdf = I_2 \otimes \bbfx^T?

In this chapter we worked through the calculations necessary to implement variational inference through CAVI in a multivariate generalization of ARD, taking in consideration the correlation between the responses. Since our model has a Normal likelihood, Normal-Wishart joint prior distribution and compatible variational family, our model is simple enough that it is possible to calculate the joint predictive distribution, that can be useful to perform more accurate predictions. We can also get a credible region, rather than two separate credibility intervals as we did in Chapter \ref{vi_ard}, achieving a narrower, more precise set of predictions. 

In order to do that, we need to approximate the predictive distribution $\p(\bbfy|\bfx)$:
\begin{equation} \label{eq:pred-1}
\begin{split}
\p(\bbfy|\bfx) & = \iint \p(\bbfy|\bfx, \bfbeta, \bfk)\p(\bfbeta, \bfk|\bfx) \der\bfbeta \der\bfk \der\bfalpha\\
& \approx\iint \p(\bbfy|\bfx, \bfbeta, \bfk)\q(\bfbeta, \bfk)\q(\bfalpha) \der\bfbeta \der\bfk \der\bfalpha \\
& = \iint \p(\bbfy|\bfx, \bfbeta, \bfk)\q(\bfbeta, \bfk) \der\bfbeta \der\bfk \\
& = \iint \text{N}(\bbfy|\bfb^T\bbfx, \bfk)\text{N}(\bfbeta|\bfbeta_*, \bfk \otimes \bfm_*)\text{W}(\bfk|\nu_*, \bfv_*)\der\bfbeta \der\bfk \\
\end{split}    
\end{equation}

Here it is worth stopping to comment on a few steps of Equation \ref{eq:pred-1}. The mean-field hypothesis aligned with $\bfalpha$'s absence in the likelihood allows this variable  to be integrated out.  Also note that $\bfb^T\bbfx = (\bbfx^T\bfb)^T$, so the notation in Equation \ref{eq:pred-1} does not clash with the distribution set in Section \ref{sec:found_mard}. Writing it in this manner provides a column vector that will enable us to carry on the calculations. We also need to replace $\bfb$ by some form of $\bfbeta$, as exploring a Multivariate Normal distribution can be more easily explored than a Matrix Normal distribution. If we define the matrix $\bfxi = \bfi_2 \otimes \bbfx^T$ ($\bfi_2$ being the $2\times2$ identity matrix) we can write $\bfb^T\bbfx$ as $\bfxi\bfbeta$ and continue as:

\begin{equation} \label{eq:pred-2}
\begin{split}
\p(\bbfy|\bfx) & = \iint \text{N}(\bbfy|\bfxi\bfbeta, \bfk)\text{N}(\bfbeta|\bfbeta_*, \bfk \otimes \bfm_*)\text{W}(\bfk|\nu_*, \bfv_*)\der\bfbeta \der\bfk \\
& = \int \Bigg[ \int \text{N}(\bbfy|\bfxi\bfbeta, \bfk)\text{N}(\bfbeta|\bfbeta_*, \bfk \otimes \bfm_*)\der\bfbeta\Bigg] \text{W}(\bfk|\nu_*, \bfv_*) \der\bfk \\
\end{split}
\end{equation}

To calculate the inner integral we resort to the following Theorem in \cite{Bis06}:

\begin{theorem}{\textbf{Convolution of Normal densities:}}
\label{theo:normal}
 Consider the following distributions:
\begin{equation} \nonumber
\begin{split}
    \p(\bbfx) &= \text{N}(\bbfx | \bm{\mu}, \bm{\psi}) \\
    \p(\bbfy | \bbfx) &= \text{N}(\bbfy | \textbf{A}\bbfx + \textbf{b}, \textbf{L}).
\end{split}
\end{equation}

The marginal distribution of $\bbfy$, given by $ \int \p(\bbfy, \bbfx)~\der \bbfx = \int \p(\bbfy | \bbfx)\p(\bbfx)~\der\bbfx$ is written as
\begin{equation} \nonumber
\begin{split}
    \p(\bbfy) = \text{N}(\bbfy | \textbf{A}\bm{\mu} + \textbf{b}, (\textbf{L}^{-1} + \textbf{A}\bm{\psi}^{-1}\textbf{A}^T)^{-1}).
\end{split}
\end{equation}
\end{theorem}

Adapting Theorem \ref{theo:normal} to our problem, we carry on Equation \ref{eq:pred-2} as:

\begin{equation} \label{eq:pred-3}
\begin{split}
\p(\bbfy | \bbfx) & = \int \Bigg[ \int \text{N}(\bbfy|\bfxi\bfbeta, \bfk)\text{N}(\bfbeta|\bfbeta_*, \bfk \otimes \bfm_*)\der\bfbeta\Bigg] \text{W}(\bfk|\nu_*, \bfv_*) \der\bfk \\
& = \int  \text{N}(\bbfy|\bfxi\bfbeta_*, (\bfk^{-1} + \bfxi(\bfk \otimes \bfm_*)^{-1}\bfxi^T)^{-1}) \text{W}(\bfk|\nu_*, \bfv_*) \der\bfk \\
\end{split}    
\end{equation}

In order to write \ref{eq:pred-3} as a Normal-Wishart distribution let us develop the Normal's precision:
\begin{equation} \nonumber
\begin{split}
\bfk^{-1} + \bfxi(\bfk \otimes \bfm_*)^{-1}\bfxi^T & = \bfk^{-1} + (\bfi_2 \otimes \bbfx^T)(\bfk \otimes \bfm_*)^{-1}(\bfi_2 \otimes \bbfx^T)^T \\
& = \bfk^{-1} + (\bfi_2 \otimes \bbfx^T)(\bfk \otimes \bfm_*)^{-1}(\bfi_2 \otimes \bbfx) \\
& = \bfk^{-1} + (\bfi_2 \otimes \bbfx^T)(\bfk^{-1} \otimes \bfm_*^{-1}\bbfx)\\
& = \bfk^{-1} +  \bfk^{-1} \otimes \bbfx^T\bfm_*^{-1}\bbfx\\
& = \bfk^{-1} + (\bbfx^T\bfm_*^{-1}\bbfx)\bfk^{-1}\\
& = [1 + \bbfx^T\bfm_*^{-1}\bbfx]\bfk^{-1},\\
\end{split}
\end{equation}
where the penultimate equality is due to the fact that $\bbfx^T\bfm_*\bbfx$ is a real number and not a matrix, rendering the Kronecker product unnecessary. Continuing Equation \ref{eq:pred-3}:
\begin{equation} \label{eq:pred-4}
\begin{split}
\p(\bbfy | \bbfx) & = \int  \text{N}(\bbfy|\bfxi\bfbeta_*, (\bfk^{-1} + \bfxi(\bfk \otimes \bfm_*)^{-1}\bfxi^T))^{-1} \text{W}(\bfk|\nu_*, \bfv_*) \der\bfk \\
& = \int  \text{N}(\bbfy|\bfxi\bfbeta_*, ([1 + \bbfx^T\bfm_*^{-1}\bbfx]\bfk^{-1})^{-1}) \text{W}(\bfk|\nu_*, \bfv_*) \der\bfk. \\
\end{split}
\end{equation}

Now, let us work on the marginalization of $\bfk$. To make notation easier, let us consider $1 + \bbfx^T\bfm_*\bbfx = \varphi$. Since $\varphi$ is a number, $(\varphi \bfk^{-1})^{-1} = \varphi^{-1}\bfk$. From Equation \ref{eq:pred-4} we have:
\begin{equation} \nonumber
\begin{split}
\p(\bfk) & = \frac{1}{2^{\nu_* d/2| \bfv_*|^{\nu_*/2}} \Gamma_d(\nu_*/2)} |\bfk|^{(\nu_* - d - 1)/2} \exp\left(-\frac{1}{2}\tr(\bfv_*^{-1}\bfk)\right) \\
\p(\bfbeta | \bfk) & = \frac{|\varphi^{-1} \bfk|^{1/2}}{(2 \pi)^{d/2}} \exp\left(-\frac{1}{2}(\bfbeta - \bfxi\bfbeta_*)^T \varphi^{-1} \bfk (\bfbeta - \bfxi\bfbeta_*)\right), \end{split}    
\end{equation}
where $d=2$ in our case. The marginalization can be computed as:
\begin{equation} \label{marg-1}
\begin{split}
 \p(\bfbeta) & = \int \p(\bfbeta | \bfk) \p(\bfk) \der\bfk \\
& \propto \int \left[|\bfk|^{(\nu_* - d - 1)/2} \exp\left(-\frac{1}{2}\tr(\bfv_*^{-1}\bfk)\right)\right]\\
& ~~~~ \times \left[|\varphi^{-1} \bfk|^{1/2} \exp\left(-\frac{1}{2}(\bfbeta - \bfxi\bfbeta_*)^T \varphi^{-1} \bfk (\bfbeta - \bfxi\bfbeta_*)\right)\right]\der\bfk \\
 & =    \int  \left[ |\bfk|^{(\nu_* - d - 1)/2}  \exp \left( -\frac{1}{2}\tr(\bfv_*^{-1}\bfk)  \right) \right]\\
& ~~~~ \times   \left[ |\varphi^{-1} \bfk|^{1/2}  \exp \left( -\frac{1}{2} \tr(\varphi^{-1}(\bfbeta  -  \bfxi\bfbeta_*) (\bfbeta  -  \bfxi\bfbeta_*)^T  \bfk)  \right) \right] \der\bfk \\
 &= \int \varphi^{-d/2} |\bfk|^{(\nu_* - d - 1)/2 + 1/2} \exp\left(-\frac{1}{2}\tr[(\bfv_*^{-1} + \varphi^{-1}(\bfbeta - \bfxi\bfbeta_*) (\bfbeta - \bfxi\bfbeta_*)^T)\bfk]\right)\der\bfk \\
 &\propto \int |\bfk|^{(\nu_* + 1 - d - 1)/2} \exp\left(-\frac{1}{2}\tr[(\bfv_*^{-1} + \varphi^{-1}(\bfbeta - \bfxi\bfbeta_*) (\bfbeta - \bfxi\bfbeta_*)^T)\bfk]\right)\der\bfk.
\end{split}
\end{equation}

Note that this last line on Equation \ref{marg-1} is the kernel of a Wishart distribution with parameters $\nu_* + 1$ and $(\bfv_*^{-1} + \varphi^{-1}(\bfbeta - \bfxi\bfbeta_*) (\bfbeta - \bfxi\bfbeta_*)^T)^{-1}$, being the integral equal to the inverse of the respective normalizing constant:
\begin{equation} \label{eq:marg-2}
\begin{split}
\p( \bfbeta) &\propto \int |\bfk|^{(\nu_* + 1 - d - 1)/2} \exp\left(-\frac{1}{2}\tr[(\bfv_*^{-1} + \varphi^{-1}( \bfbeta -  \bfxi\bfbeta_*) ( \bfbeta -  \bfxi\bfbeta_*)^T)\bfk]\right)\der\bfk \\
&= 2^{(\nu_* + 1)d/2} |(\bfv^{-1} + \varphi^{-1}( \bfbeta -  \bfxi\bfbeta_*) ( \bfbeta -  \bfxi\bfbeta_*)^T)^{-1}|^{(\nu_* + 1)/2} \Gamma_d((\nu_* + 1)/2) \\
&\propto |\bfv_*^{-1} + \varphi^{-1}( \bfbeta -  \bfxi\bfbeta_*) ( \bfbeta - \bfxi\bfbeta_*)^T|^{-(\nu_* + 1)/2}.
\end{split}
\end{equation}

In order to carry on, let us explore the following relation: $|\textbf{A} + \bfv\bfv^T| = |\textbf{A}|(1 + \bfv^T\textbf{A}^{-1}\bfv)$. Therefore:
\begin{equation} \label{eq:marg-3}
\begin{split}
\p(\bfbeta) &\propto |\bfv_*^{-1} + \varphi^{-1}(\bfbeta - \bfxi\bfbeta_*) (\bfbeta - \bfxi\bfbeta_*)^T|^{-(\nu_* + 1)/2} \\
&\propto |\bfv_*^{-1}|^{-(\nu_* + 1)/2}[1 + \varphi^{-1} (\bfbeta - \bfxi\bfbeta_*)^T \bfv_* (\bfbeta - \bfxi\bfbeta_*)]^{-(\nu_* + 1)/2} \\
&\propto [1 + \varphi^{-1} (\bfbeta - \bfxi\bfbeta_*)^T \bfv_* (\bfbeta - \bfxi\bfbeta_*)]^{-(\nu_* + 1)/2}.
\end{split}
\end{equation}
This expression is similar to the kernel of a multivariate $t$-distribution, so with some algebra we can arrive at:
\begin{equation} \label{eq:marg-4}
\begin{split}
\p(\bfbeta) &\propto [1 + \varphi^{-1} (\bfbeta - \bfxi\bfbeta_*)^T \bfv_* (\bfbeta - \bfxi\bfbeta_*)]^{-(\nu_* + 1)/2} \\
& = \left[1 + \frac{\nu_* - d + 1}{\nu_* - d + 1} \varphi^{-1} (\bfbeta - \bfxi\bfbeta_*)^T \bfv_* (\bfbeta - \bfxi\bfbeta_*)\right]^{-((\nu_* - d + 1) + d)/2} \\
& = \left[1 + \frac{1}{\nu_* - d + 1} (\bfbeta - \bfxi\bfbeta_*)^T (\varphi^{-1} (\nu_* - d + 1) \bfv_*) (\bfbeta - \bfxi\bfbeta_*)\right]^{-((\nu_* - d + 1) + d)/2},
\end{split}
\end{equation}
which can be recognized as a multivariate $t$-distribution with $\nu_* - d + 1$ degrees of freedom ($\nu_* -2 +1$ in our case), centered at $\bfxi\bfbeta_*$ and scale matrix given by $ \frac{ \varphi^{-1}}{\nu_* - d + 1} \bfv_*^{-1}$.

Therefore, we can write Equation \ref{eq:pred-4} as:
\begin{equation} \label{eq:pred-5}
\begin{split}
\p(\bbfy | \bfx) &= \int \text{N}(\bbfy | \bfxi\bfbeta_*, [1 + \bbfx^T\bfm_*\bbfx) \text{W}(\bfk| \nu_*, \bfv_*)\der\bfk\\
& = t\left(\bbfy \left| \nu_* - 2 + 1, \bfxi\bfbeta_*, \left[\frac{(1 + \bbfx^T\bfm_*\bbfx)^{-1}}{\nu_* - 2 + 1}\right] \bfv_*^{-1} \right.\right).
\end{split}
\end{equation}

Since this distribution has a closed form and is from a known family of distributions, it is easy to sample using Algorithm \ref{alg:sample-mard} from and allows us to calculate approximated credible regions for each prediction. 

\begin{algorithm}[H] \label{alg:sample-mard}
\SetAlgoLined

For $i = 1, \dots, \text{Niter}$:
    
\begin{itemize}
    \item [i) ] Sample $\bfk_{(i)} \sim \text{W}(\bfk | \nu_*, \bfv_*)$
        
    \item [ii) ] Sample $\bfbeta_{(i)} | \bfk_{(i)} \sim \text{N}(\bfbeta |\bfbeta_*, (\bfk_{(i)} \otimes \bfm_*)^{-1})$
        
    \item [iii) ] Sample $\bbfy_{(i)} |\bfbeta_{(i)}, \bfk_{(i)} \sim \text{N}(\bbfy | \bfb_{(i)}^T\bbfx, \bfk_{(i)}^{-1})$, being $\bfbeta_{(i)} = \mathrm{\vect}(\bfb_{(i)})$
\end{itemize}
    
Return: From the set $(\bbfy_{(i)},\bfbeta_{(i)}, \bfk_{(i)})_{i = 1, \dots, \text{Niter}}$ of samples from the joint distribution of $\bbfy$, $\bfbeta$ and $\bfk$, keep only $(\bbfy_{(i)})_{i = 1, \dots, \text{Niter}}$.
\caption{Sampling from MARD's predictive distribution}
\end{algorithm}

\section{Testing MARD in a controlled setting}\label{sec:mard-test}

To access MARD's overall performance, it was firstly tested in a controlled scenario. The dummy dataset built for this purpose will have $2$ responses and $p=100$ features, where only $20$ differ from $0$. These features will be stored in a matrix $\bfb_{p \times 2}$.
The input data $ x_i \in \mathbb{R}^p$ will follow a multivariate normal distribution with mean $[0~0]$ and identity covariance matrix, for $i = 1,\dots, n$, and is stored in the matrix $ \bfx_{n \times p}$.
In this scenario, the generated errors $\varepsilon_i$ are normally distributed with mean zero and covariance matrix $\bfk^{-1}$, for $i = 1, \dots, n$ and stored in $\bfe_{n, 2}$. 
Two choices will be made for $K^{-1}$: first as a multiple of the identity matrix for uncorrelated responses, then $\left[\begin{array}{cc}
     100 & 85  \\
     85 & 100
\end{array} \right]$ for a scenario closer to ours, with a $0.57$ correlation for the responses.
The observed data will be $\bfy = \bfx\bfb + \bfe$ and split between training sets of size $n= 1000, 500, 100$ and test sets of size $n=1000$. 
Unfortunately, due to the nature of MARD's algorithm, we cannot work with datasets features $n<p$ in this format, as CAVI still requires inverting the matrix $(\bfx^T\bfx)$, only possible if $n\geq p$. This does not exclude the possibility of implementing MARD in a way where this operation in not necessary, but we will refrain from exploring this here.

\begin{figure}[H]
    \centering
        \subfloat{\includegraphics[width=0.5\textwidth]{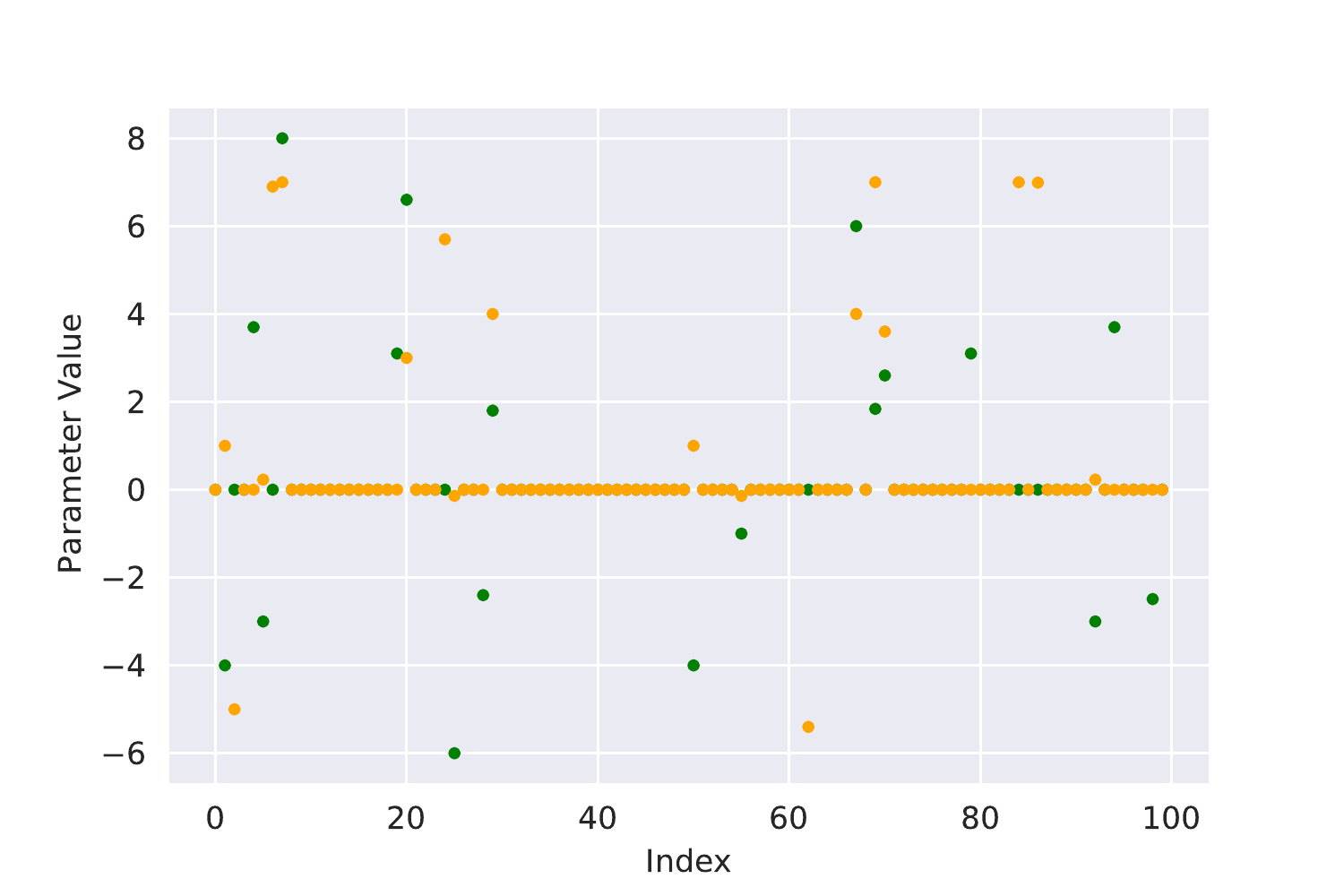}}
        \subfloat{\includegraphics[width=0.5\textwidth]{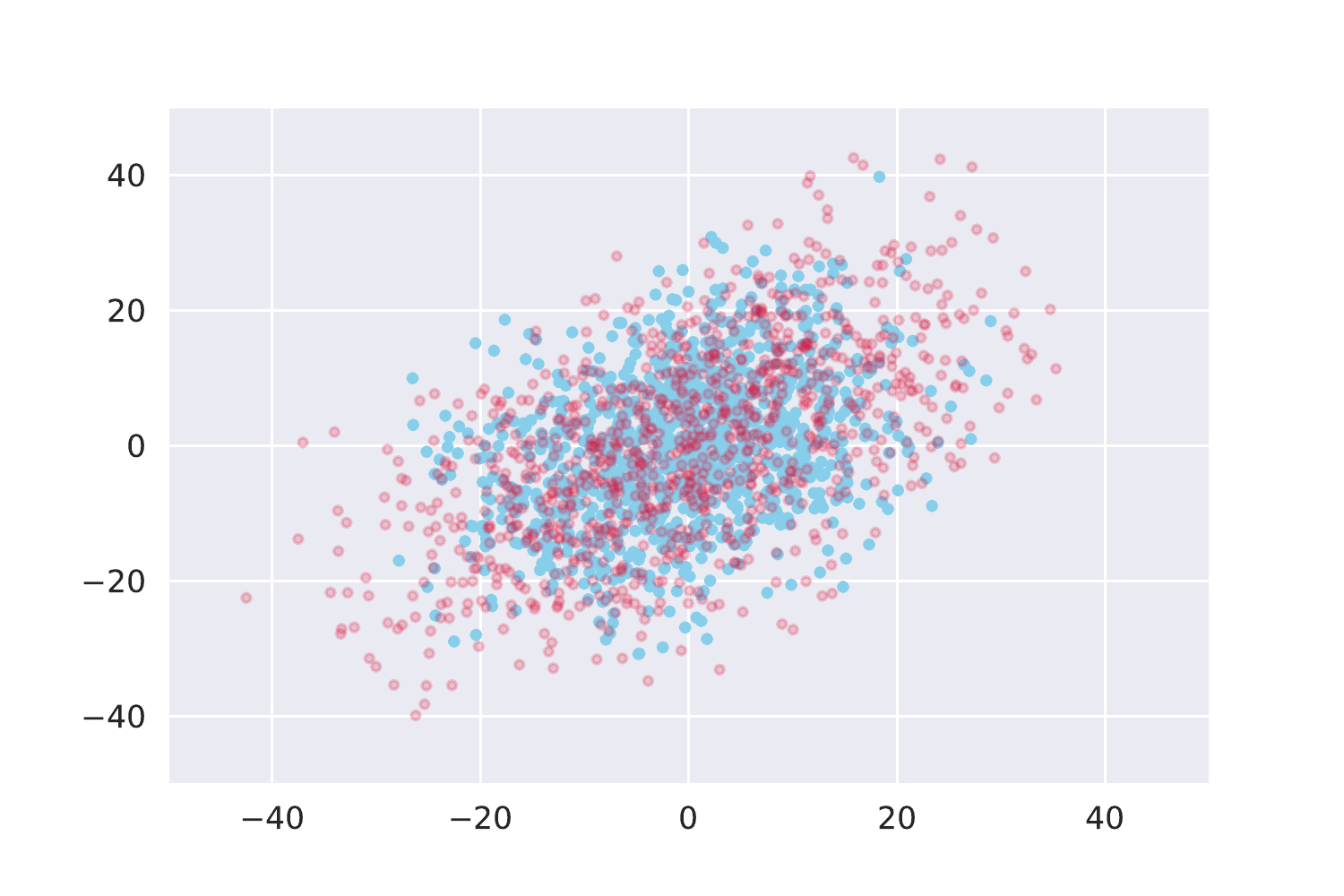}}
        \caption[Manufactured dataset display of parameters and the effects of the errors in the observations (right)]{Manufactured dataset display of parameters with one response variable in blue and the other in orange (left) and the effects of the errors in the observations (right).}
        \label{fig:fake_dataset}
\end{figure}

We provide a visual elaboration of the dataset in Figure \ref{fig:fake_dataset}. On the left we can see the parameters $\bfb$ divided in two colors, one for each corresponding response. On the right we can see how the errors affect the data's dispersion. $\bfx\bfb$ are displayed in blue, versus $\bfx\bfb + \bfe$ in pink and hope to observe how well the model will interpret this behavior.

Now let us analyse how well each model calculated the coefficients for each size of dataset. For $n=1000$, both perform similarly, with standard deviations of around $0.25$ for MARD and $0.27$ for ARD. For $n=500$ and $100$, the error greatly increases, with standard deviations reaching around $0.4$ for both models, than $2.8$ for MARD and $1.8$ for ARD, meaning the latter performed better when faced with less data.

\begin{figure}[H]
    \centering
    \includegraphics[width=1.1\textwidth]{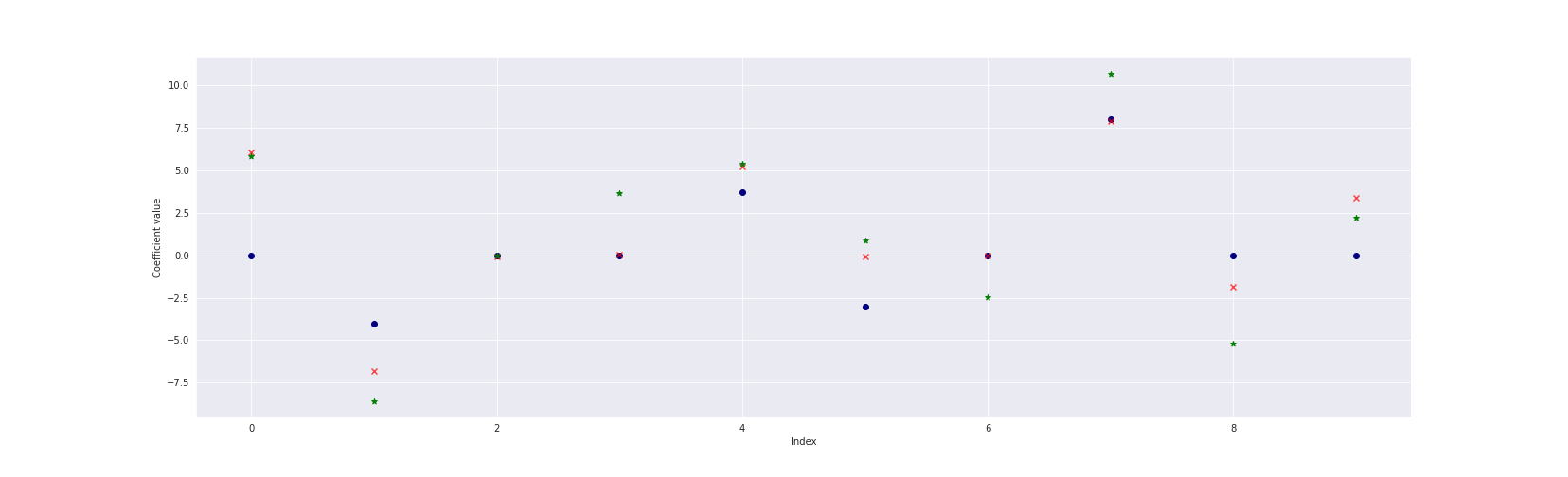}
        \caption{ARD and MARD performances in predicting coefficients for $n=1000, 500, 100$, respectively..}
        \label{fig:test-fake_dataset}
        
\end{figure}

In Figure \ref{fig:test-fake_dataset} we provide a visual aid for the information stade above. Each image displays a excerpt of $10$ coefficients and it is easy to see how the predictions stray from the targets with each decrease in the size of the training data. In navy blue the real values are displayed, versus MARD's green stars and ARD's red crosses. All result are displayed for the coefficients corresponding a single response, which is reasonable considering that, unlike the DEAM dataset, this responses have equal modeling difficulties. For $n=1000$ both show similar reasonable results and in the second image we can see that for $n=500$ the results are not as accurate, and it is not clear which model performed better. Finally, with $n=100$ predictions are very scattered and it is evident we did not provide enough data to train either model. 

% We have also calculated the RMSE for each case and, for the coefficients shown, there is no real distinction between ARD and MARD in the first two cases and a more significant gap in the last, with ARD showing some superiority over MARD as we can see in Table \ref{table:results_dummy_coefficients}. This has to do with the fact that MARD is being implemented through CAVI, and having problems dealing with $p \approx n$.

% \begin{table}[H]
% \begin{center}
% \begin{tabular}{||c c c c ||} 
% \hline
% \textbf{n} & 5000 & 500 & 100  \\ [0.5ex] 
%  \hline\hline
% ARD & $0.22$ & $0.56$ & $1.61$\\
% \hline
% MARD & $0.20$ & $0.55$ & $32$ \\[1ex]
% \hline
%  \hline
% \end{tabular}
% \end{center}
% \caption{Comparison between RMSE of coefficient predictions for three different dataset sizes achieved by ARD and MARD.}
% \label{table:results_dummy_coefficients}
% \end{table}

\begin{figure}[H]
    \centering
    \includegraphics[width=1.1\textwidth]{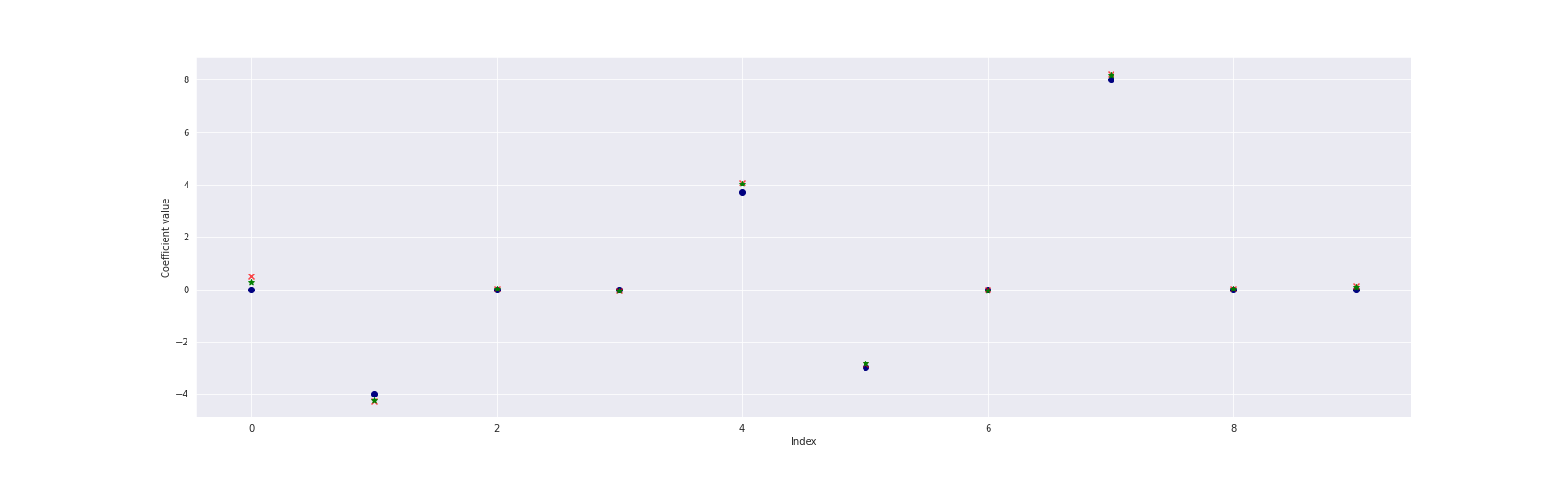}
\end{figure}

\begin{figure}[H]
    \centering
        \includegraphics[width=1.1\textwidth]{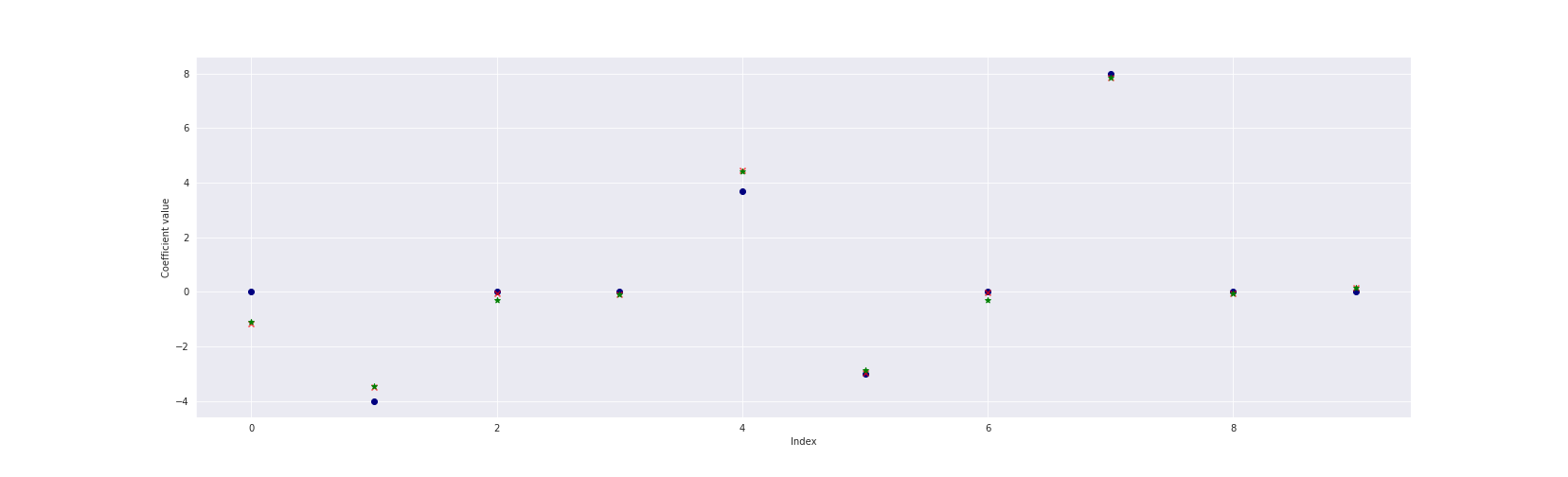}    
\end{figure}

\begin{figure}[H]
    \centering
        \subfloat{\includegraphics[width=0.5\textwidth]{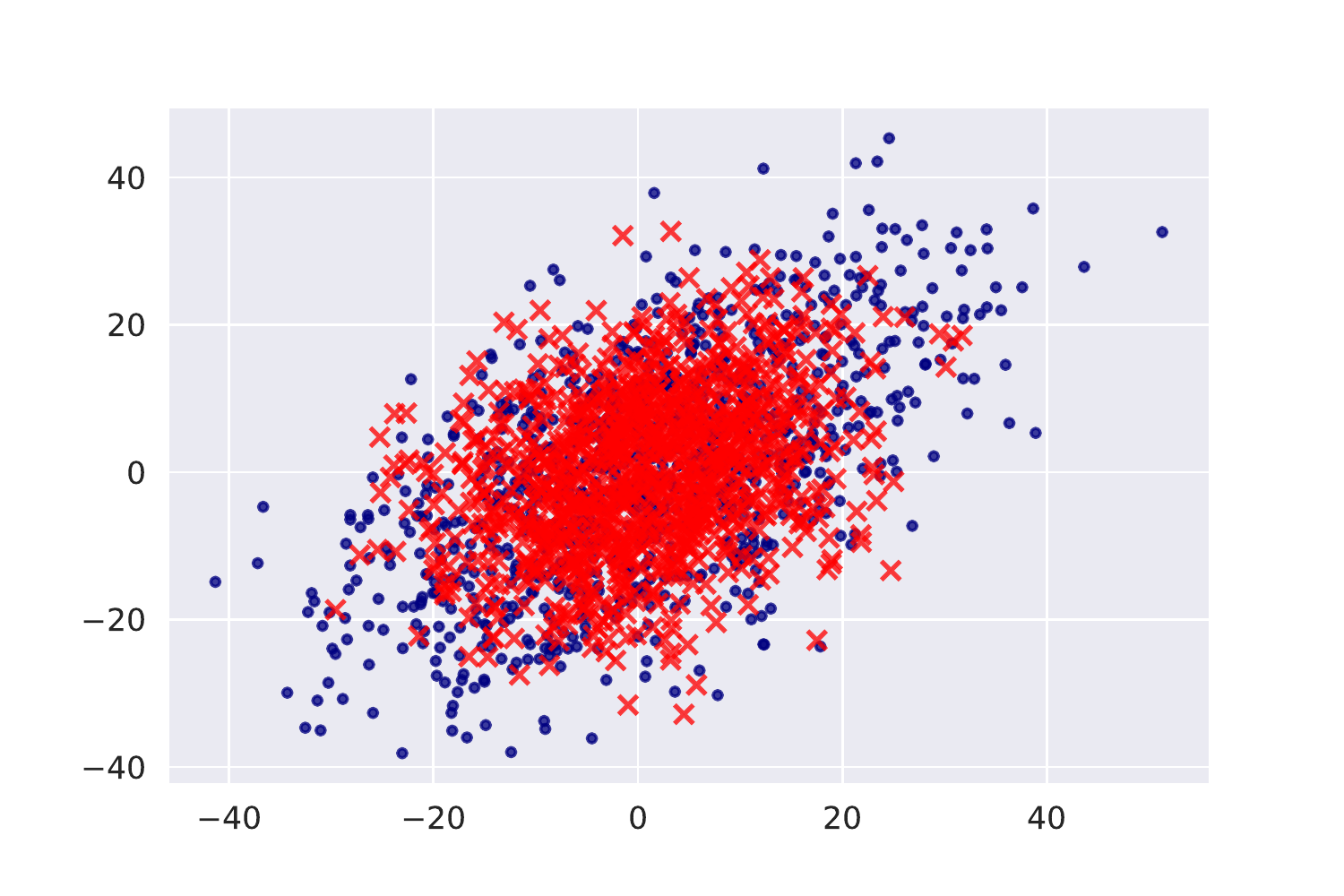}}
        \subfloat{\includegraphics[width=0.5\textwidth]{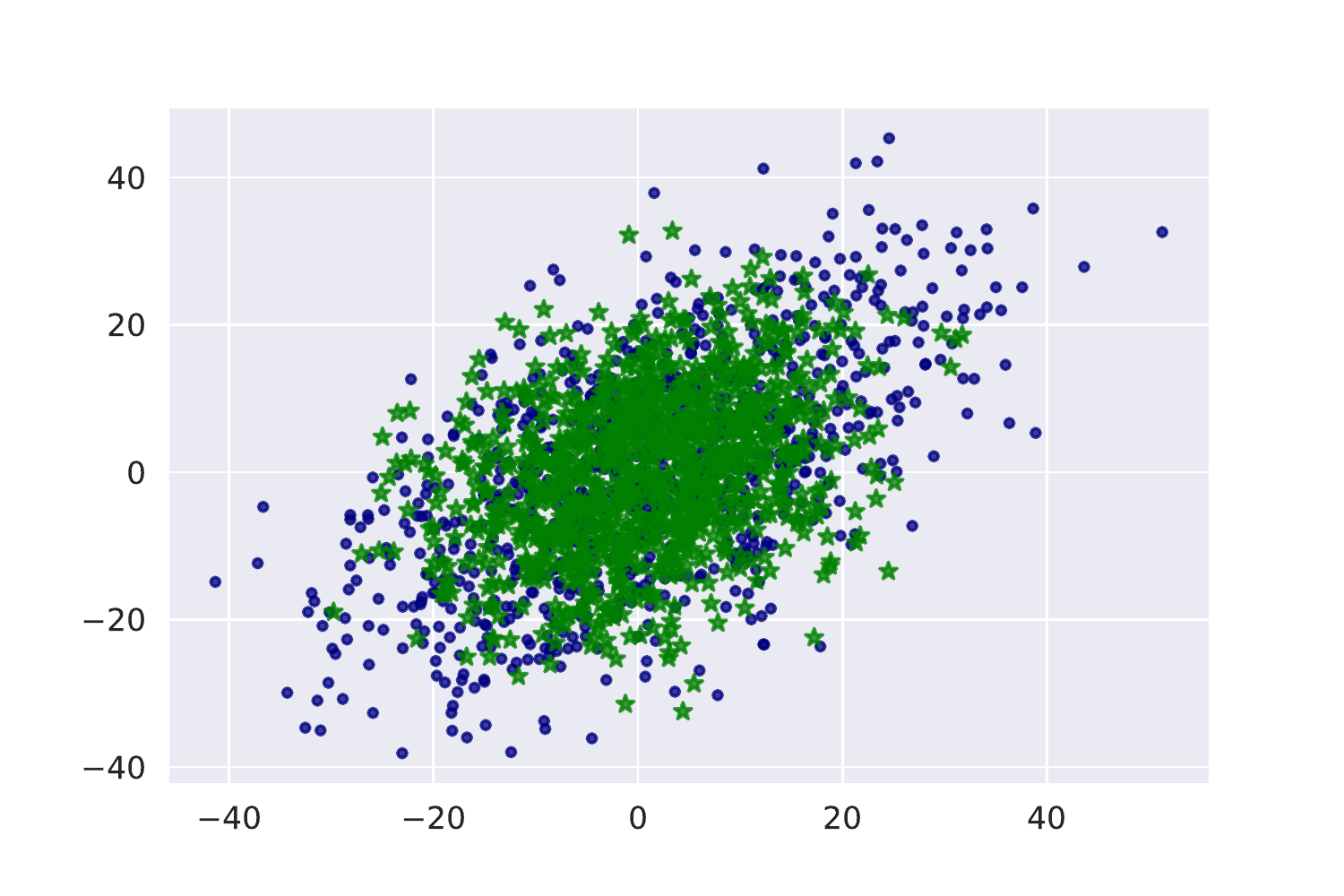}}
        \caption{Comparison between predictions achieved by ARD and MARD  for $\bfy$ with $n=500$.}
        \label{fig:test2-fake_dataset}
\end{figure}

Coefficients aside, let us analyse the predictions each model made on the test set for $n=500$. We sampled from the predictive distribution using Algorithm \ref{alg:sample-mard}:

\vspace{5mm}

Figure \ref{fig:test2-fake_dataset} compares the spread of predictions made by ARD (red) and MARD (green) to the real values (blue). At a first glance, it is clear that both model struggle with the dispersion crated by the models variance in Figure \ref{fig:fake_dataset} and seem to provide similar results. The RSME values are also indistinguishable: $10.51$ for ARD and $10.48$ for MARD. For $n=100$, ARD outperforms MARD by reaching a RMSE of $16.6$, half of MARD's value.

\begin{figure}[H]
    \centering
        \includegraphics[width=0.5\textwidth]{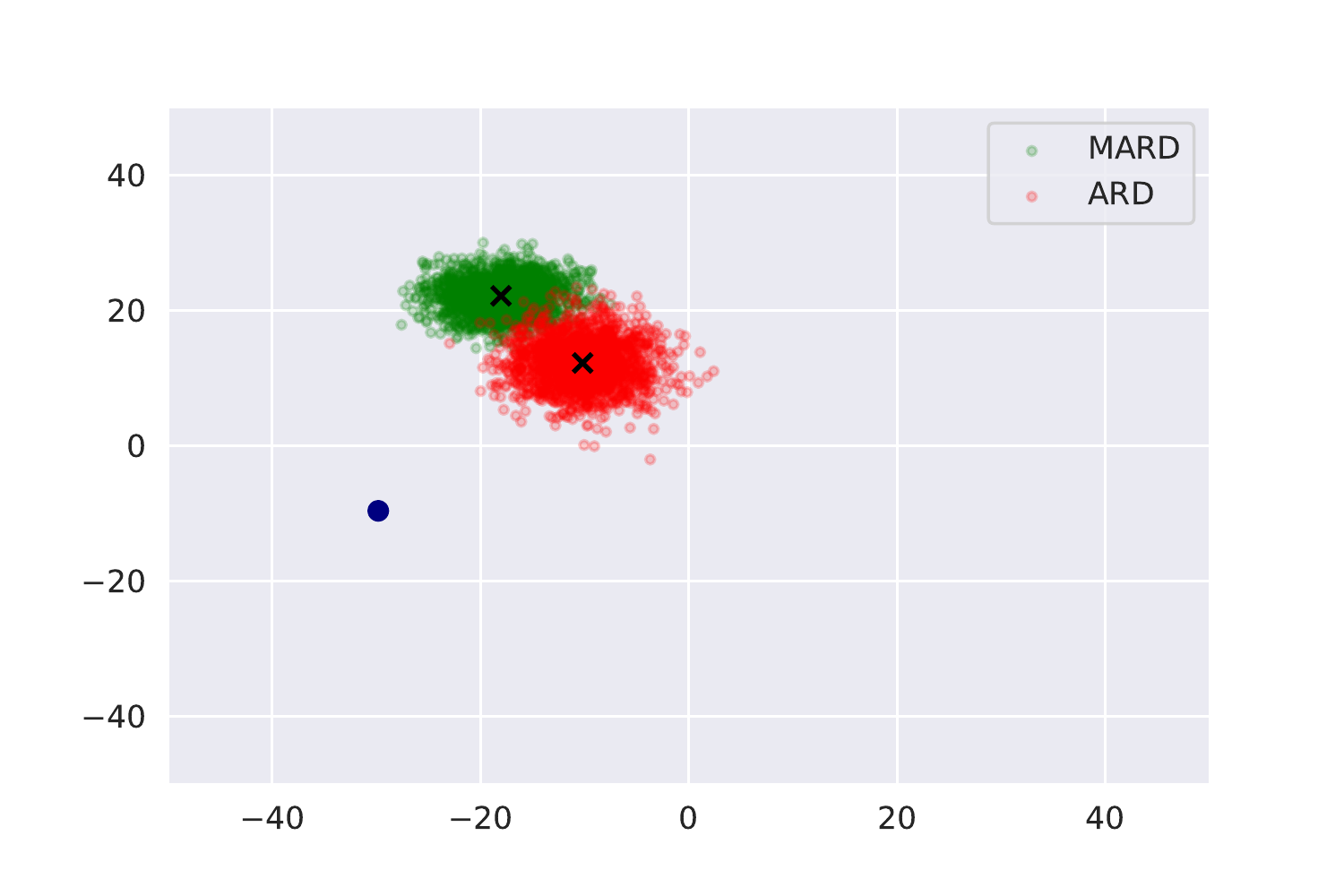}
        \caption{Display of preditive samples by ARD and MARD predictions for a single response with $n=100$.}
        \label{fig:intervalsfake_dataset}
\end{figure}

To analyse the main metric used in the previous models, we have calculated credible regions by sampling from the predictive distribution and displayed them in Figure \ref{fig:intervalsfake_dataset}. The black X marks the values each model predicted, while the blue dot is the real value the models were aiming to predict. The credible region is created by the dispersion dots sampled from the predictive distribution, with red representing ARD and green MARD.

Overall, the results are show in Table \ref{table:intervals_dummy}. In the manufactured dataset contains $1000$ samples and we can see a significant drop in performance when both work with $100$ samples in the training set, with ARD taking the lead.

\begin{table}[H]
\begin{center}
\begin{tabular}{||c c c c||} 
\hline
\textbf{n} & 5000 & 500 & 100  \\ [0.5ex] 
 \hline\hline
ARD &  $165$  & $397$ & $432$ \\
\hline
MARD  & $167$ & $379$  & $194$ \\[1ex]
\hline
 \hline
\end{tabular}
\end{center}
\caption{Comparison between adequate intervals created by ARD and MARD for three different dataset sizes.}
\label{table:intervals_dummy}
\end{table}

\section{Results from MARD} \label{sec:mard-results}

\begin{figure}[H]
    \centering
        \subfloat{\includegraphics[width=0.5\textwidth]{Images/Beta_A_ARD.pdf}}
        \subfloat{\includegraphics[width=0.5\textwidth]{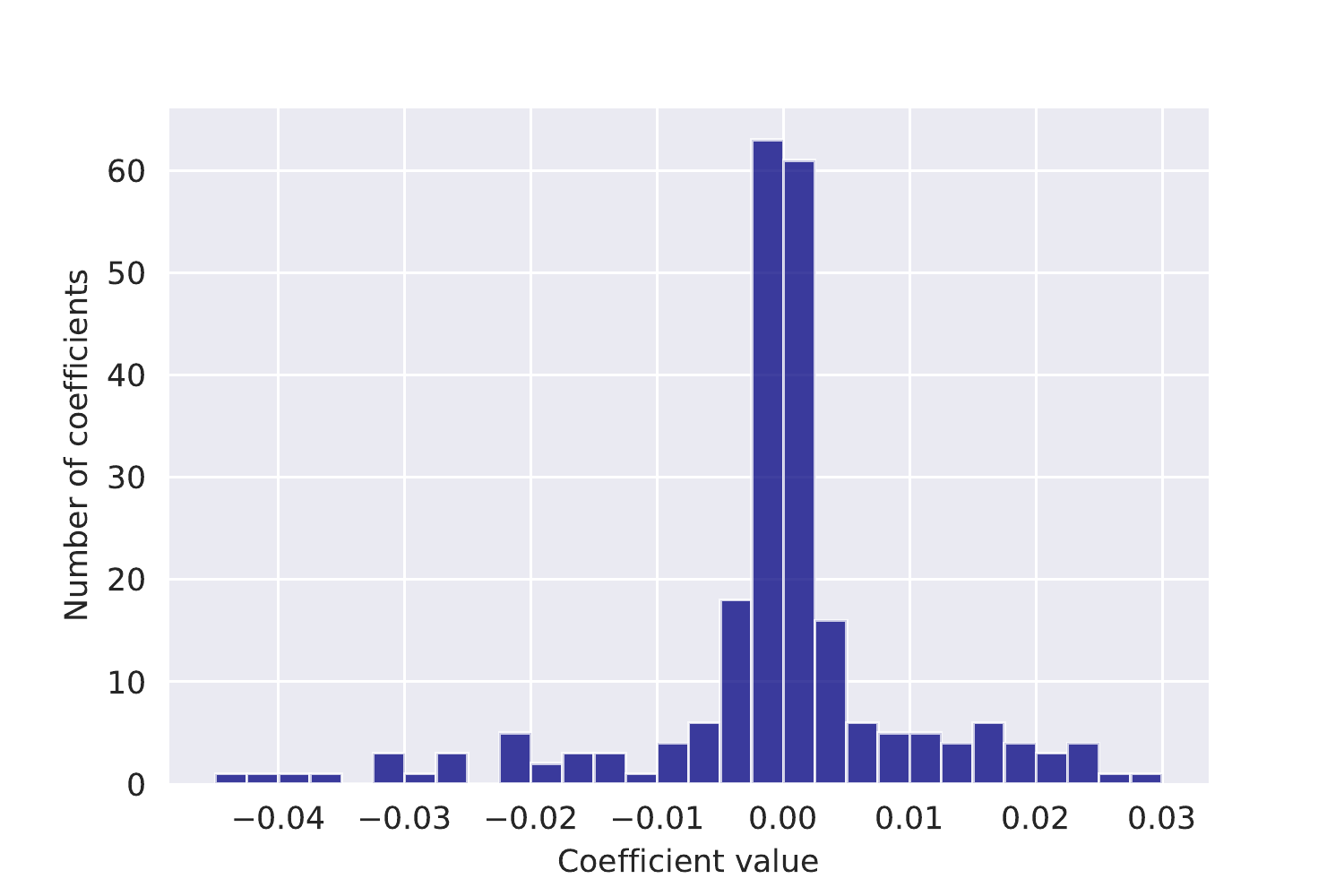}}
        \caption{Histograms of linear regression coefficients for Arousal with feature selection via ARD (left) and  MARD (right).}
        \label{fig:beta-a-mard}
\end{figure}

\begin{figure}[H]
    \centering
        \subfloat{\includegraphics[width=0.5\textwidth]{Images/Beta_V_ARD.pdf}}
        \subfloat{\includegraphics[width=0.5\textwidth]{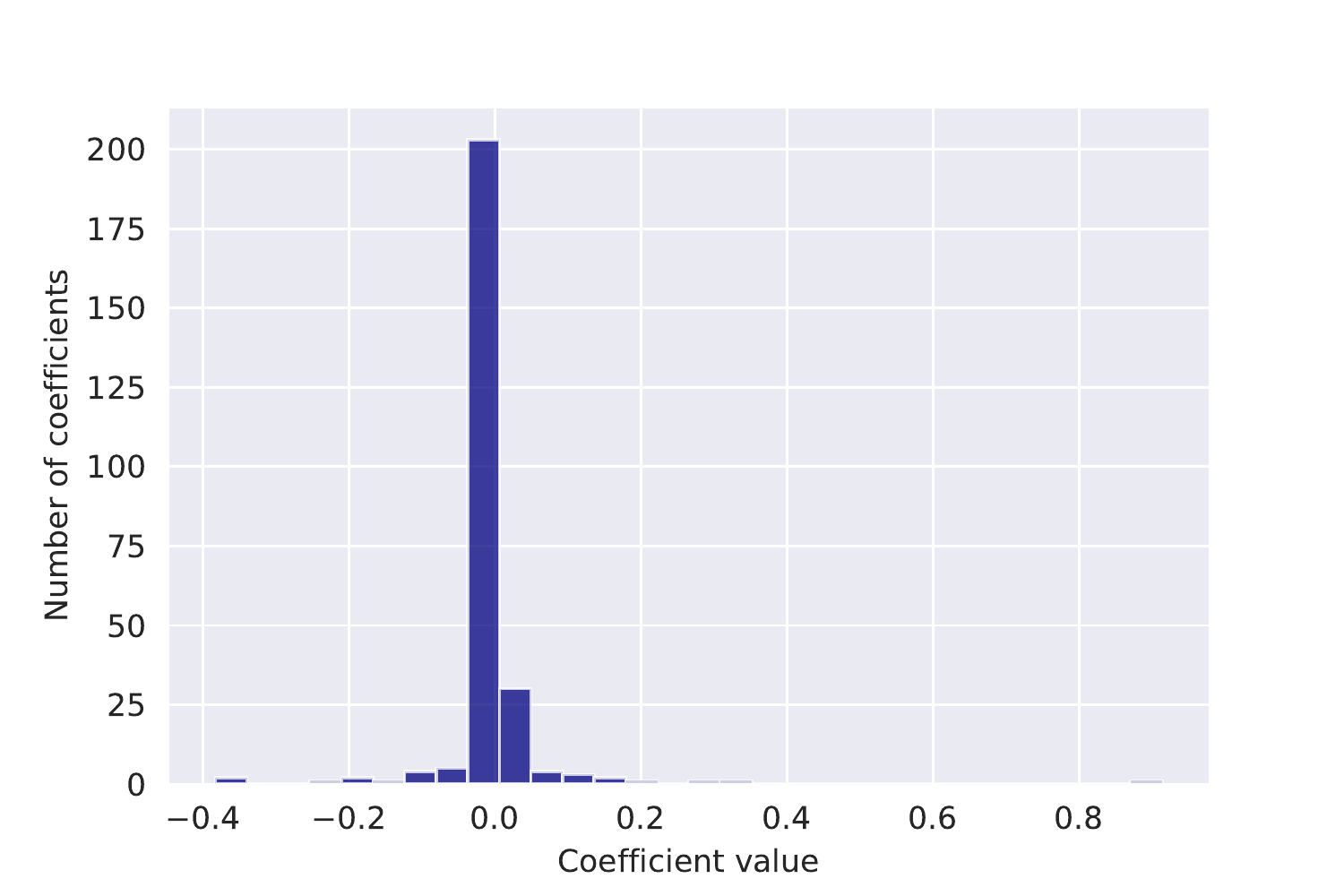}}
        \caption{Histograms of linear regression coefficients for Valence with feature selection via ARD (left) and  MARD (right).}
        \label{fig:beta-v-mard}
\end{figure}

Now that we have built some insight on how MARD operates in regard to ARD and acknowledged that it performs adequately, let us try it on the DEAM dataset and see if it brings any developments to the problem at hand. First let us compare MARD's parameter estimation in comparison to ARD's in Figures \ref{fig:beta-a-mard} and \ref{fig:beta-v-mard}. While MARD turned a slightly higher number of of parameters to zero for Arousal,  the work done for Valence was a lot more drastic, with most of $\beta$'s being discarded or remaining very low. 

As done before, we will carry on observing some performance measures. Even though $R^2$ values are very similar, there has been a huge improvement considering the credible regions, with both of them almost reaching the total number of test samples ($540$). Let us take this information this further.

\begin{table}[H]
\begin{center}
\begin{tabular}{||c c c c c||} 
\hline
\textbf{Arousal} & Classic & Bayesian & ARD & MARD \\ [0.5ex] 
 \hline\hline
Training $R^2$ &  $0.22$  & $0.60$ & $0.75$ & $0.74$\\
\hline
Test $R^2$ & $0.15$ & $0.58$  & $0.66$ & $0.66$\\
\hline
Credibility intervals & - & $201$ & $270$ & $517$\\[1ex]
\hline
 \hline
\end{tabular}
\end{center}
\caption{Comparison between classical, Bayesian ARD and MARD models for Arousal.}
\label{table:results3A}
\end{table}

\begin{table}[H]
\begin{center}
\begin{tabular}{||c c c c c||} 
\hline
\textbf{Valence}  & Classic & Bayesian & ARD & MARD \\ [0.5ex] 
 \hline\hline
Training $R^2$ &  $0.12$  & $0.45$ & $0.59$ & $0.56$\\
\hline
Test $R^2$ & $0.06$ & $0.25$  & $0.29$ & $0.33$\\
\hline
Credibility intervals & - & $185$ & $231$ & $518$\\[1ex]
\hline
 \hline
\end{tabular}
\end{center}
\caption{Comparison between classical, Bayesian ARD and MARD models for Valence.}
\label{table:results3V}
\end{table}

\begin{figure}[H]
    \centering      
        \includegraphics[width=1\textwidth]{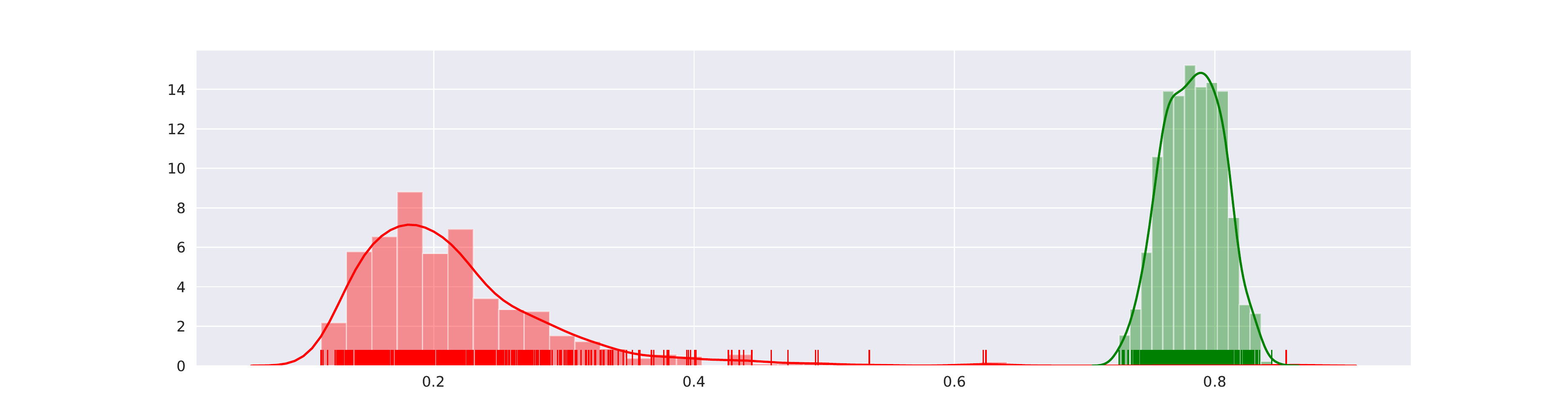}
        \caption{Spread of credibility intervals created by ARD (red) and MARD (green) for Arousal (first) and Valence (second).}
        \label{fig:intervals}
\end{figure}

The Figure \ref{fig:intervals} contains a histogram of the intervals' amplitudes. In it we can see that the intervals created by MARD are much larger, what at least partially accounts for the results we obtained, meaning a recommendation system created using MARD would be more generalist.

\begin{figure}[H]
    \centering
        \includegraphics[width=1\textwidth]{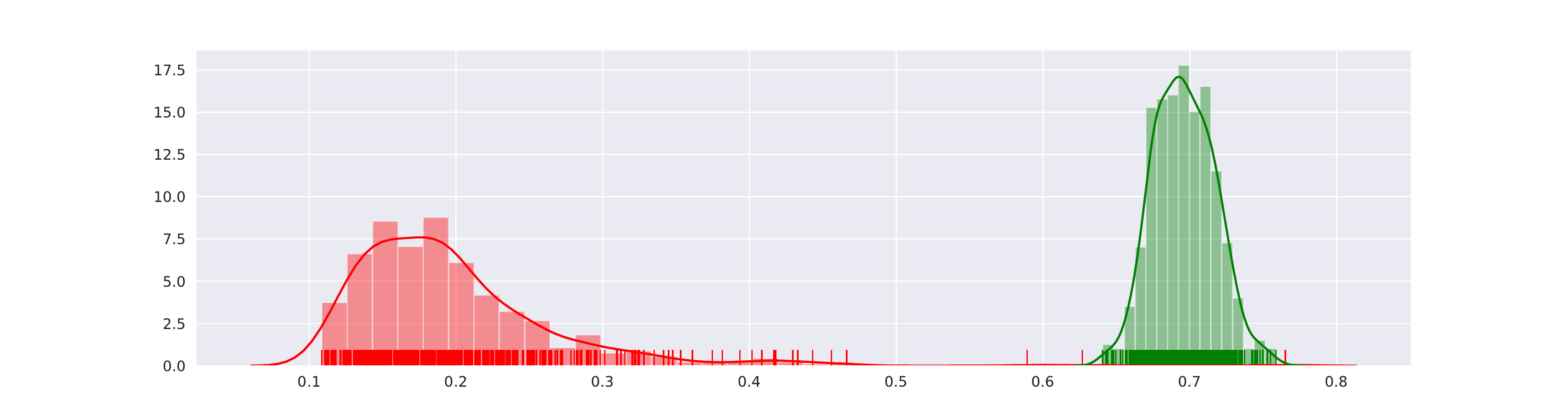}
\end{figure}

\begin{figure}[H]
    \centering      
        \includegraphics[width=1\textwidth]{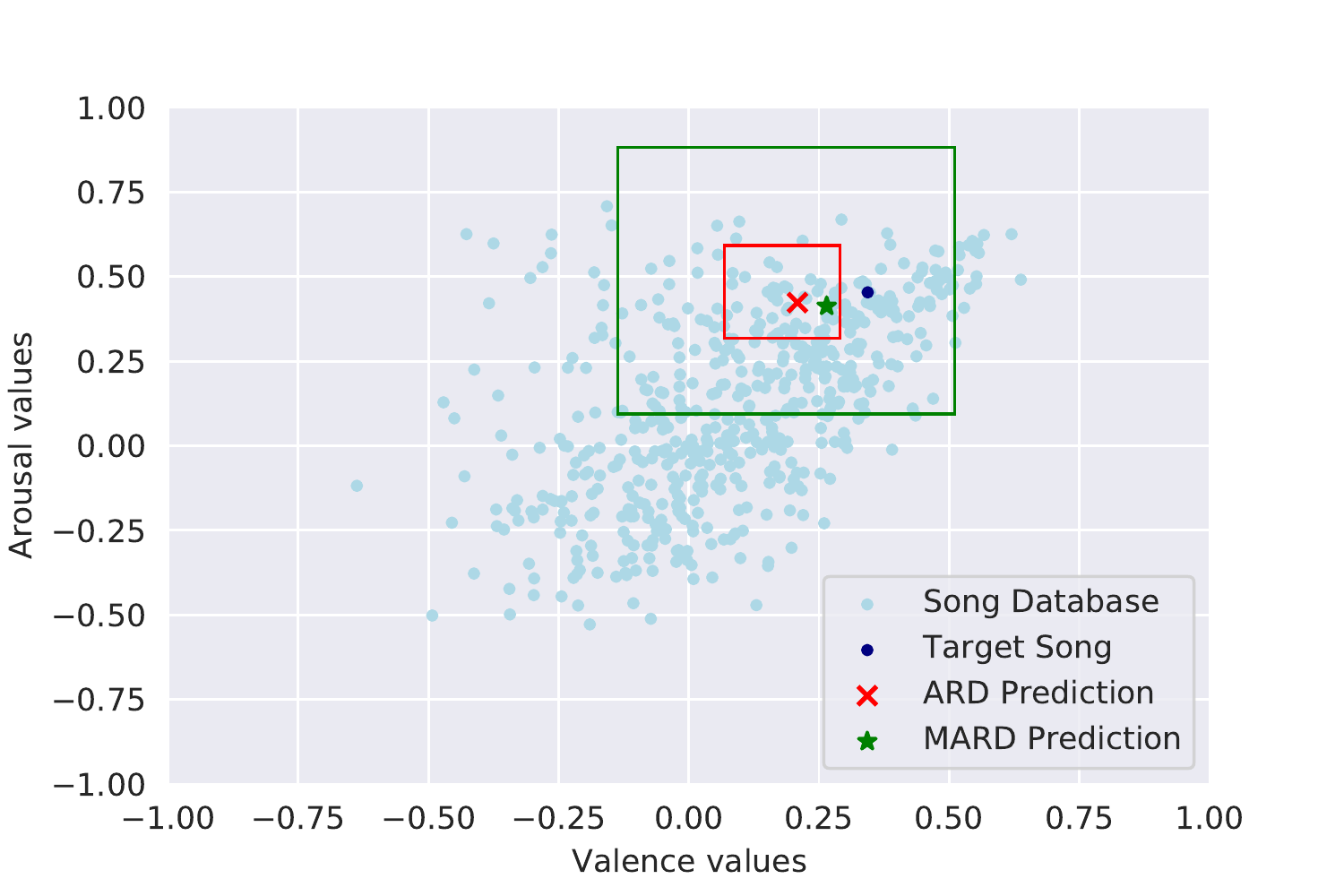}
        \caption{Example of how a few pieces would be recommended by ARD and MARD considering a single song.}
        \label{fig:recom}
\end{figure}

Let us consider, for instance, that a user has chosen a particular piece of music, portrayed in navy blue in Figure \ref{fig:recom} and wishes to listen to $n$ other songs of similar emotion from the dataset, in light blue. A recommendation system employing ARD would recommend the ones contained in the red rectangle, while MARD would choose from the green one.

\subsection{Conclusions}

In this Chapter, we went through a lengthy process in order to transform ARD into a multivariate model with hopes that it would capture more information and return better results. Even though comparisons between the two approaches did not differ much in the controlled setting, we were able to see great disparities in the actual case of MER. While ARD return a narrower region, MARD creates very broad intervals, more capable to encapsulate the target while also risking comprehending data that does not relate to the objective. The possible ramifications of this will be discussed in the following and final Chapter.

\chapter{Conclusion}\label{Conclusions}
    
In this work, we have explored a relationship between Music Information Retrieval and Statistical knowledge and how we can use new Bayesian inference methods to expand possibilities in an old problem. 

Although Bayesian techniques can be very computationally taxing, here we find a way to work around that issue by bringing in Variotinal Inference. Albeit it returns an approximated result, this approach manages to return good results in mere seconds in the database tested here.

\begin{figure}[H]
    \centering
        \subfloat{\includegraphics[width=0.5\textwidth]{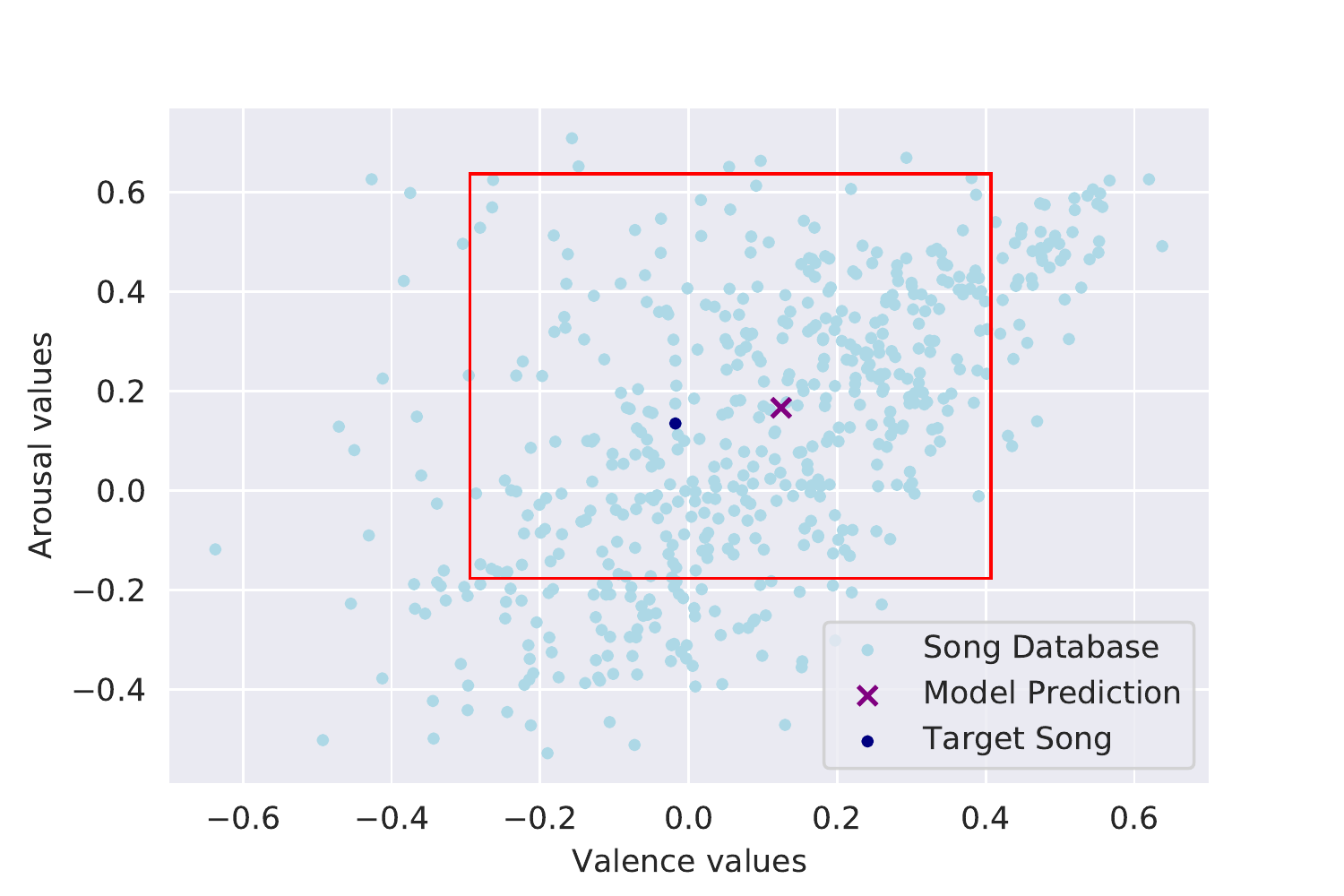}}
        \subfloat{\includegraphics[width=0.5\textwidth]{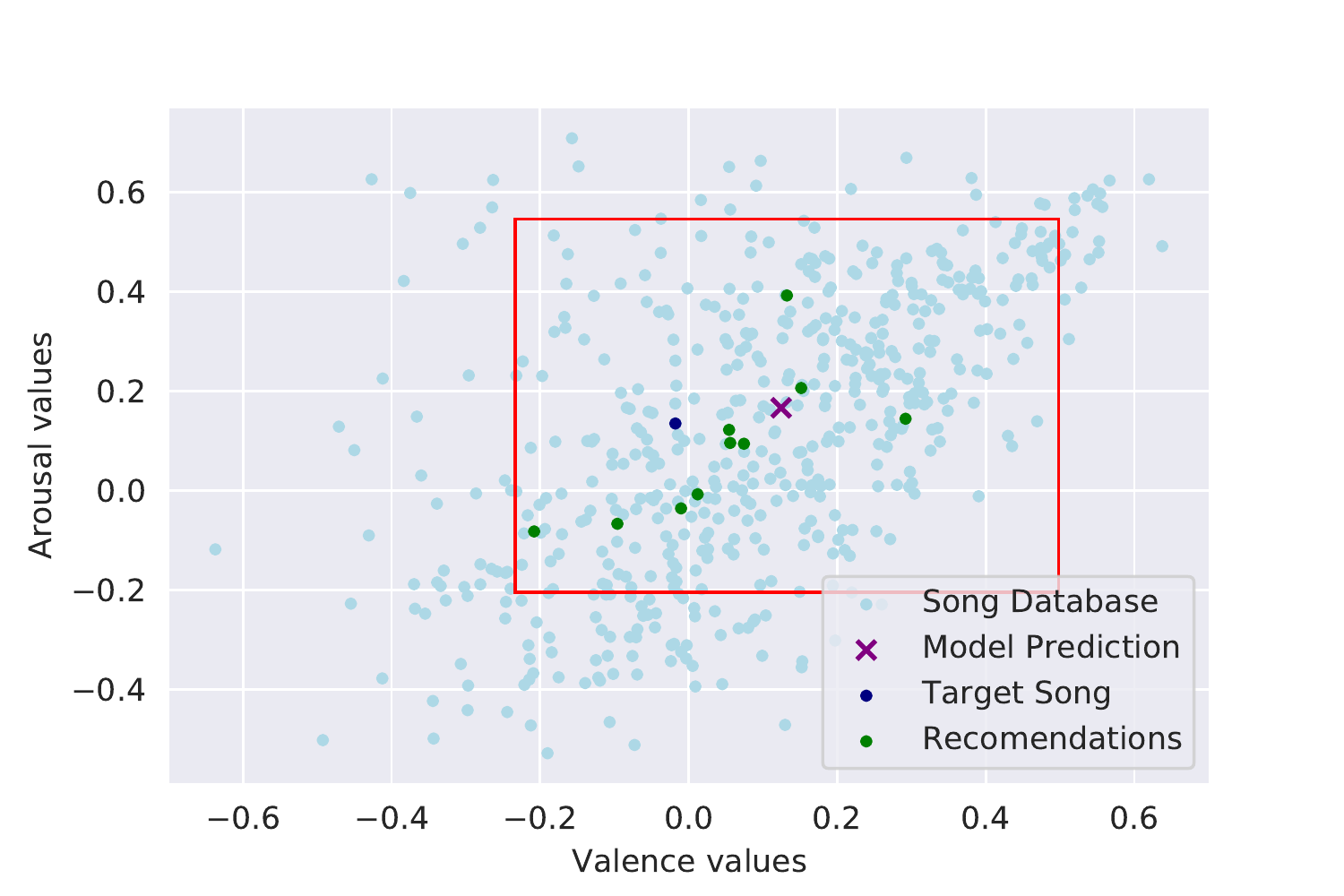}}
        \caption{Recommendation system created by the Gibss, ARD and MARD model, respectively.}
        \label{fig:recom_sys}
        
\end{figure}

As mentioned throughout this work, the metric proposed for the Bayesian models can be used to build an emotion-based recommendation system where all songs with AV values within a chosen song's interval can be recommended to a user. Let us see in Figure \ref{fig:recom_sys} below how that would work for each model, Note that the images were enlarged and do not range between $[-1,1]$, to improve understanding. We have displayed the results for Gibbs first, then for ARD and finally for MARD. A particular song was chosen and marked by the blue dot. A prediction was made by each model and marked by the red $x$ and around this prediction we have built the Cartesian product of the credibility intervals calculated for this prediction as the red rectangle. Inside each rectangle, ten songs were chosen as recommendations for the user. As previously discussed, MARD's creates much wider intervals and its recommendations can stray further from the target than the other methods.

\begin{figure}[H]
    \centering
        \subfloat{\includegraphics[width=0.5\textwidth]{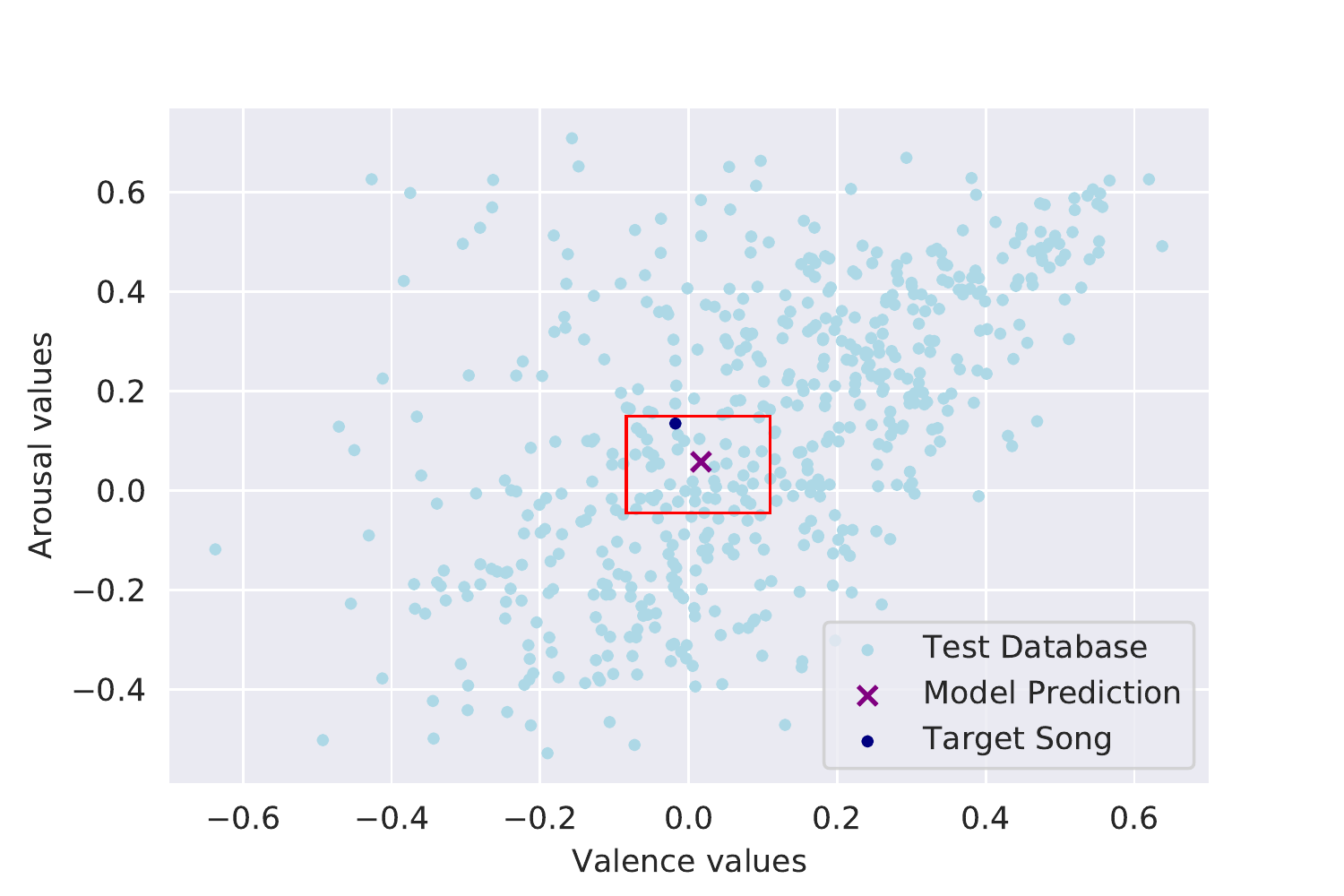}}
        \subfloat{\includegraphics[width=0.5\textwidth]{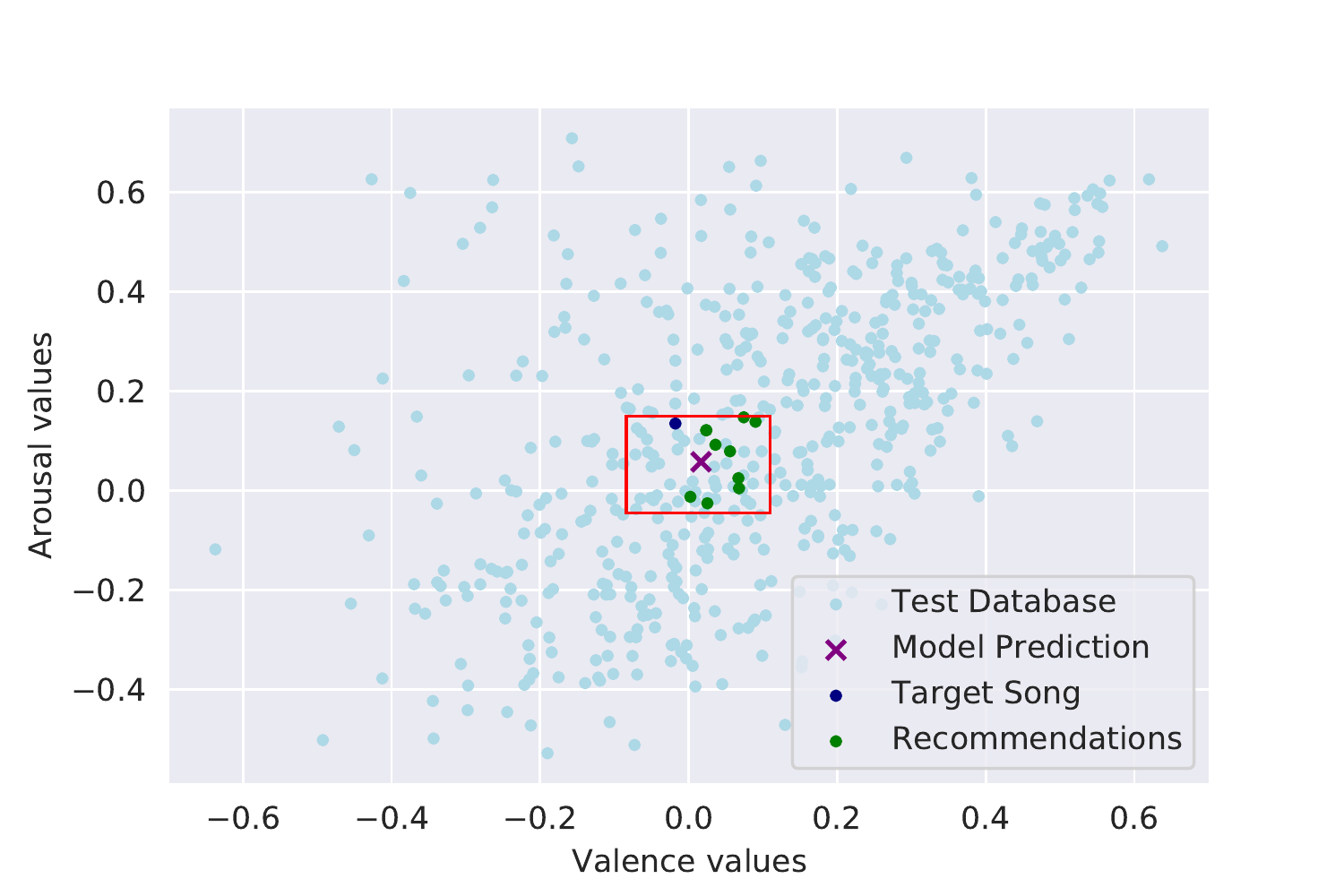}}
\end{figure}

\begin{figure}[H]
    \centering
        \subfloat{\includegraphics[width=0.5\textwidth]{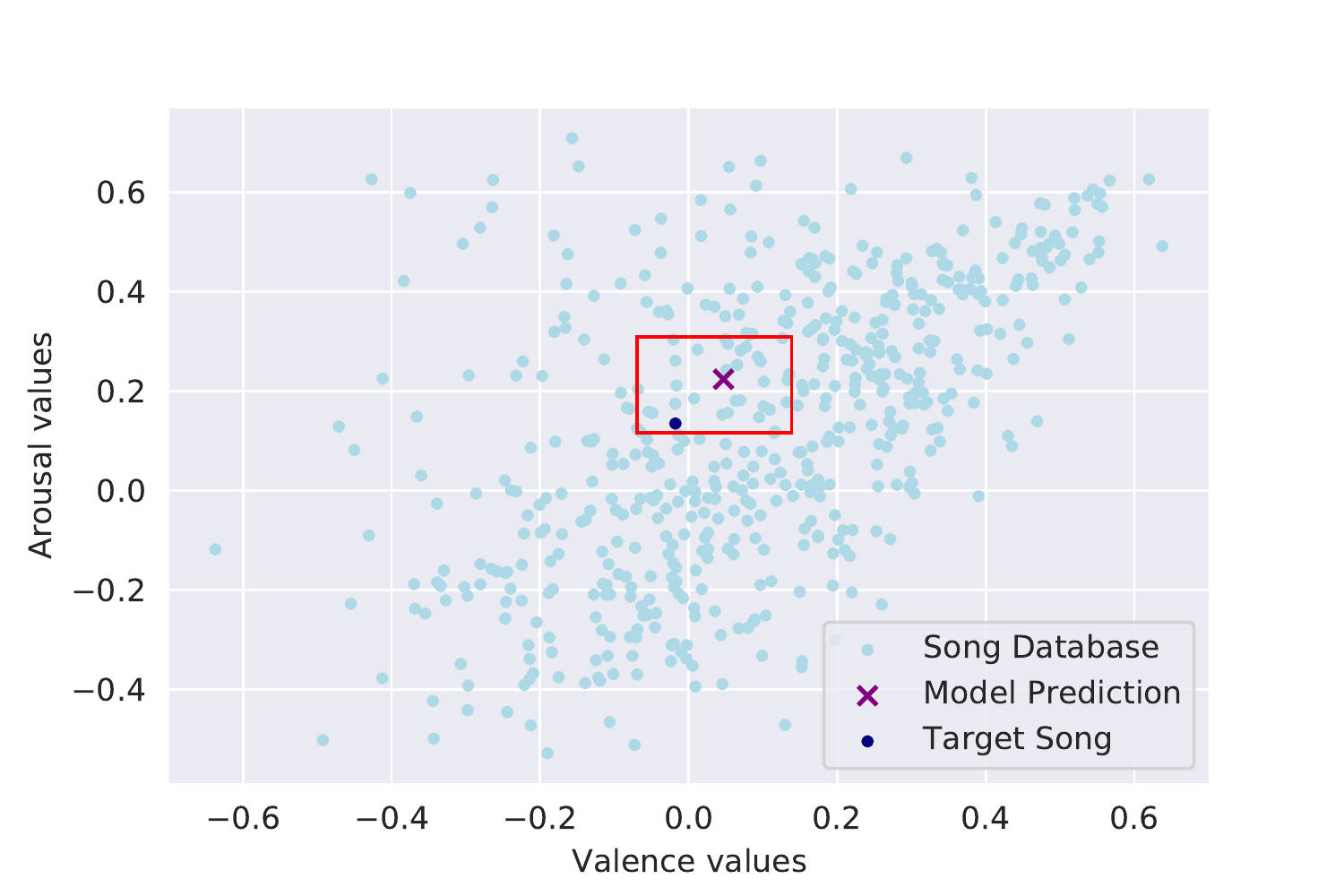}}
        \subfloat{\includegraphics[width=0.5\textwidth]{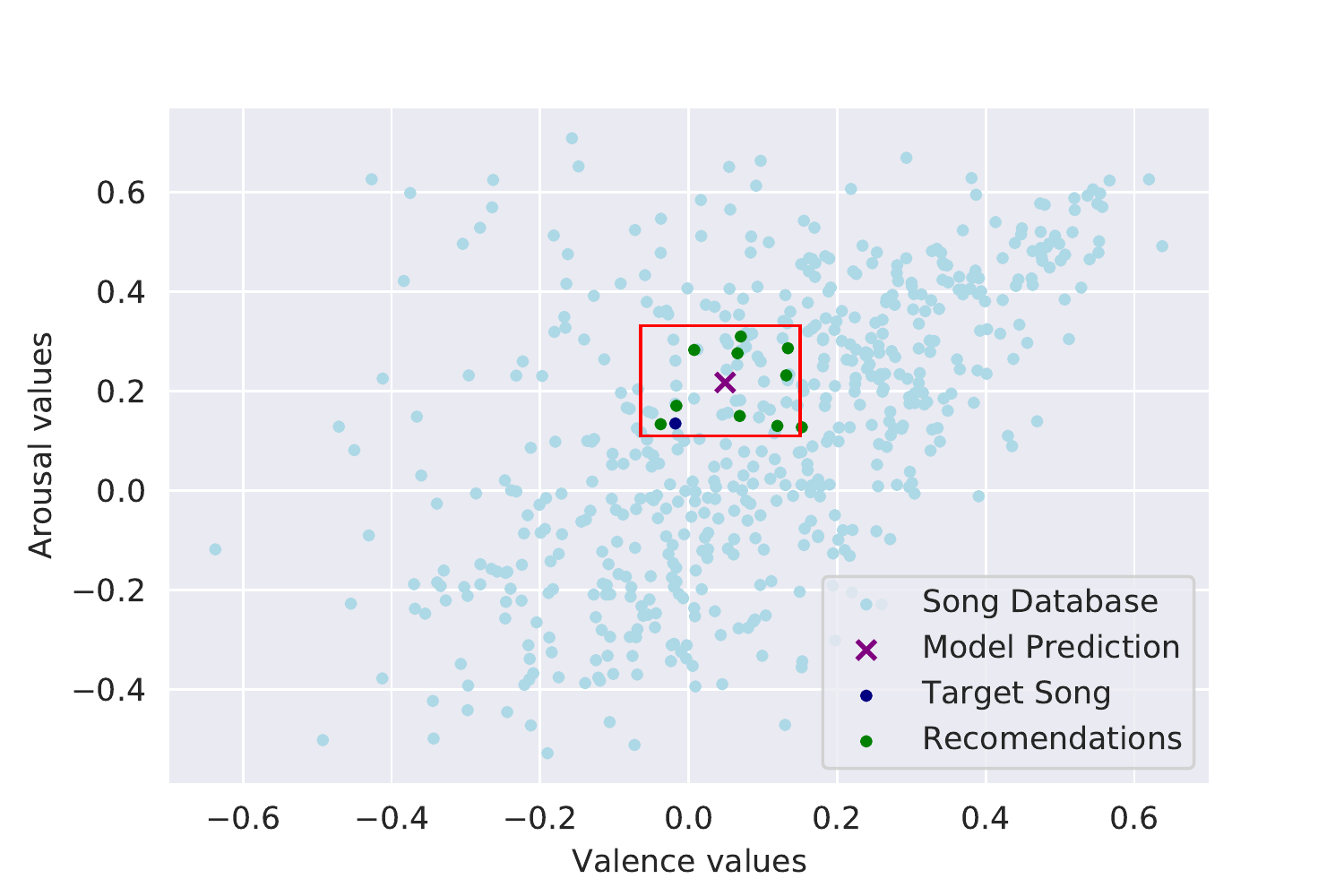}}
\end{figure}

However, there is something slightly deceitful about Figure \ref{fig:recom_sys}. We have deliberately chosen a song that all three models were able to predict somewhat precisely making MARD's system look much poorer that the others. Let us remember that this is not the case for most of the songs, as not even half were captured by ARD's intervals and even less by Gibbs. Let us analyse one setting where the models do not fare so well.

\begin{figure}[H]
    \centering
        \subfloat{\includegraphics[width=0.5\textwidth]{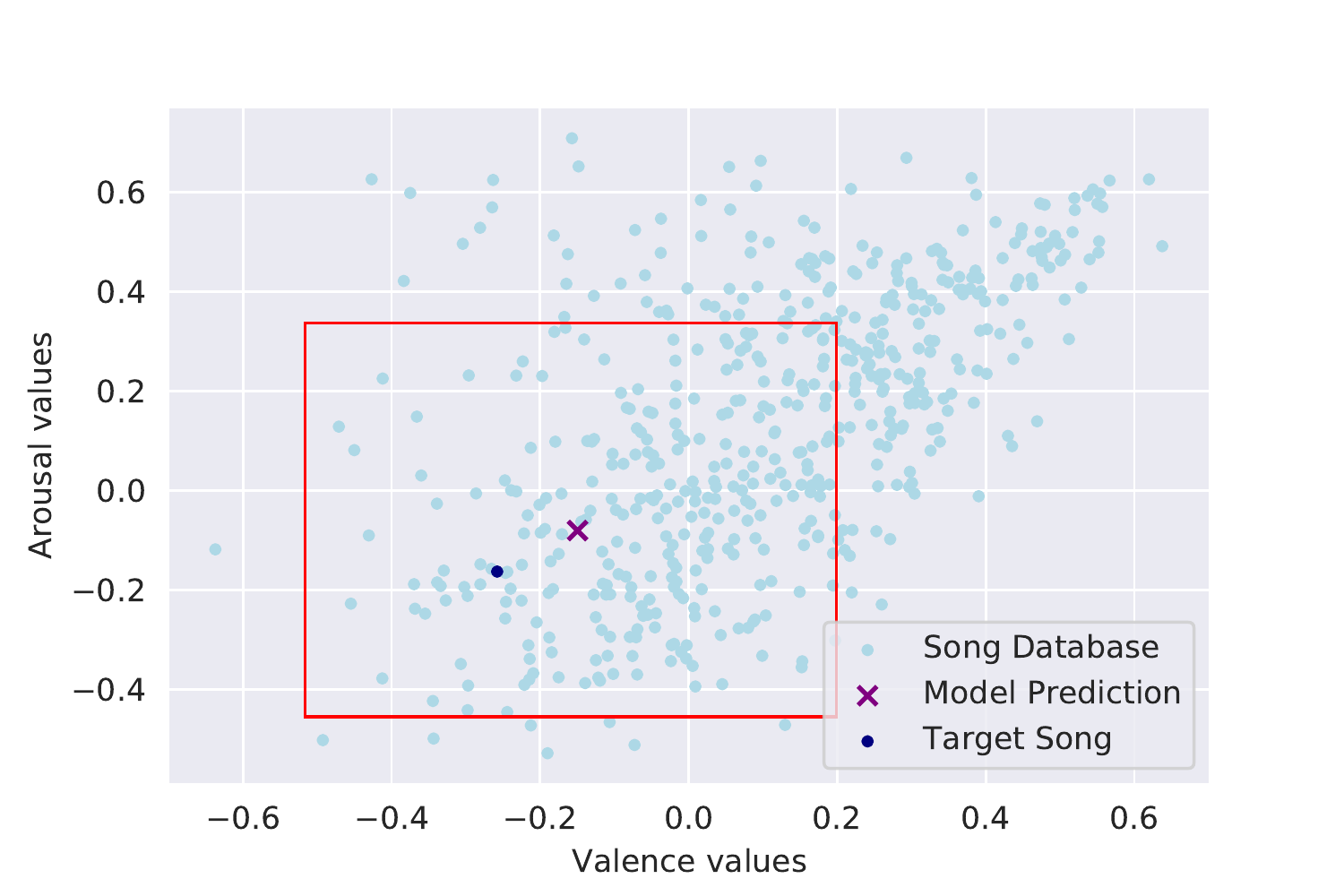}}
        \subfloat{\includegraphics[width=0.5\textwidth]{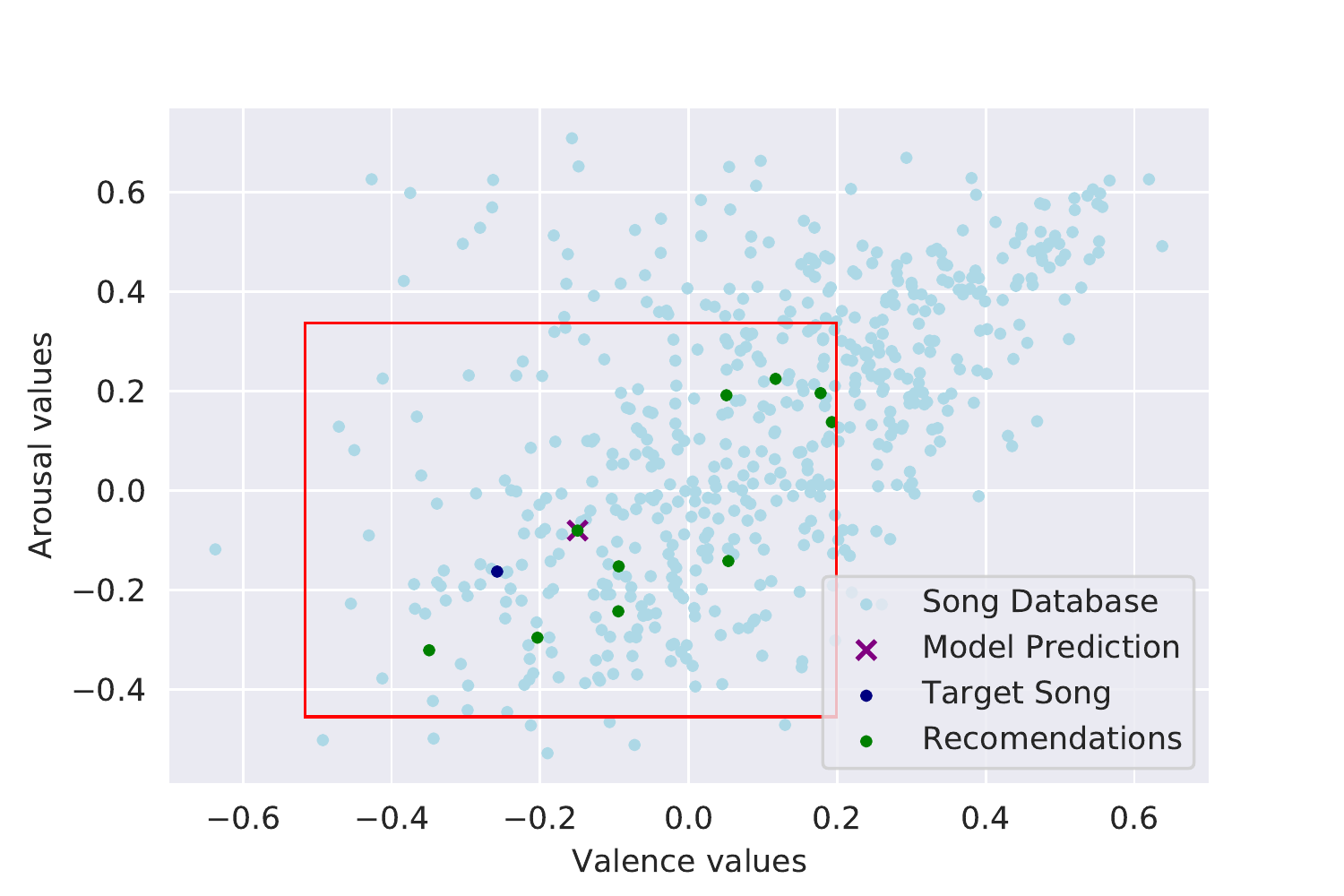}}
        \caption{Recommendation system created by the Gibss, ARD and MARD model, respectively. In this case, for a piece not as well predicted.}
        \label{fig:recom_sys_2}
        
\end{figure}

In Figure \ref{fig:recom_sys_2} we can see how the other models can end up stuck far from the target and choose poorly, while MARD has a bigger probability to succeed. Not only that, but if the user had few songs in his database that matched with the emotion he was aiming for, Gibbs and ARD could return very few pieces or nothing at all, due to their limited space. Or if the user wanted $20$ ou more songs rather than just $10$, both would also fall short, while MARD would be able to provide a reasonable estimate. 

\begin{figure}[H]
    \centering
        \subfloat{\includegraphics[width=0.5\textwidth]{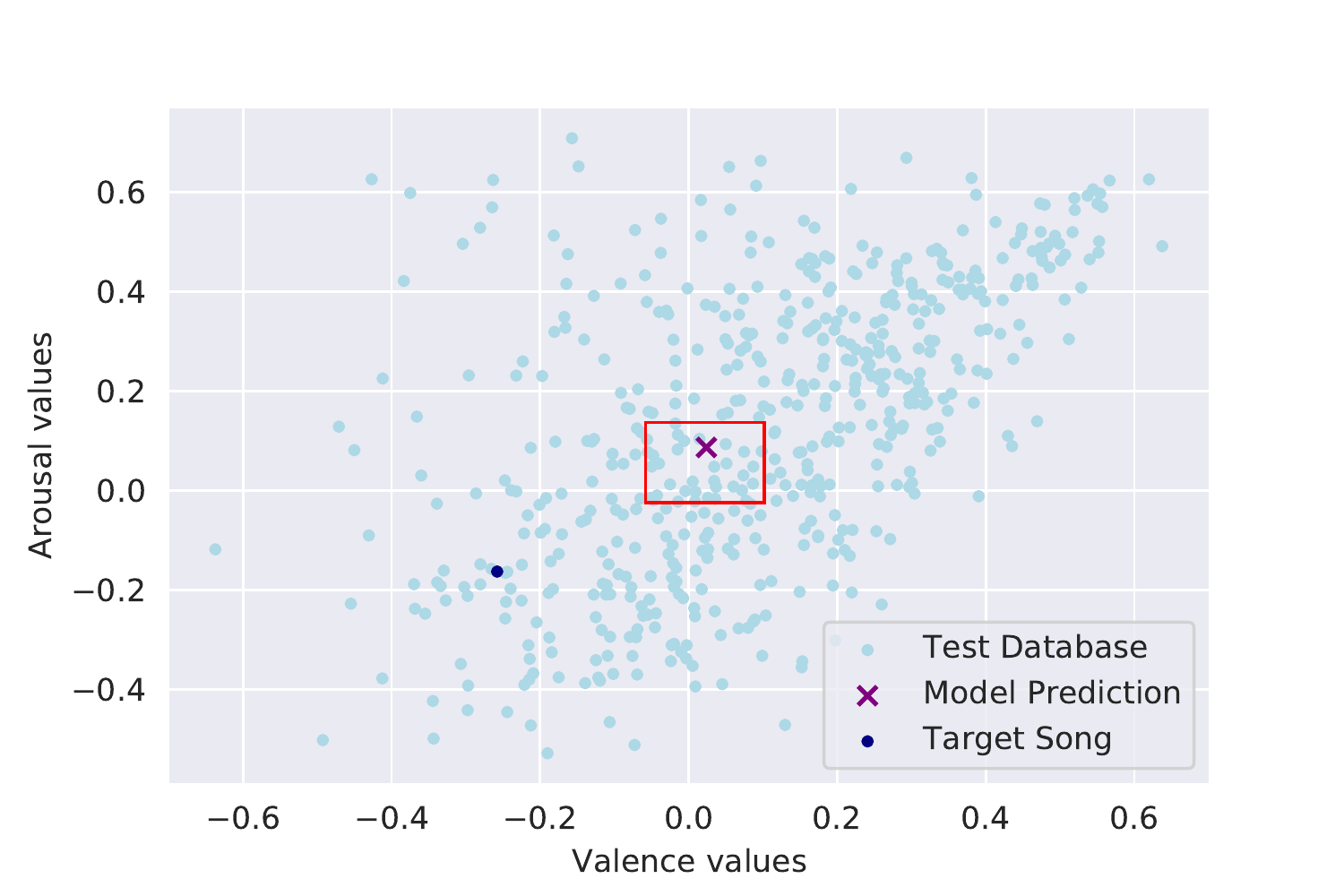}}
        \subfloat{\includegraphics[width=0.5\textwidth]{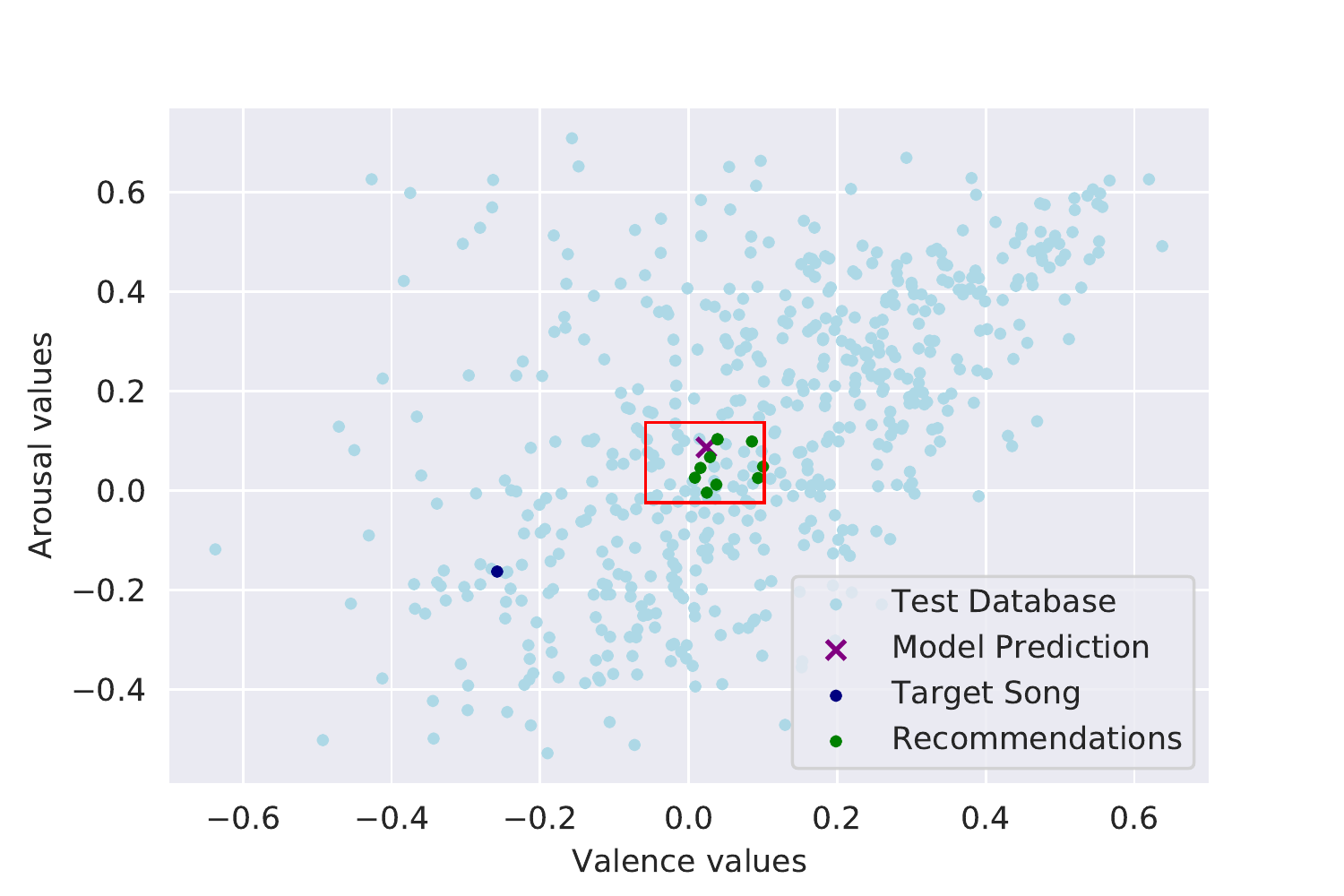}}
\end{figure}

\begin{figure}[H]
    \centering
        \subfloat{\includegraphics[width=0.5\textwidth]{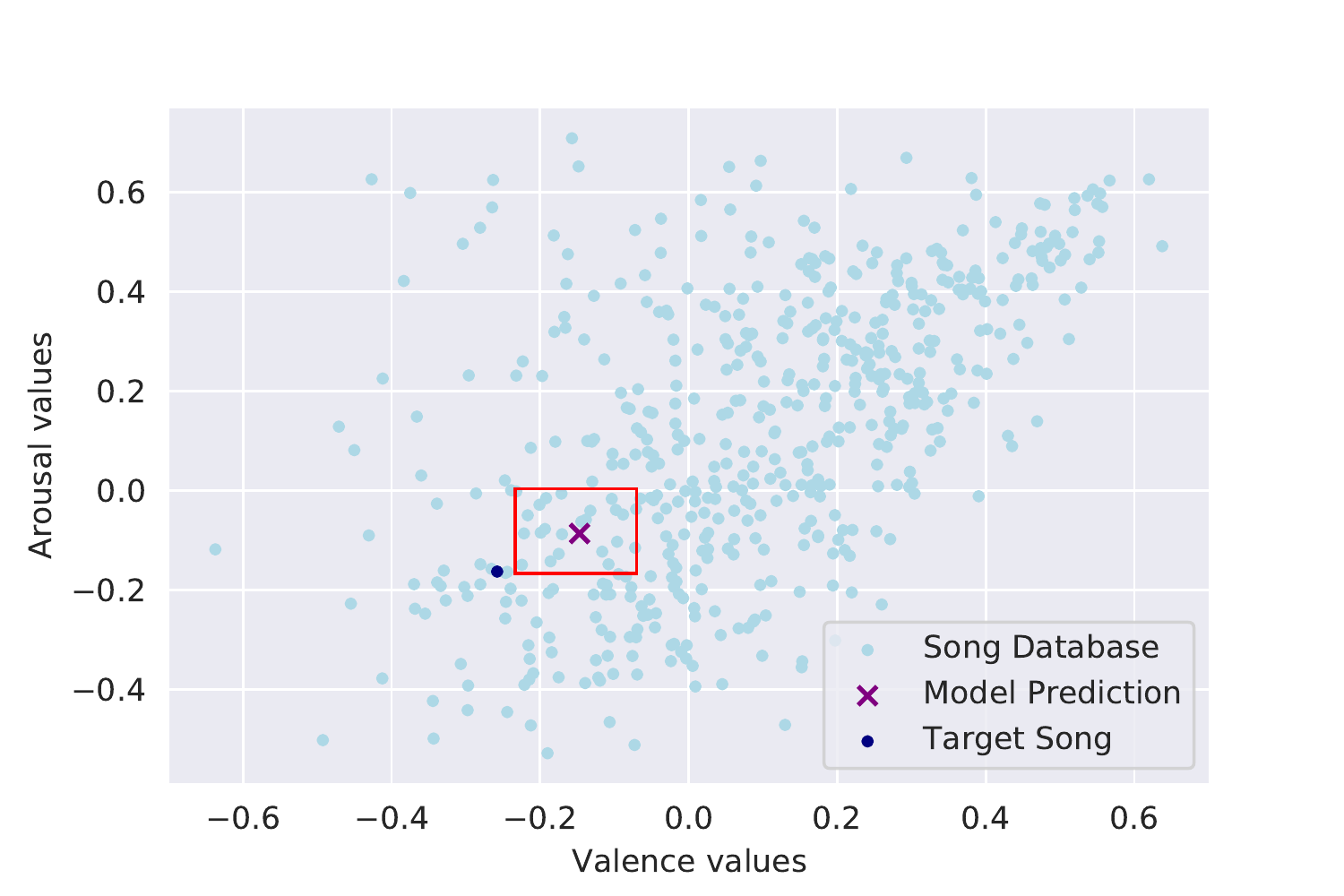}}
        \subfloat{\includegraphics[width=0.5\textwidth]{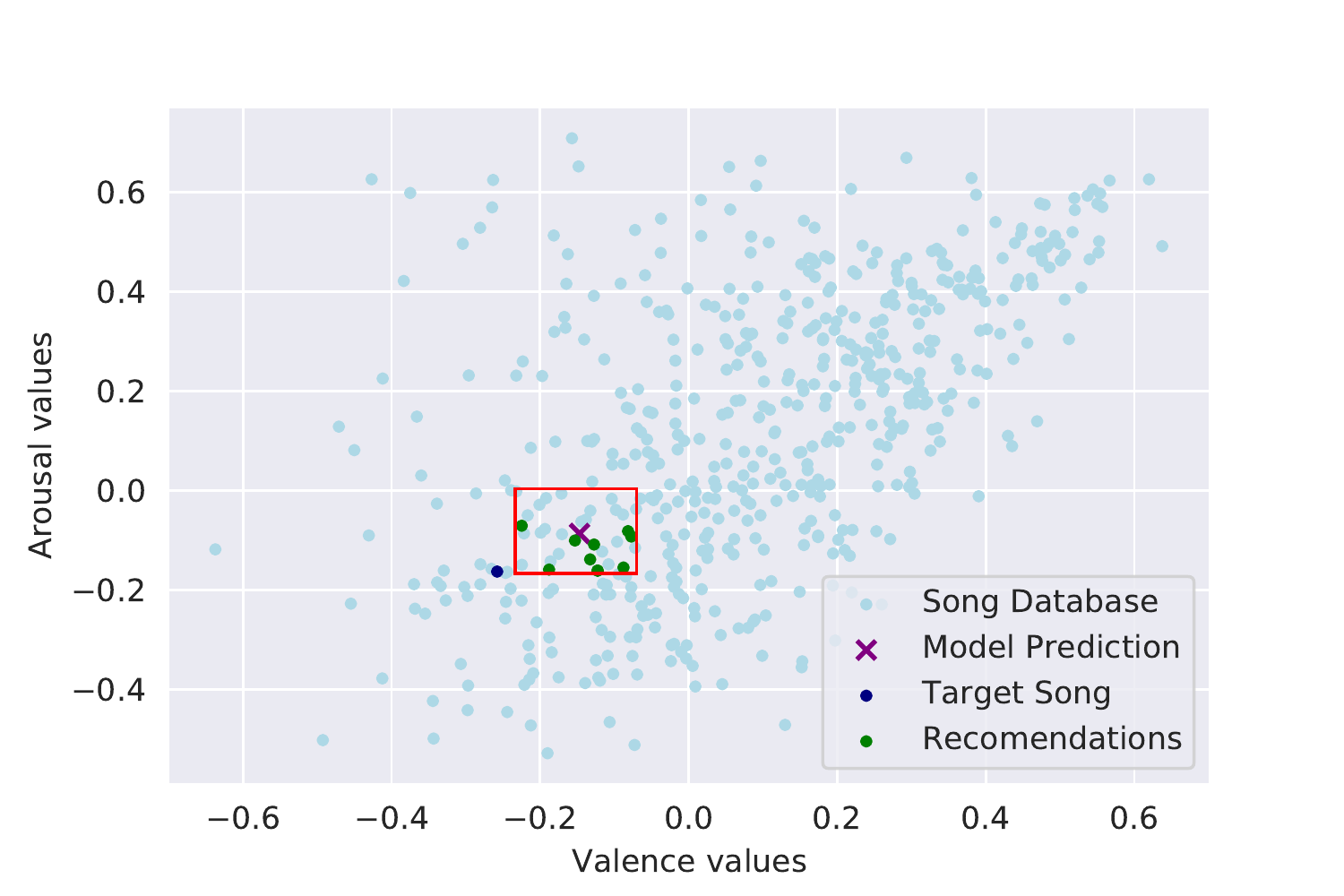}}
\end{figure}

Overall, the usage of modern statistical methods offered some very interesting insights and possibilities as the models proposed were able to explore this data in new forms, generating unseen results. We hope that this endeavour encourages more statisticians to look into Music Emotion Recognition and keep proposing new developments to the field.

\appendix
\chapter{Equivalence between the Bayesian and classical LASSO}\label{appendix:A}

Here let us prove that model described by lines $1, 2$ and $4$ in Equation \ref{eq:bayes-1} in Chapter \ref{bayes} has the same maximum a posteriori estimation as the classical LASSO. In this scenario, we consider $\lambda$ is fixed and $\sigma^2$ is known. Let $\bfx \in \mathbb{R}^{n \times p}$ be the matrix containing the explanatory variables, where $n$ in the number of observations and $p$ the number of parameters; $\bbfy \in \mathbb{R}^{n}$ will represent the response variable. Assume that each $\beta_j$ follows independent Laplace distributions with common parameter lambda and work through the necessary calculations to find the equivalence between these approaches.

We can assume the response variable $\bbfy$ to follow a normal distribution with mean $\beta_0 + \sum_{j=1}^p \beta_jx_{ij}$ and $\sigma^2$ variance. Joining the likelihood $L$ with the Laplace prior distribution previously defined we find the following posterior $p(\beta|\bfx, \bbfy)$:

\begin{equation}
\begin{split}
p(\beta|\bfx,\bbfy) & \propto L(\bbfy| \beta, \bfx)p(\beta) \\
&  \propto \left( \frac{1}{\sqrt{2\pi\sigma^2}} \right)^n \!\! \exp \left(-\frac{1}{2\sigma^2}\sum_{i=1}^n \left(\bbfy_i - \beta_0 - \sum_{j=1}^p \beta_jx_{ij}\right)^2\right)\frac{1}{2b}\exp \left(-\frac{|\beta|}{b}\right)\\
& \propto  \left( \frac{1}{\sqrt{2\pi\sigma^2}} \right)^n \!\! \frac{1}{2b}  \exp \left(-\frac{1}{2\sigma^2}\sum_{i=1}^n \left(\bbfy_i - \beta_0 - \sum_{j=1}^p \beta_jx_{ij}\right)^2-\frac{|\beta|}{b}\right)
\end{split}
\end{equation}

Now we must verify that the $\beta$ estimated by this posterior is equal to the one estimated by LASSO regularization. First let us apply the logarithmic function to this equation, since maximizing a function is analog to maximizing its logarithm:

\begin{equation}
\begin{split}
\log(p(\beta|\bfx,\bbfy)) & \propto  \log \left(\left( \frac{1}{\sqrt{2\pi\sigma^2}} \right)^n  \frac{1}{2b}  \exp  \left(-\frac{1}{2\sigma^2}\sum_{i=1}^n \left(\bbfy_i - \beta_0 - \sum_{j=1}^p \beta_jx_{ij}\right)^2-\frac{|\beta|}{b}\right)\right)\\
& \propto  \log \left[\left( \frac{1}{\sqrt{2\pi\sigma^2}} \right)^n \frac{1}{2b}\right] - \left(\frac{1}{2\sigma^2}\sum_{i=1}^n \left(\bbfy_i - \beta_0 - \sum_{j=1}^p \beta_jx_{ij}\right)^2-\frac{|\beta|}{b}\right)
\end{split}
\end{equation}
Since we wish for the value of $\bfbeta$ which maximizes the posterior:
\begin{equation}
\begin{split}
\underset{\bfbeta \in \mathbb{R}^p}{\mathrm{argmax}} (p(\beta|\bfx,\bbfy)) & = \underset{\bfbeta \in \mathbb{R}^p}{\mathrm{argmax}}\ \!\! \log \left[\left( \frac{1}{\sqrt{2\pi\sigma^2}} \right)^n \!\! \frac{1}{2b}\right] \! - \! \left(\frac{1}{2\sigma^2}\sum_{i=1}^n \left(\bbfy_i - \beta_0 - \sum_{j=1}^p \beta_jx_{ij}\right)^2 \!\! - \frac{|\beta|}{b}\right)\\
& = \underset{\bfbeta \in \mathbb{R}^p}{\mathrm{argmin}} \left(\frac{1}{2\sigma^2}\sum_{i=1}^n \left(\bbfy_i - \beta_0 - \sum_{j=1}^p \beta_jx_{ij}\right)^2-\frac{|\beta|}{b}\right)\\
& = \underset{\bfbeta \in \mathbb{R}^p}{\mathrm{argmin}} \left( \frac{1}{2\sigma^2}\sum_{i=1}^n \left(\bbfy_i - \beta_0 - \sum_{j=1}^p \beta_jx_{ij}\right)^2-\frac{1}{b} \sum_{j=1}^p|\beta_j| \right),
\end{split}
\end{equation}
which matches the LASSO regularization:
$$\underset{\bfbeta \in \mathbb{R}^p}{\mathrm{argmin}} \left( \sum_{i=1}^n \left(\bbfy_i - \beta_0 - \sum_{j=1}^p \beta_jx_{ij}\right)^2+\lambda \sum_{j=1}^p|\beta_j|\right)$$
with $\lambda = -\frac{2\sigma^2}{b}$.

%\appendix
\chapter{Computation of the full conditional distributions of the Bayesian LASSO}\label{Appendix:B}

Here we will develop the necessary calculations in order to implement the Gibbs sampler used in Chapter \ref{bayes}. That is, we will find the model's conditional posteriors we intend to sample from. The model is identical to the one laid out in Equation \ref{eq:bayes-2}:

\vspace{-29mm}

\begin{align} \label{eq:bayes-2-calculation}
\begin{split}
\bbfy|\bfbeta, \bfbeta_0, \sigma^2 &\sim \text{N}(\mathbf{x_i}^T\bfbeta, \sigma^2),~i = 1, \dots n \\
\beta_0 &\sim \text{Flat} \\
\sigma^2 &\sim \text{IG}(a, b) \\
\beta_j | \gamma_j &\sim \text{N}(0, \gamma_j),~j = 1, \dots, p \\
\gamma_j | \lambda &\sim \exp(\lambda/2),~j = 1, \dots, p \\
\lambda &\sim \Gamma(c, d).
\end{split}
\end{align}

We can derive the following calculations for implementing the Gibbs sampler:

\begin{equation} \label{eq:post_cond}
\begin{split}
\p(\beta_0, \bfbeta, \bfgamma, \lambda, \sigma^2 | \bfx) & \propto \p(\bbfy | \beta_0, \bfbeta, \sigma^2, \bfx)\p(\beta_0)\p(\sigma^2)\p(\lambda)\p(\bfgamma|\lambda)\p(\bfbeta|\bfgamma) \\
& \propto \Bigg[\prod_{i=1}^n \frac{1}{\sqrt{\sigma^2}} \exp \Bigg\{-\frac{1}{2\sigma^2}\Bigg(y_i - \beta_0- \mathbf{x}_i^T\bfbeta \Bigg)^2\Bigg\} \Bigg] \\
& ~~~~ \Bigg[1 \Bigg]\Bigg[\frac{1}{(\sigma^2)^{a+1}} \exp \Bigg\{- \frac{b}{\sigma^2} \Bigg\} \Bigg]\Bigg[\lambda^{c-1} \exp\Bigg\{ -d \lambda\Bigg\} \Bigg] \\
& ~~~~ \Bigg[\prod_{j=1}^p \frac{\lambda}{2} \exp\Bigg\{-\frac{\lambda}{2}\gamma_j \Bigg\} \Bigg]\Bigg[\prod_{j=1}^p \frac{1}{\sqrt{\gamma_j}} \exp\Bigg\{-\frac{1}{2 \gamma_j}\beta_j^2 \Bigg\}\Bigg] \\
& = \Bigg[\frac{1}{(\sigma^2)^{n/2}} \exp \Bigg\{-\frac{1}{2\sigma^2}\sum_{i=1}^n \Bigg(y_i - \beta_0- \mathbf{x}_i^T\bfbeta\Bigg)^2\Bigg\} \Bigg] \\
& ~~~~ \Bigg[\frac{1}{(\sigma^2)^{a+1}} \exp \Bigg\{ \frac{b}{\sigma^2} \Bigg\} \Bigg]\Bigg[\lambda^{c-1} \exp\Bigg\{ -d \lambda\Bigg\} \Bigg] \\
& ~~~~ \Bigg[\Bigg( \frac{\lambda}{2} \Bigg)^p \exp\Bigg\{-\frac{\lambda}{2}\sum_{j=1}^p\gamma_j \Bigg\} \Bigg]\Bigg[\prod_{j=1}^p \frac{1}{\sqrt{\gamma_j}} \exp\Bigg\{-\frac{1}{2 \gamma_j}\beta_j^2 \Bigg\}\Bigg] \\
\end{split}
\end{equation}

Observing the full posterior in Equation \ref{eq:post_cond} we can calculate the conditional posterior for each variable by discarding the terms not related to the variable of interest and observing which distribution it can be recognized as. In some cases, such as $\beta_0$ and $\beta_j$ the quadratic form stands out quite easily, while $\sigma^2$ and $\lambda$ were respectively identified as an Inverse Gamma and  Gamma distributions, with $\gamma_j$ having presented characteristics of a more unusual distribution, the Generalized Inverse Gaussian.

\section{Conditional distribution of $\beta_0$}

In order to derive the parameters of the distribution for $\beta_0$ we will explore the first and second moments of the normal distribution:

\begin{equation}
\begin{split}
\beta_0 |\bfbeta, \bfgamma, \lambda, \sigma^2, \bfx & \sim N(\beta_0^M, \beta_0^{\sigma^2}) \\
 g(\beta_0) & = \sum_{i=1}^n \frac{(y_i - \beta_0- \bfbeta^T\mathbf{x}_i)^2}{2 \sigma^2}\\
 g'(\beta_0) & = -\frac{1}{2 \sigma^2} \sum_{i=1}^n 2(y_i - \beta_0- \bfbeta^T\mathbf{x}_i)\\
 & = -\frac{1}{\sigma^2}\Bigg[ \sum_{i=1}^n (y_i - \bfbeta^T\mathbf{x}_i) - n\beta_0 \Bigg] = 0\\
 & \Rightarrow \beta_0^M = \frac{1}{n} \sum_{i=1}^n (y_i - \mathbf{x}_i^T\bfbeta)\\
 g''(\beta_0) & = \frac{n}{\sigma^2} \Rightarrow \beta_0^{\sigma^2} = \frac{\sigma^2}{n}
\end{split}
\end{equation}

\section{Conditional distribution of $\sigma^2$}

\begin{equation}
\begin{split}
\sigma^2 | \beta_0, \bfbeta, \bfgamma, \lambda, \bfx & : \frac{1}{(\sigma^2)^{\frac{n}{2}+\nu+ 1}} \exp \Bigg\{\sum_{i=1}^n \frac{(y_i - \beta_0- \bfbeta^T\mathbf{x}_i)^2 - 2\nu}{2 \sigma^2} \Bigg\} \\
\sigma^2 | \beta_0, \bfbeta, \bfgamma, \lambda, \bfx & \sim IG \Bigg( \frac{n}{2}+\nu , \frac{1}{2}\sum_{i=1}^n (y_i - \beta_0- \bfbeta^T\mathbf{x}_i) - \nu \Bigg)
\end{split}
\end{equation}

\section{Conditional distribution of $\lambda$}

\begin{equation}
\begin{split}
\lambda | \sigma^2, \beta_0, \bfbeta, \bfgamma, \bfx & : \lambda ^{p + c -1} \exp \Bigg\{-\lambda \Bigg[d + \frac{1}{2} \sum_{j=1}^p\gamma_j \Bigg] \Bigg\} \\
\lambda | \sigma^2, \beta_0, \bfbeta, \bfgamma, \bfx & \sim \Gamma \Bigg(p + c, d + \frac{1}{2} \sum_{j=1}^p\gamma_j \Bigg)
\end{split}
\end{equation}

\section{Conditional distribution of $\gamma_j$}

\begin{equation}
\begin{split}
\gamma_j | \lambda, \sigma^2, \beta_0, \bfbeta, \bfx & :\exp \Bigg\{- \frac{\lambda}{2}\gamma_j \Bigg\} \frac{1}{\sqrt{\gamma_j}} \exp \Bigg\{- \frac{1}{2\gamma_j}\beta_j^2 \Bigg\} \\
& = \frac{1}{\sqrt{\gamma_j}} \exp - \Bigg\{ \frac{\lambda}{2}\gamma_j + \frac{\beta_j^2}{2\gamma_j} \Bigg\} \\
& = \gamma^{-\frac{1}{2}} \exp  \Bigg\{ -\frac{1}{2}\Bigg[ \lambda\gamma_j + \frac{\beta_j^2}{\gamma_j}\Bigg]  \Bigg\} \\
\gamma_j | \lambda, \sigma^2, \beta_0, \bfbeta, \bfx &  \sim \text{GIG} \Bigg(p =  \frac{1}{2}, a = \lambda, b = \beta_j^2 \Bigg)
\end{split}
\end{equation}

\section{Conditional distribution of $\beta_j$}

 Here, we will replicate the process done for $\beta_0$ to discover the parameters of $\beta_j$'s distribution:
 
\begin{equation}
\begin{split}
\beta_j | \gamma_j, \lambda, \sigma^2, \beta_0, \bfx & :\exp \Bigg\{- \underbrace{ \Bigg[\frac{1}{2\gamma_j}\beta_j^2 + \frac{1}{2\sigma^2} \sum_{i=1}^n (y_i - \beta_0 - \beta_1 x_i^{(1)} - \dots - \beta_j x_i^{(j)} - \dots - \beta_p x_i^{(p)} )^2 \Bigg]}_{g(\beta_j)} \Bigg\} \\
g'(\beta_j) & = \frac{1}{2 \gamma_j}2 \beta_j + \frac{1}{2 \sigma^2} \sum_{i=1}^n (y_i - \beta_0 - \bfbeta^T\mathbf{x}_i)2(-x_i^{(j)}) \\
& = \frac{\beta_j}{\gamma_j} - \frac{1}{\sigma^2} \sum_{i=1}^n x_i^{(j)} (y_i - \beta_0 - \beta_1 x_i^{(1)} - \dots - \beta_j x_i^{(j)} - \dots - \beta_p x_i^{(p)}) \\
& = \frac{\beta_j}{\gamma_j} - \frac{1}{\sigma^2} \sum_{i=1}^n x_i^{(j)} (y_i - \beta_0 - \bfbeta_{(-j)}^T \mathbf{x}_i^{(-j)} - \beta_jx_i^{(j)} ) \\
& = \frac{\beta_j}{\gamma_j} - \frac{1}{\sigma^2} \sum_{i=1}^n x_i^{(j)} (y_i - \beta_0 - \bfbeta_{(-j)}^T \mathbf{x}_i^{(-j)}) + \frac{1}{\sigma^2} \beta_j \sum_{i=1}^n \Bigg[x_i^{(j)} \Bigg]^2 = 0 \\
\beta_j^M & = \Bigg[ \frac{1}{\gamma_j} + \frac{1}{\sigma^2} \sum_{i=1}^n \Bigg[x_i^{(j)} \Bigg]^2 \Bigg]^{-1}\frac{1}{\sigma^2} \sum_{i=1}^n x_i^{(j)} (y_i - \beta_0 - \bfbeta_{(-j)}^T \mathbf{x}_i^{(-j)}) \\
g''(\beta_j) & = \frac{1}{\gamma_j} + \frac{1}{\sigma^2} \sum_{i=1}^n \Bigg[x_i^{(j)} \Bigg]^2 \Rightarrow \beta_j^{\sigma^2} = \Bigg[ \frac{1}{\gamma_j} + \frac{1}{\sigma^2} \sum_{i=1}^n \Bigg[x_i^{(j)} \Bigg]^2 \Bigg]^{-1} \\
\beta_j | \gamma_j, \lambda, \sigma^2, \beta_0, \bfx & \sim N(\beta_j^M, \beta_j^{\sigma^2})
\end{split}
\end{equation}

\chapter{Automatic Relevance Determination Calculations}\label{Appendix:C}

Here we will develop the calculations mentioned in Chapter \ref{vi_ard}. Let us begin with $\q^*(\bfbeta,\tau,\bfalpha)$.

First let us remember the posterior distribution, where $\bbfy$ can mean either Arousal or Valence, since they are modeled identically. It is also worth reminding that the normal distribution for $\bbfy$ is parametrized by the precision, rather than variance.

\begin{equation} \label{eq:posterior_model_ard_1_ap}
\begin{split}
\p  (\bbfy, \bfbeta, \tau, \bfalpha| \bfx)&  = \left[ \prod_{i=1}^n \text{N}( y_i|\mathbf{x_i}^T\bfbeta, \tau) \right] \times [\text{N}(\bfbeta|\mathbf{0}, \tau \diag(\bfalpha)) \Gamma(\tau|a_0,b_0)] \times \\
& ~~~~ \left[\prod_{d=1}^p \Gamma(\alpha_d|c_0, d_0) \right] 
\end{split}
\end{equation}

Recalling that the variational family will be factorized as

\begin{equation}
\q(\bfbeta, \tau, \bfalpha ) = \q(\bfbeta, \tau) \prod_{j = 1}^{p} \q( \alpha_j )
\end{equation}

Where each $\q(\alpha_j)$ is independent.

Now let us find the CAVI's updating rules  $\q^*(\bfbeta, \tau)$ and  $\q^*(\bfalpha)$. Both calculations are somewhat straightforward, with the posterior being reorganized. It is worth noting that we progressively discarded values that do not relate to the variables in question as the calculations advanced and all expected values were calculated with respect to the variational distribution.

Finding $\q^*(\bfbeta, \tau)$:

\begin{equation} \label{eq:q-beta-tau_ap}
\begin{split}
\q^*(\bfbeta, \tau) & \propto \exp\{\bbe_{\bfalpha}[\log \p  (\bbfy, \bfbeta, \tau, \bfalpha|\bfx)]\} \\
& \propto \exp\{\bbe_{\bfalpha}[\log \p  (\bbfy| \bfbeta, \tau, \bfx) + \log \p  (\bfbeta, \tau | \bfalpha) + \log \p  (\bfalpha)] \} \\
& \propto \exp\{\bbe_{\bfalpha}[\log \p  (\bbfy| \bfbeta, \tau, \bfx) + \log \p  (\bfbeta, \tau | \bfalpha)] \} \\
& \propto \exp \Bigg\{\bbe_{\bfalpha} \Bigg[\sum_{i=1}^n \Bigg[\frac{1}{2} \log(\tau) - \frac{1}{2} \tau (\bbfy_i - \mathbf{x_i}^T\bfbeta)^2  \Bigg] + \frac{1}{2} \log(\tau^p\alpha_1, \dots, \alpha_p) - \\
& ~~~~ \frac{1}{2} \bfbeta^T\Sigma^{-1}\bfbeta + (a_0 - 1) \log(\tau) -b_0\tau \Bigg] \Bigg\} \\
& \propto \exp \Bigg\{ \frac{n}{2} \log(\tau) - \frac{\tau}{2} \sum_{i=1}^n (\bbfy_i - \mathbf{x_i}^T\bfbeta)^2 + \frac{p}{2}\log(\tau) + \\
& ~~~~\frac{1}{2}\bbe_{\bfalpha}[\log(\alpha_1, \dots, \alpha_p)] -  \frac{1}{2}\bfbeta^T \bbe_{\bfalpha} [\Sigma^{-1}]\bfbeta + (a_0 -1) \log(\tau) - b_0 \tau\Bigg\} \\
& \propto \exp \Bigg\{\log(\tau) \Bigg[\frac{n}{2} + \frac{p}{2} + a_0 -1 \Bigg] - \frac{\tau}{2} \Bigg[\sum_{i=1}^n (\bbfy_i - \mathbf{x_i}^T\bfbeta)^2 - \\
& ~~~~ \bfbeta^T \bbe_{\bfalpha} [\diag(\bfalpha)]\bfbeta + 2b_0 \Bigg] \Bigg\} \\
& \propto \exp \Bigg\{ \log(\tau) \Bigg[\frac{n}{2} + \frac{p}{2} + a_0 - 1 \Bigg] - \frac{\tau}{2} \Bigg[\bfbeta^T \Bigg(\sum_{i=1}^n \mathbf{x_i}\mathbf{x_i}^T - \bbe_{\bfalpha}[\diag(\bfalpha)] \Bigg)\bfbeta \\
& ~~~~ -2\bfbeta^T \sum_{i=1}^n \mathbf{x_i}\bbfy_i + \sum_{i=1}^n \bbfy_i^2 + 2b_0  \Bigg]\Bigg\} \\
& = \exp \Bigg\{\log(\tau) \Bigg[\frac{n}{2} + \frac{p}{2} + a_0 - 1 \Bigg] - \frac{\tau}{2}\Bigg[(\bfbeta - \bfbeta_*)^T V_*^{-1}(\bfbeta - \bfbeta_*) - \\
& ~~~~ \bfbeta_*^T V_*^{-1} \bfbeta_* +  \sum_{i=1}^n \bbfy_i^2 + 2b_0 \Bigg]\Bigg\} \\
& = \underbrace{\Bigg[\tau^{\frac{p}{2}} e^{-\frac{\tau}{2}(\bfbeta - \bfbeta_*)^T V_*^{-1}(\bfbeta - \bfbeta_*)} \Bigg]}_{\text{N}(\bfbeta|\bfbeta_*; \tau V_*^{-1})} \underbrace{\Bigg[\tau^{\frac{n}{2} + a_0 - 1} e^{- \tau(b_0 + \frac{1}{2}\sum_{i=1}^n \bbfy_i^2 - \frac{1}{2}\bfbeta_*V_*^{-1}\bfbeta_*)} \Bigg]}_{\Gamma(\tau| a_*, b_*)} \\
\end{split}
\end{equation}

Now we need to find $\q^*(\alpha_d)$:

\begin{equation}
\label{eq:q-alpha-model-ard-1_ap}
\begin{split}
\q^*(\alpha_d) & \propto \exp \Bigg\{\bbe_{-\alpha_d} \Bigg[\log \p  (\bbfy, \bfbeta, \tau, \bfalpha|\bfx) \Bigg]   \Bigg\} \\
& = \exp \Bigg\{ \bbe_{-\alpha_d} \Bigg[\log \p  (\bbfy| \bfbeta, \tau, \bfalpha, \bfx) + \log \p  (\bfbeta, \tau| \bfalpha) + \log \p  (\bfalpha) \Bigg]   \Bigg\} \\
& \propto \exp \Bigg\{\bbe_{-\alpha_d} \Bigg[\log (\text{N}(\bfbeta|\mathbf{0}, \tau \diag(\bfalpha))\Gamma(\tau|a_0,b_0) + \log (\Gamma(\alpha_d|c_0,d_0)) \Bigg]   \Bigg\} \\
& \propto \exp \Bigg\{\bbe_{-\alpha_d} \Bigg[ \frac{1}{2} \log(\tau^p \alpha_{1} \dots \alpha_{p}) - \frac{1}{2} \bfbeta^T \Sigma^{-1}\beta + (a_0-1)\log(\tau) - b_0 \tau + \\
& ~~~~ (c_0-1) \log(\alpha_d) -d_0\alpha_d \Bigg]   \Bigg\} \\
& \propto \exp \Bigg\{ \frac{p}{2} \bbe_{-\alpha_d} \Bigg[\log(\tau^p)\Bigg] + \frac{1}{2} \bbe_{-\alpha_d} \Bigg[\log(\alpha_{1} \dots \alpha_{p}) \Bigg]  -\frac{1}{2} \bbe_{-\alpha_d} \Bigg[\bfbeta^T \Bigg]\Sigma^{-1}\bbe_{-\alpha_d} \Bigg[\bfbeta \Bigg] + \\ &\qquad\qquad\qquad(a_0-1)\bbe_{-\alpha_d} \Bigg[\log(\tau)\Bigg] - b_0\bbe_{-\alpha_d} \Bigg[\tau \Bigg] + (c_0 - 1)\log(\alpha_d) - d_0\alpha_d \Bigg\} \\
& \propto \exp \Bigg\{\frac{p}{2}\log(\alpha_d) - \frac{1}{2}\alpha_d \bbe_{-\alpha_d} \Bigg[\tau \bfbeta^T\bfbeta \Bigg] + (c_0 - 1) \log(\alpha_d) - d_0 \alpha_d \Bigg\} \\
& \propto \exp \Bigg\{\log(\alpha_d) \Bigg(\frac{p}{2} + c_0 - 1 \Bigg) - \alpha_d \Bigg(d_0 - \frac{1}{2} \bbe_{-\alpha_d}\Bigg[\tau \bfbeta^T\bfbeta \Bigg] \Bigg) \Bigg\} \\
& = \exp \underbrace{\Bigg\{\log(\alpha_d) \Bigg(\frac{p}{2} + c_0 - 1 \Bigg) - \alpha_d \Bigg(d_0 - \frac{1}{2} V_* + \bfbeta_*^T\bfbeta_* \frac{a_*}{b_*} \Bigg) \Bigg\}}_{\log(\Gamma(\alpha_d|c_*,d_*))} \\
\end{split}
\end{equation}

After computing the update formulas in Equation \ref{eq:elbo}, we find that:

\begin{equation} \label{eq:ard-model-1-dists_ap}
\begin{split}
\q(\bfbeta, \tau) & \sim \text{N}\left(\bfbeta|\bfbeta_*, \tau V_*^{-1} \right)\Gamma\left(a_*, b_*\right)\\ \q(\alpha_j) & \sim \Gamma (c_{d*},d_{d*}),~j = 1, \dots, p 
\end{split}
\end{equation}
where

\begin{equation} \label{eq:parameters-ard-model-1_ap}
\begin{split}
\bfbeta_* & = V_*\Bigg[\sum_{i=1}^n \mathbf{x_i}\bbfy_i \Bigg] \\
V_*^{-1} & = \sum_{i=1}^n \mathbf{x_i}\mathbf{x_i}^T + \diag(c_{1*}/d_{1*}, \dots, c_{p*}/d_{p*}) \\
a_* & = a_0 + \frac{n}{2} \\
b_* & = b_0 + \frac{1}{2}\Bigg(\sum_{i=1}^n \bbfy_i^2 - \bfbeta_*^T V_*^{-1}\bfbeta_*\Bigg) \\
c_{j*} & = c_0 + \frac{1}{2},~j = 1, \dots, p \\
d_{j*} & =  d_0 + \frac{1}{2} \Bigg[[V_*]_{jj} + \beta_{*j}^2 \frac{a_*}{b_*}\Bigg],~j = 1, \dots, p.\\
\end{split}
\end{equation}

\begin{comment}
Notas sobre a diagramação do TCC:
* Matrizes: maiúsculas em negrito
* Vetores: minúsculas em negrito
* Dados: $\bfx$
* Distribuição dos $\alpha$'s: $\prod_{i=1}^p \alpha_d^{C_0 - 1} \exp^{-\alpha_d_0}$
\end{comment}

%text ------------------------------------------------------- 

\normalsize
\bibliographystyle{apalike}{}
\bibliography{references.bib}

\end{document}